%% file: 0-main.tex
\documentclass{article}

\usepackage{my-icml24}

\icmltitlerunning{Simplicity Bias of Two-Layer Networks beyond Linearly
Separable Data}

\begin{document}

\twocolumn[
\icmltitle{Simplicity Bias of Two-Layer Networks beyond Linearly Separable
Data}



\icmlsetsymbol{equal}{*}

\begin{icmlauthorlist}
\icmlauthor{Nikita Tsoy}{insait}
\icmlauthor{Nikola Konstantinov}{insait}
\end{icmlauthorlist}

\icmlaffiliation{insait}{INSAIT, Sofia University, Bulgaria}

\icmlcorrespondingauthor{Nikita Tsoy}{nikita.tsoy@insait.ai}

\icmlkeywords{simplicity bias, two-layer network, gradient flow, out-of-distribution}

\vskip 0.3in
]



\printAffiliationsAndNotice{} 

\begin{abstract}
    Simplicity bias, the propensity of deep models to over-rely on simple
    features, has been identified as a potential reason for limited
    out-of-distribution generalization of neural networks  \citep{s20p}.
    Despite the important implications, this phenomenon has been theoretically
    confirmed and characterized only under strong dataset assumptions, such as
    linear separability \cite{l21g}. In this work, we characterize simplicity
    bias for general datasets in the context of two-layer neural networks
    initialized with small weights and trained with gradient flow.
    Specifically, we prove that in the early training phases, network features
    cluster around a few directions that do not depend on the size of the
    hidden layer. Furthermore, for datasets with an
    XOR-like pattern, we precisely identify the learned features and
    demonstrate that simplicity bias intensifies during later training stages.
    These results indicate that features learned in the middle stages of
    training may be more useful for OOD transfer. We support this hypothesis
    with experiments on image data.
\end{abstract}

\input{1-intro}

\input{2-rel-work}

\input{3-set}

\input{4-gen}

\input{5-spec}
\input{5b-spec-emp}

\input{6-exper}

\input{7-conc}

\section*{Acknowledgments}

This research was partially funded from the Ministry of Education and Science  of Bulgaria (support for INSAIT, part of the Bulgarian National Roadmap for  Research Infrastructure). The authors thank Ivan Kirev and Kristian Minchev for
their helpful feedback and discussions on this work.

\section*{Impact Statement}

This paper presents work whose goal is to advance the field of Machine
Learning. There are many potential societal consequences of our work, none
which we feel must be specifically highlighted here.

\bibliography{bibliography}

\newpage

\onecolumn

\appendix

\begin{center}
  {\LARGE Supplementary Material}
\end{center}

The supplementary material is structured as follows.
\begin{itemize}
    \item \cref{sec:add-notation} explains some additional notation used in
        proofs.
    \item \cref{sec:proof-gen-phase1} contains the proof of
        \cref{thm:gen-phase1}.
    \item \cref{sec:proof-gen-phase2} contains the proof of
        \cref{thm:gen-phase2}.
    \item \cref{sec:proof-spec} contains the proofs of the results from
        \cref{sec:spec}.
    \item \cref{sec:conv-graphs} contains the additional experimental results
        for \cref{sec:spec}.
    \item \cref{sec:exper-details} contains additional experimental results for
        \cref{sec:exper}.
\end{itemize}

\section{Additional Notation}
\label{sec:add-notation}

Denote $\g(\v) \defeq \nabla_{\v} G(\v)$.

\input{proof-gen-phase1}

\input{proof-gen-phase2}

\input{proof-spec-phase12}
\input{proof-spec-phase3}

\input{conv-graphs}

\input{exper-details}

\end{document}

%% file: 1-intro.tex
\section{Introduction}

Out-of-distribution (OOD) generalization is a key challenge towards
the widespread adoption of machine learning. Specifically, since training data
may not always cover all possible test scenarios, networks often rely on
shortcuts: spurious rules that hold on the training distribution but not in
more complicated real-world situations \citep{g20s}. For example,
convolutional networks often prioritize texture over shape \citep{ge18i}, or
transformers might rely on simplistic heuristics in natural language
inference \citep{m19r}. 

One possible mechanism behind shortcuts is \textit{simplicity bias}, the
propensity of neural networks to rely only on ``simple'' features. As
\citet{s20p} demonstrated, this bias might be persistent and hurt OOD
generalization in image classification tasks. Simplicity bias is also a
peculiar phenomenon from a theoretical perspective. Since many neural
architectures are universal function approximators \citep{c89a,h89m}, one might
hope that models will learn other, more sophisticated patterns within the
data.

Despite the importance of simplicity bias, a thorough theoretical understanding
of this phenomenon is still lacking. To the best of our knowledge, existing
works on simplicity bias \citep{l21g,s22o,m23s} only demonstrate its emergence
by employing stringent assumptions, which restrict training data to be linearly
separable or one-dimensional.

\paragraph{Contributions} We give the first proof of the existence of
simplicity bias and a precise mathematical characterization of the features
learned during training beyond linearly separable data. We do so for
two-layer neural networks trained with gradient flow from a small
initialization, a model popular in the theoretical literature
\cite{l21t,l21g,b22g}. We characterize simplicity bias as a property that
only a small set of \emph{prominent} neurons governs the behavior of the
network. These prominent neurons cluster in several directions, which do not
depend on the size of the hidden layer.

Specifically, our theoretical analysis divides training into three stages.
During the first two stages (Section 4), in which the weights grow from small
to constant scale, we prove for general datasets that the most prominent
features recovered by the training dynamics cluster around the extrema of a
data-dependent function, which does not depend on the number of neurons and
disentangles the interactions between them. For the last stage, where the
network converges to zero loss, we prove that simplicity bias can become
extreme even in non-linearly-separable datasets. In this stage, we cover the
case of XOR-like data under an assumption about the convergence of a 4-neuron
network, which we validate experimentally. On a methodological level, our work
generalizes the analysis of \citet{l21g} beyond linearly separable data and
provides an implicit description of the most prominent features for general
datasets.

Our theoretical findings additionally lead to a hypothesis with potential
practical implications: networks trained to a very small loss may be prone to
stronger simplicity bias and, hence, may be harder to finetune to new tasks. We
test this intuition with experiments on a domino dataset of MNIST-CIFAR10 pairs
\citep{s20p} and observe experimental support for our
hypothesis.\footnote{Replication files are available at
\url{https://github.com/nikita-tsoy98/simplicity-bias-beyond-linear-replication}}

%% file: 2-rel-work.tex
\section{Related Work}

\paragraph{Simplicity bias of features} The simplicity bias of neural
networks was observed and linked to generalization and OOD performance
in prior work \citep{v18d,s20p}. Several works show that two-layer networks
provably learn a linear decision boundary on linearly separable datasets
\citep{b18s,p20a,s21t,p21t,l21g,e22a,f23i,k23i,m23s,w23u,c23l,m24e}. While we
use some of the techniques developed by these works, we focus on the
non-linearly-separable case, which requires further theoretical analysis.
\citet{s22o} prove that, on one-dimensional data, two-layer networks converge
to a model with few linear regions. In contrast, we study datasets in
$\mathbb{R}^d$. \citet{b19w} also show simplicity bias for XOR-like data, but
only for a 4-point dataset in $\mathbb{R}^2$, while we focus on datasets of
arbitrary size in $\mathbb{R}^d$.

\paragraph{Simplicity bias of training dynamics} A direction related to our
work is simplicity bias in terms of training dynamics, the propensity of neural
networks to learn simple patterns first in training. This property was observed
in several empirical works \citep{a17a,x18t,r19o,k19s}
and demonstrated theoretically in certain settings \citep{a19f,b20f,l21t,b22i}.
Another closely related topic is the distributional simplicity bias proposed by
\citet{r23n}. In contrast to these works, we seek to characterize the
learned features instead of analyzing some complexity invariant.

\paragraph{Small initialization and initial condensation} Our results in
\cref{sec:gen} provably demonstrate initial condensation, the propensity of
neural networks to condense neurons in few directions during early stages
of training, for two-layer networks with small initialization. Several
theoretical works have analyzed this phenomenon previously. \citet{m18g}
study condensation in regression and classification problems with two-layer
ReLU networks. While they also recognize the function $G$ we use to describe
the network dynamics as an important factor in learning, their derivations and
discussions of the near-zero initialization regime, which we study
in our paper, are informal. In addition, their bounds for the speed of neuron
alignment
are insufficient to differentiate between prominent and non-prominent neurons,
as we do in \cref{sec:gen}. \citet{z22t} give another theoretical description
of condensation in regression tasks. However, their results do not apply to
ReLU activation or our differentiable approximation of ReLU (due to
irregularity at $0$). \citet{b22g} also identify the condensation phenomenon in
regression tasks, but only for orthogonal data, while we put considerably
milder assumptions on the data. Additionally, \citet{x23o} recognize the
condensation phenomenon in a regression setting with a one-neuron teacher
network, which is similar to a linearly separable case because the direction of
the gradients will be correlated with the direction of the teacher neuron.

Finally, the concurrent works of \citet{b24e} and \citet{k24d} study the
initial condensation and prove the result similar to our \cref{thm:gen-phase1}.
However, these results do not quantify the differences in the neurons' growth
rates and, hence, can not differentiate between prominent and non-prominent
neurons. Moreover, there are several technical differences. \citet{b24e} work
directly with (leaky) ReLU and make milder assumptions on the initialization,
but require the function $G$ to have no saddle points and only show alignment
results for neurons that satisfy a certain technical condition (Condition 1 in
their manuscript). \citet{k24d} work in a more general setting of
two-homogeneous neural networks, but require their initialization to be
non-branching.

\paragraph{Small initialization and mean-field regime} Since we work in the
small initialization regime of two-layer networks, our setting is similar to
the mean-field regime \citep{c18o,m18a,w19r,l20l,g21u}. However, in our
setting, we do not increase the number of neurons when we decrease the
initialization scale \citep[which corresponds to the condensed regime in the
classification of][]{l21p}. Thus, in our limit, at the beginning of training,
the neurons evolve independently, while, in the mean-field limit, they interact
via the velocity field (the derivative of a loss function). At the same time,
some techniques in our work are similar to the mean-field techniques since both
we and the mean-field works analyze the behavior of a loss function near zero.

\paragraph{Neural collapse} The constraint on features imposed by simplicity
bias might resemble the phenomenon of neural collapse \citep{p20p}, under which
last-layer features collapse to a set of class-specific highly symmetric
features. While these phenomena might be related, they also have some important
differences. First, simplicity bias occurs early in training, while neural
collapse occurs at the end of training. Second, under simplicity bias, the
features are not necessarily associated with some class, and the neural network
might not classify all examples correctly. Third, the mechanisms behind these
phenomena are probably different. While the suggested mechanism for feature
collapse arises from the unconstrained feature model \citep[e.g.,][]{z21n},
which focuses on the last-layer classifier, our mechanism for simplicity bias
arises from the implicit bias of SGD, i.e., the interaction of the first- and
last-layer weights during training.

%% file: 3-set.tex
\section{Setting}
\label{sec:set}

Throughout the paper, we analyze feature learning on a binary classification
problem with a two-layer network initialized with small random parameter
values.

\paragraph{Notation} We use the following notation:
$v^j$ is $j$th component of vector $\v$, so that $\v = (v^1, v^2, \dots,
v^d)^\tran$, $\norm{\v} \defeq \sqrt{\sum_{j=1}^d (v^j)^2}$ is the usual
$l_2$-norm of $\v$, $\vu \defeq \frac{\v}{\norm{\v}}$ is the unit direction of
$\v$, $\P_{\v} \defeq \I - \vu \vu^\tran$ is the projector on the space
orthogonal to $\v$, $[k] \defeq \{1, \dots, k\}$, $\S^{d-1} \defeq \Bc*{\x \in
\mathbb{R}^d \given \norm{\x} = 1}$ is the unit sphere, $\D^d \defeq \Bc*{\x
\in \mathbb{R}^d \given \norm{\x} < 1}$ is the unit disk.

\paragraph{Objective} Denote by $f(\prm, \cdot)$ a network parameterized by
$\prm$. The sign of $f$ stands for the classification result. Denote by $D
\defeq (\x_i \in \mathbb{R}^d, y_i \in \{-1, 1\})_{i=1}^n$ an arbitrary
training dataset, such that $\forall i \: \norm{\x_i} \le 1$. We consider
networks trained to minimize the cross-entropy loss
\[
  L(\prm) \defeq \frac{1}{n} \sum_{i=1}^n \ell(f(\prm, \x_i) y_i), \text{ where
  } \ell(z) \defeq \ln(1 + \e^{-z}).
\]

\paragraph{Architecture} We consider two-layer networks
\[
  f(\prm, \x) \defeq \sum_{j=1}^m u_j \phi(\v_j, \x),
\]
where $\prm \defeq (u_1, \dots, u_m, \v_1^\tran, \dots, \v_m^\tran)^\tran$,
$u_j \in \mathbb{R}$ and $\v_j \in \mathbb{R}^d$, are the network parameters
and $\phi$ is an activation function. We denote $\forall A \subseteq [m] \:
\norm{\prm}_A \defeq \max_{j \in A} \max(\abs{u_j}, \norm{\v_j})$.

One of the most commonly used activation functions is ReLU, for which $\phi(\v,
\x) = (\v^\tran \x)_+ = \max (\v^\tran \x, 0)$. In our paper, for the technical
reasons, we need the activation to be smooth to avoid some degenerate cases in
the dynamics of gradient flow. Since this property does not hold for ReLU, we
consider a
differentiable approximation,
\[
  \phi(\v, \x) \defeq \phi_{Q, \xi}(\v, \x) \defeq \int_{\mathbb{R}^d}
  (\v^\tran (\x + \xi \z))_+ Q(\d \z),
\]
where $\xi > 0$, $Q$ is the uniform measure on $\D^d$, and $(z)_+ \defeq
\max(0, z)$. For the purposes of our analysis, $\xi$ could be set to be much
smaller than machine precision. Thus, there is no practical difference between
our activation and ReLU. We also note that in our experiments in Sections
\ref{sec:exper-b} and \ref{sec:exper} we use the usual ReLU activation.

We call a function $f(x,y)$ $k$-positively homogeneous in $x$ if $f(cx, y) =
c^k f(x,y)$ for all vectors $x,y$ and all $c>0$. Notice that $\phi(\v, \x)$ is
1-positively homogeneous in $\v$, i.e., $\forall c > 0 \: \phi(c \v, \x) = c
\phi(\v, \x)$. This property implies that $f(\prm, \x)$ is 2-positively
homogeneous in $\prm$.

\paragraph{Optimization} We consider the training of $f$ via gradient flow,
$\odv{\prm}{t} = -\nabla L(\prm)$, which implies dynamics
\begin{equation}
  \label{eq:set-gf}
  \begin{aligned}
    \odv{u_j}{t} &= \frac{1}{n} \sum_{i=1}^n (-\ell'(f(\prm, \x_i) y_i))
    \phi(\v_j, \x_i) y_i,\\
    \odv{\v_j}{t} &= \frac{u_j}{n} \sum_{i=1}^n (-\ell'(f(\prm, \x_i) y_i))
    \nabla_{\v} \phi(\v_j, \x_i) y_i.
  \end{aligned}
\end{equation}
We initialize the system with a small weights $\prm(0)
= \sigma \prm^0$, where $\sigma \approx 0$, similarly \citet{l21g}. For
simplicity, we assume that $\abs{u^0_j} = \norm{\v^0_j}$. The results of
\citet{d18a} imply that $u_j^2 - \norm{\v_j}^2 = \text{const}$. Then,
$\sign(u_j) \eqdef s_j$ is constant \citep[Lemma 1,][]{b22g}.

%% file: 4-gen.tex
\section{Simplicity Bias for General Data}
\label{sec:gen}

First, we analyze \cref{eq:set-gf} in the early (\cref{sec:small_scale}) and
middle (\cref{sec:const_scale}) training phases, in which $\prm$ remains small
or grow to a constant scale, respectively. These phases correspond to Phases 1
and 2 of \citet{l21g}.

\paragraph{Key challenge} The key challenge in
characterizing the features learned throughout training is the lack of
universal training invariants to trace. To compensate for this, most of the
works that describe features of neural networks make structural assumptions
about the dataset, such as linear separability \citep{l21g}, orthogonality or
near-orthogonality \citep[e.g.,][]{b19w,p21t,f23r} or high-dimensionality
\citep[e.g.,][]{ba22h}.

In contrast to these works, we do not make structural assumptions about the
data. Instead, we exploit that, for small weights, a simpler data-dependent
function, $G$, which does not depend on the number of neurons, can approximate
the network dynamics. Specifically, we link the features learned by the
original system to the global extrema of $G$. While our results do not
explicitly characterize the learned features, they are sufficient to prove the
presence of simplicity bias.

\subsection{Feature Learning from a Small Initialization}
\label{sec:small_scale}

\paragraph{Disentangling training dynamics} First, we informally motivate our
approximation of \cref{eq:set-gf} for a small initialization. By the mean value
theorem,
\[
    \begin{alignedat}{2}
        && \ell'(f(\prm, \x) y) - \ell'(0) = & \ell''(\zeta) f(\prm, \x) y\\
        \implies && \abs{\ell'(f(\prm, \x) y) - \ell'(0)} \le & \abs{f(\prm,
        \x)} \sup_{z \in \mathbb{R}} \abs{\ell''(z)},
    \end{alignedat}
\]
for some $\zeta \in [0, f(\prm, \x) y]$. Since $f(\prm, \x)$ is 2-homogeneous,
we get $\abs{\ell'(f(\prm, \x) y) - \ell'(0)} = \O(\norm{\prm}^2)$. Thus, when
$\sigma \approx 0$, \cref{eq:set-gf} behaves similarly to the following
system with linearized loss \citep{m18g},
\begin{equation}
  \label{eq:gen-gf-lin}
  \begin{aligned}
    \odv{\ul_j}{t} &\defeq G(\vl_j) \defeq \frac{1}{n} \sum_{i=1}^n (-\ell'(0))
    \phi(\vl_j, \x_i) y_i,\\
    \odv{\vl_j}{t} &\defeq \ul_j \nabla G(\vl_j),
  \end{aligned}
\end{equation}
where $\ul_j(0) = u_j(0)$ and $\vl_j(0) = \v_j(0)$.

Note that the neurons in \cref{eq:gen-gf-lin} evolve independently of each
other, while the neurons in \cref{eq:set-gf} interact via $\ell'(f(\prm, \x_i)
y_i)$. This property significantly facilitates the analysis of
\cref{eq:gen-gf-lin} compared to the original system.

The following theorem formalizes the link between the two systems and links the
features learned by the original dynamics (\ref{eq:set-gf}) to the global
extrema of the function $G$.

\begin{theorem}[Proof in \cref{sec:proof-gen-phase1}]
  \label{thm:gen-phase1}
  Assume that $\prm$ follows \cref{eq:set-gf}, $\forall i \:
  \norm{\x_i} \le 1$, $d \ge 2$, and $d$ is odd. Then $\exists \kappa^* > 0, P
  \subseteq [m], (\kappa_j > 0, u^*_j \in \mathbb{R}, \vu^*_j \in
  \S^{d-1})_{j=1}^m$ such that for $\sigma = r^{1 + \kappa^*}$, $T_1
  \defeq \frac{1}{\lambda} \ln\Par*{\frac{r}{\sigma}}$, and $r \to 0$, we get
  \[
    \begin{aligned}
        \forall j \in P && \abs{u_j(T_1) - r u^*_j} \le & \O(r^{1 +
        \kappa^*}),\\
        && \norm{\vu_j(T_1) - \vu^*_j} \le & \O(r^{\kappa^*}), \: s_j
        G(\vu^*_j) = \lambda,\\
        \forall j \in R && \abs{u_j(T_1)} = \|v_j(T_1)\| \le & \O(r^{1 +
        \kappa_j}),
    \end{aligned}
  \]
  where $R \defeq [m] \setminus P$, $\lambda \defeq \max_{\vu \in
  \S^{d-1}} \abs{G(\vu)}$ and $G$ is defined by \cref{eq:gen-gf-lin}.

  Moreover, to ensure that a particular global extrema $\vu^*$ of $\abs{G}$
  ($\abs{G(\vu^*)} = \lambda$) is captured ($\exists j \in P : \vu^*_j =
  \vu^*$) with probability at least $1 - \delta$ over isotropic initialization
  of $\prm^0$, we need $m = \O(-\ln(\delta))$ neurons, where the constants in
  the big-O notation depend only on the properties of data.
\end{theorem}

\begin{remark}
    We expect that the result can be extended to the case of even $d$. However,
    this extension will require exploiting the concept of o-minimal structures
    and proving a corresponding Lojasiewicz inequality for the
    o-minimal structure containing $\arcsin$ function (see our proof,
    \citet{j20d}, and Example 1.5 of \citet{l10l}). Since such an
    analysis is not necessarily informative from a machine learning
    perspective, we stick to $d$ being odd for simplicity.
\end{remark}

\paragraph{Discussion} \cref{thm:gen-phase1} suggests that, when we start
training from small initialization ($\sigma \to 0$), the neurons either align
with the global extrema of $G$ ($j \in P$) or grow very slowly ($j \in R$). In
particular, for the neurons in $R$, $\abs{u_j(T_1)} = \norm{v_j(T_1)} = \O(r^{1
+ \kappa_j})$. Therefore, their contribution to the decision boundary is
negligible compared to the \textit{prominent} neurons in $P$, for which
$\abs{u_j(T_1)} = \norm{\v_j(T_1)} = \Theta(r \abs{u_j^*}) = \Theta(r)$. Thus,
at the start of the training, the network exhibits simplicity bias: regardless
of the number of neurons, $m$, only prominent ones contribute to the network's
decision boundary. Moreover, these prominent neurons are aligned with the
global extrema directions of $G$, which do not depend on $m$. When the number
of neurons is sufficiently large ($m = \Omega(-\ln(\delta))$), the network will
learn all global extrema directions of $G$, which makes the characterization
very precise for small $\sigma$.

\subsection{Feature Growth to a Constant Scale}
\label{sec:const_scale}

Next, we extend our analysis beyond the stage studied in \cref{thm:gen-phase1}
to the point when the network weights reach a constant scale. Specifically, we
show that the prominent features preserve their alignment and that the network
essentially behaves like a smaller $p$-neuron network, where $p$ is the number
of extrema of $G$.

\paragraph{Embedding function} To formalize our claim, we consider a
specific smaller network that describes \cref{eq:set-gf} well. Since the
prominent neurons ($j \in P$) cluster around the global extrema of $G$ at the
end of the first phase, we can divide them according to their direction
\[
    \begin{split}
        P = \sqcup_{k=1}^p P_k \text{ s.t. } & \forall k \forall i, j \in P_k
        \vu^*_i = \vu^*_j\\
        & \land \forall k \neq k' \forall i \in P_k, j \in P_{k'} \: \vu^*_i
        \neq \vu^*_j.
    \end{split}
\]
We denote by $\vu^*_{P_k}$ the direction of neurons in $P_k$ and by $s_{P_k}
\defeq \sign(G(\vu^*_{P_k}))$. Now, consider the following auxiliary
system
\begin{equation}
  \label{eq:gen-gf-emb}
  \begin{aligned}
    \odv{\ue_k}{t} &= \frac{1}{n} \sum_{i=1}^n (-\ell'(f(\prme, \x_i)
    y_i)) \phi(\ve_k, \x_i) y_i,\\
    \odv{\ve_k}{t} &= \frac{\ue_k}{n} \sum_{i=1}^n (-\ell'(f(\prme, \x_i)
    y_i)) \nabla_{\v} \phi(\ve_k, \x_i) y_i,
  \end{aligned}
\end{equation}
where
\[
  \ue_k(T_1) = s_{P_k} r \sqrt{\sum_{j \in P_k} (u^*_j)^2}, \:
  \ve_k(T_1) = r \abs{\ue_k(T_1)} \vu^*_{P_k}.
\]
These equations describe the dynamics of a $p$-neuron network $\prme = (\ve_1,
\ve_2, \ldots, \ve_p, \ue_1, \ue_2, \ldots, \ue_p)$, initialized in a way that
preserves the alignment and scale of $\prm$.

Our goal will be to show that each neuron in the original network can be
approximated by corresponding neuron in the $p$-neuron network. To do it,
below, we define an embedding function \citep{l21g} that maps the $p$-neuron
network $\prme$ to a $m$-neuron network $\prmt \defeq \chi(\prme)$
\begin{equation}
  \label{eq:gen-emb}
  \begin{aligned}
      \forall j \in P_k && \ut_j = & \frac{r u^*_j}{\ue_k(T_1)} \ue_k, & \vt_j
      = & \frac{r u^*_j}{\ue_k(T_1)} \ve_k,\\
      \forall j \in R && \ut_j = & 0, & \vt_j = & \vec{0}.
  \end{aligned}
\end{equation}

\paragraph{Propagation of simplicity bias} We prove that the original network
behaves approximately as an image of the embedding above. We only require a
mild assumption on the data that avoids degenerate cases in which a data point
perfectly aligns with an extrema of $G$.
\begin{definition}
  A direction $\vu$ is \emph{$\varDelta$-regular} if $\forall i \:
  \abs{\vu^\tran \x_i} \ge \varDelta$.
\end{definition}
The following result holds under the assumption that all global extrema of $G$
are regular. We note that this is a generalization of Assumption 4.5 of
\citet{l21g} for the case of non-linearly-separable data.

\begin{theorem}[Proof in \cref{sec:proof-gen-phase2}]
    \label{thm:gen-phase2}
    In the setting of \cref{thm:gen-phase1}, consider \cref{eq:gen-gf-emb} and
    assume that $\forall j \in P \: \vu^*_j$ is $(\xi + 2 \varDelta)$-regular.
    Then, $\forall \varepsilon \le \min\{\varDelta, \nicefrac{1}{2}\}$, the
    following holds. First, $\exists \prm^{\varepsilon,*} \defeq \lim_{r \to 0}
    \prme(T^\varepsilon_2)$, where $T^\varepsilon_2 \defeq T_1 +
    t^\varepsilon_4, t^\varepsilon_4 \defeq \frac{1}{2 \lambda}
    \ln\Par[\Big]{\frac{\lambda \varepsilon}{2 a r^2 \norm{\prm^*}_{[m]}^2}}$,
    and $a \defeq \frac{m (1+\xi)^2}{4}$. Second, denote
    $\prm^{\chi,\varepsilon,*} \defeq \chi(\prm^{\varepsilon,*})$, then, as $r
    \to 0$, we have
    \[
        \begin{aligned}
            \forall j \in P && \abs{u_j(T^\varepsilon_2) -
            u^{\chi,\epsilon,*}_j} = &\O(r^\kappa + r^2),\\
            && \norm{\vu_j(T^\varepsilon_2) - \vu^{\chi,\varepsilon,*}_j} = &
            \O(r^\kappa + r^2),\\
            \forall j \in R && \abs{u_j(T^\varepsilon_2)} =
            \norm{\v_j(T^\varepsilon_2)} = & \O(r^{\kappa_j}),
        \end{aligned}
    \]
    where $\kappa \defeq \min\{\kappa^*, \min_{j \in R} \kappa_j\}$.

    Moreover, $\abs{u^{\chi,\varepsilon,*}_j} = \Theta(\sqrt{\varepsilon})$ and
    $\forall j \in P \: \norm{\vu^{\chi,\varepsilon,*}_j - \vu^*_j} \le
    \varepsilon$.
\end{theorem}

\begin{remark}
    The assumption that the critical directions $\vu^*$ are regular is
    essentially needed only to show that the Hessian matrix $\nabla^2
    \abs{G}(\z)$ is negative semi-definite in some $(\xi + 2
    \varDelta)$-neighborhood of $\vu^*$. Since the Hessian matrix is
    semi-negative at $\z = \vu^*$ and $G$ is twice continuously differentiable,
    this assumption appears rather mild.
\end{remark}

\paragraph{Discussion} Similarly to \cref{thm:gen-phase1}, the neurons in $R$
have a negligible effect on the network since $u_j$ and $v_j$ are of smaller
magnitude compared to the remaining weights. \cref{thm:gen-phase2} suggests
that the network experiences the simplicity bias not only at the start of the
training but also until the weights grow to a constant scale
$\Theta(\sqrt{\varepsilon})$. Even in the second phase, the prominent neurons
in $P$ stay near the extrema of $G$, which they learned initially. In addition,
$\prm(T^\varepsilon_2) \approx \prm^{\chi,\varepsilon,*}$ implies that the
original network is an approximate embedding of the $p$-neural network above.

Theorems \ref{thm:gen-phase1} and \ref{thm:gen-phase2} show an
interesting separation of the training dynamics of two-layer networks trained
from small initialization. First, the hidden layer features
traverse the unit sphere until some become prominent, capturing a
supremum direction of $G$ and aligning with it. Then, the prominent features
grow without much change in direction.

%% file: 5-spec.tex
\section{Extreme Simplicity Bias for Specific Data}
\label{sec:spec}

Our results so far describe the features learned by two-layer networks in the
early stages of training, as the parameters go from small to constant scale.
However, as indicated by \citet{l21g,s20p}, the simplicity bias might persist
not only in the initial stages of training but also when the network reaches
perfect accuracy on a training dataset. To test to what extent our mechanism
outlined in \cref{sec:gen} can explain this empirical observation, we extend
our analysis to the infinite time training limit.

A key challenge in this setup is the lack of convergence guarantees for
non-convex models, which necessitates at least some assumptions on the train
data. Given our focus on non-linearly-separable data, we focus on datasets in
$\mathbb{R}^d$ that feature an XOR-like pattern in a $2$-dimensional subspace
as a prime example that breaks the linear separability.

\subsection{Data with an XOR-pattern}
\label{subsec:xor_data_defn}

We consider train data in $\mathbb{R}^d$ that follows an XOR pattern in the
$2$-dimensional subspace generated by coordinate vectors $\ee_1$ and $\ee_2$.
Specifically, the points cluster around four vectors, $\ee_1, -\ee_1, \ee_2,
-\ee_2$. They are symmetric w.r.t. the permutation of the first and second
coordinate, the reflection of the first and second coordinate axes, and the
reflection through the hyperplane generated by the first and second axes.
Points that cluster around the directions $\ee_1$ and $-\ee_1$ have positive
labels; others have negative labels. Finally, we make an additional assumption
on the non-alignment of data points with coordinate axes (which would allow us
to apply \cref{thm:gen-phase2} to this dataset). Notice that similar
assumptions appear in \citet{l21g}, but in our case the resulting dataset is
not linearly separable.

To formalize these assumptions, denote $\mat{P} \defeq \I + (\ee_2 - \ee_1)
\ee_1^\tran + (\ee_1 - \ee_2) \ee_2^\tran$ (permutation of the first and second
coordinates), $\forall a \in \{1, 2\} : \R_a \defeq \I - 2 \ee_a \ee_a^\tran$
(reflection of $a$th coordinate), $\R_r \defeq 2 (\ee_1 \ee_1^\tran + \ee_2
\ee_2^\tran) - \I$ (reflection of the rest of the coordinates), and $D_{\x}
\defeq \{\x_i\}_{i=1}^n$ for $\x_i \in \mathbb{R}^d$. Then, the formal
assumptions will be the following.

\begin{assumption}
  \label{ass:xor}
   Denote $a(k) \defeq k \bmod 2$, $b(k) \defeq
   2 \Floor*{\frac{k-1}{2}} - 1$. There exists $\{S_1, S_2, S_3, S_4\} :
   \sqcup_k S_k = [n]$ such that the following properties hold.
  \begin{enumerate}
    \item $
        \begin{aligned}
            \exists \delta < \nicefrac{\sqrt{2}}{2} : \forall i \in S_k \: &
            \norm{\x_i - b(k) \ee_{a(k) + 1}} \le \delta\\
            & \land y_i = 1 - 2 a(k).
        \end{aligned}$
    \item $\forall a \in \{1, 2, r\} \: \R_a D_{\x} = D_{\x}$.
    \item $\mat{P} D_{\x} = D_{\x}$.
    \item $\exists \varDelta > 0 : \forall i, k \: \abs{\ee_k^\tran \x_i} \ge
      \xi + 2 \varDelta$.
  \end{enumerate}
\end{assumption}

\subsection{Initial stages of training} 
\label{subsec:xor_early_stages}

We first use the results of \cref{sec:gen} to analyze the behavior of a
two-layer network trained from small initialization on our XOR-like dataset.
To apply \cref{thm:gen-phase1}, we first describe the global extrema of $G$.

\begin{lemma}[Proof in \cref{sec:proof-spec-g-extr}]
  \label{lem:spec-g-extr}
  If \cref{ass:xor} holds and $\xi + \delta < \nicefrac{1}{6}$, the function
    $\abs{G}$ have four extrema directions: $\ee_1, -\ee_1, \ee_2, -\ee_2$.
\end{lemma}

\cref{lem:spec-g-extr} and \cref{thm:gen-phase1} imply that, at the start of
the training, the big neurons will converge in the four directions: $\pm \ee_1$
and $\pm \ee_2$. To evaluate the probability of the network capturing all
extrema of $G$, we show the following fact.

\begin{lemma}[Proof in \cref{sec:proof-spec-init}]
    \label{lem:spec-init}
    Assume the setting of \cref{lem:spec-g-extr}. The probability of successful
    initialization that will capture all extrema of $G$ is greater than $(1 -
    h^m)^4 \ge 1 - 4 \Par*{\nicefrac{3}{4}}^m (1 + \O(\delta + \xi)) -
    \O\Par*{\Par*{\nicefrac{9}{16}}^m}$ for $m \to \infty$ and $\delta + \xi
    \to 0$, where $h \defeq 1 - \frac{1}{2} \frac{\Vol(A)}{\Vol(\S^{d-1})}$ and
    $A = \Bc*{\x \in \S^{d-1} \given \ee_1^\tran \x \ge \delta + \xi}$.
\end{lemma}

The lemmas above and Theorems \ref{thm:gen-phase1} and \ref{thm:gen-phase2}
suggest that, with high probability, at the end of Phase 2, $\prm^{\varepsilon,
*}$, the four-neuron approximation of the original network, has features
aligned with the cluster directions of the data, $\norm{\vu^{\varepsilon,*}_k -
b(k) \ee_{a(k)+1}} \le \varepsilon$.

\subsection{Training Dynamics in the Infinite Time Limit}

Now we are interested in the training dynamics on the XOR dataset, beyond the
stage described by \cref{thm:gen-phase2} and as $T\to\infty$. To study that, we
use the fact that $\prme$ provides a good approximation of the original network
$\prm$ and formalize this beyond the time $T^\varepsilon_2$ used in
\cref{thm:gen-phase2}.

The next result proves that the limit dynamics (\ref{eq:set-gf})
of the original network converge to the same features as those learned by
the simpler 4-neuron network from \cref{subsec:xor_early_stages}.
Our result assumes that the features of the 4-neuron network remain aligned
with the extrema of $G$ from \cref{lem:spec-g-extr}.
\begin{assumption}
    \label{ass:conv}
    In the setting of \cref{lem:spec-g-extr}, the 4-neuron network initialized
    at $\prm^{\varepsilon,*}$ converges in direction to $\prmmm$, in which
    $\forall i, j \: \norm{\vmm_i} = \abs{\umm_i} = \abs{\umm_j}$ and $\forall
    k \: \vmmu_k = b(k) \ee_{a(k)+1}$.
\end{assumption}
We provide experimental and theoretical evidence that this assumption holds in
the next subsection. We also note that this assumption concerns the two-layer
4-neuron network and can be checked directly in an experimental manner. In
contrast, the result below holds for any sufficiently big two-layer neural
networks trained from small initialization.

\begin{lemma}[Proof in \cref{sec:proof-spec-extreme}]
    \label{lem:spec-extreme}
    In the setting of \cref{lem:spec-g-extr} under \cref{ass:conv}, the
    original network $\prm$ converges in direction to $\chi(\prmmm)$ if the
    initialization scale $\sigma$ is small:
    \[
        \exists \sigma^* : \forall \sigma < \sigma^* \lim_{t\to\infty}
        \bigg\lVert \frac{\prm(t)}{\norm{\prm(t)}} -
        \frac{\chi(\prmmm)}{\norm{\chi(\prmmm)}} \bigg\rVert = 0.
    \]
    (Notice $\prm(0) = \sigma \prm^0$ and $\chi$ depends on $\prm^0$ but not on
    $\sigma$.)
\end{lemma}

This lemma shows that the conclusions of Theorems \ref{thm:gen-phase1} and
\ref{thm:gen-phase2} not only propagate to the later stages of training for the
XOR data, but also exacerbate. The network forgets all features except those
learned at the beginning of training, causing an extreme simplicity bias
\citep{s20p}.

\paragraph{Proof sketch} Our proof builds upon concepts studied in
\citealt{l20g}. Following their notation, we define the normalized margin
$\gamma(\prm)$ of $f$ on dataset $D$ by
\[
    \gamma \defeq \min_i f(\prm, \x_i) y_i / \norm{\prm}^2.
\]
Notice that, due to 2-homogeneity of $f$, $\gamma$ depends only on the
direction of $\prm$: $\forall \lambda > 0 \: \gamma(\lambda \prm) =
\gamma(\prm)$. We call a direction that is a (local) solution to the max-margin
problem $\max_{\prm} \gamma(\prm)$ a (local-)max-margin direction. The main
result we build upon in our proof is the following.

\begin{theorem}[Theorem 5.6, \citealt{l21g}]
  \label{thm:spec-extr-bias}
  Consider 2-positively-homogeneous network $f$ trained with
  gradient flow on logistic loss. For any local-max-margin direction,
  $\hat{\prm}^*$, and $\zeta > 0$, $\exists \omega > 0, \rho \ge 1$ such that
  for any $\prm^0$ with $\norm{\prm^0} \ge \rho$ and $\norm{\hat{\prm}^0 -
  \hat{\prm}^*} \le \omega$, gradient flow starting with $\prm^0$ directionally
  converges to some direction $\hat{\prm}$ with the same normalized margin
  $\gamma$ as $\hat{\prm}^*$, and $\norm{\hat{\prm} - \hat{\prm}^*} \le \zeta$.
\end{theorem}

We use this result in the following manner. First, we show that the assumed
limit direction of $\prme$ is a local-max-margin direction. Second, we employ
\cref{thm:gen-phase2} and classical theorems about continuous dependency on
initial conditions for initial value problems to show that our original system
will satisfy the conditions of \cref{thm:spec-extr-bias}. Finally, we apply
\cref{thm:spec-extr-bias} to prove the desired result.

\subsection{Convergence Behavior of Four-Neuron Network}
\label{subsec:evidence_for_assumption}

Finally, we provide evidence for \cref{ass:conv}. Unfortunately, a precise
characterization of this smaller network on
our dataset is challenging. First, as far as we are aware, general results
about convergence to perfect accuracy or zero loss of two-layer networks exist
only for 1-neuron networks \citep{a23a,c23l}. Second, even if one can prove
that the network converges to zero loss, the additional challenge to analyze
the resulting limit direction remains. The only characterization of the limit
directions of networks that we are aware of is in terms of the KKT conditions
for the dual margin-maximization problem \citep{l20g}. However, in the general
case a direct analysis of the KKT conditions remains intractable.

Thus, we give a theoretical motivation for the proposed direction $\prmmm$ and
validate our assumption empirically.

\subsubsection{Theoretical Evidence}

\paragraph{Convergence to perfect accuracy} First, we prove that the network
converges to perfect accuracy.

\begin{figure*}[ht!]
    \centering
    \input{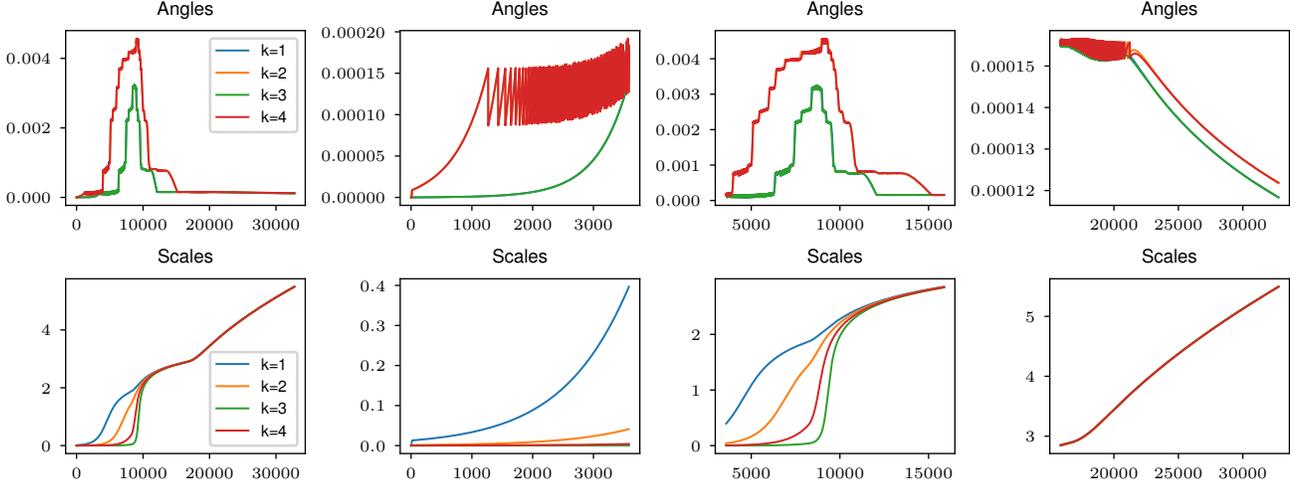}
    \caption{Evolution of 4-neuron network initialized at $(\ue_1(0), \ue_2(0),
    \ue_3(0), \ue_4(0)) = (10^{-4}, -10^{-5}, 10^{-7}, -10^{-6})$. The first
    column depicts the whole training process. We additionally depict different
    stages of training process for visual convenience: the second column
    depicts the first $3584$ training epochs; the third column depicts the
    epochs from $3584$ to $15872$; the last column depicts training after the
    $15872$th epoch. Notice that $\alpha_1 \approx \alpha_3$ and $\alpha_2
    \approx \alpha_4$, where $\alpha_i$ are the angles between the
    network features and the cluster directions.}
    \label{fig:evo}
\end{figure*}

\begin{proposition}[Proof in \cref{sec:proof-spec-conv}]
  \label{prop:spec-conv}
  Assume that $\max_{i, j} \nicefrac{\abs{\ue_i(0)}}{\abs{\ue_j(0)}} \le 1000$,
  $\max_k \abs{\ue_k(0)} \in (0.001, 0.5)$, and $\delta + \xi \le 0.01$. Then
  there exist a time $T^0$ such that $\forall i \:
  f(\prme(T^0), \x_i) y_i > 4.67 > 0$.
\end{proposition}

To comment on the plausibility of the assumptions in this result, first notice
that in the setting of \cref{thm:gen-phase1} neurons evolve almost
independently. Thus, due to the symmetry of our dataset, at the end of Phase 1
the scales $u_j^*$ will be sums of $m$ identically distributed values. Then by
the law of large numbers they will converge to the same values in the limit $m
\to \infty$. Thus, when $m$ is big, at the end of Phase 1, the ratios
$\nicefrac{\abs{\ue_i(0)}}{\abs{\ue_j(0)}}$ will be close to one. Since in
Phase 2 the scales of neurons and the deviation of features is small, we expect
these ratios to increase by no more than $1 + \O(\varepsilon)$, implying that
these ratios again will be close to one.

The assumption that $\max_k \abs{\ue_k(0)} \in (0.001, 0.5)$ is implied
by the property $u^{\varepsilon, *}_j = \Theta(\sqrt{\varepsilon})$. If we
assume that $\varDelta \approx 0.001$ and choose $\varepsilon = \varDelta$, we
can expect that $\prme$ will have the desired scale. Finally, we assume $\delta
+ \xi \le 0.01$ for technical reasons related to the proof technique. We expect
that the property will hold for wider ranges of $\delta$.

\paragraph{Implicit bias at the end of training}  The previous result about
perfect accuracy also suggests that the network may achieve small
loss at the end of training. Then, by Theorem 4.4 of \citet{l20g} and Theorem
3.1 of \citet{j20d}, the network converges to some KKT point of dual
margin-maximization problem 
\[
  \min_{\prm} \norm{\prm}^2 \text{ s.t. } \forall i \: f(\prm, \x_i) y_i \ge 1.
\]
Under the additional assumption that the features of this margin direction are
$(\xi + 2 \varDelta)$-regular for some $\varDelta > 0$, a slight modification
of
the results of \citet{v22o} implies that this direction is a local-max-margin
direction. Finally, we show that $\prmmm$ is one strict local-max-margin
direction, which motivates our hypothesis.

\begin{proposition}[Proof in \cref{sec:proof-spec-loc-max-marg}]
    \label{prop:spec-loc-max-marg}
    Assume the setting of \cref{lem:spec-g-extr}. Then, the direction of
    $\prmmm$ is a strict local-max-margin in the weight space of the $4$-neuron
    network, while its embedding, $\chi(\prmmm)$ is a strict local-max-margin
    in the original weight space.
\end{proposition}

\subsubsection{Empirical Evidence}

Here we present experimental evidence for \cref{ass:conv}. To this end, we
construct a random dataset in $\mathbb{R}^2$ that will
satisfy \cref{ass:xor} with $\delta \le 0.01$ and train a four neuron ReLU
network using gradient descent for different initializations, in which the
neurons are aligned with cluster directions, corresponding to the setting of
\cref{lem:spec-g-extr}. Specifically, after we pick initialization scales
$(\ue_1(0), \ue_2(0), \ue_3(0), \ue_4(0))$, we initialize the first layer as
$\ve_k(0) = \ue_k(0) \ee_{a(k)+1}$ and train our network using plain gradient
descent.

We used the following adaptive learning rate schedule $\eta_t$. During the
first part of training, which corresponds to the evolution initialized at the
limit point from \cref{thm:gen-phase1} to the limit point in
\cref{thm:gen-phase2}, the gradients are very stable. Therefore, we use
constant-scale learning rates $\eta_t = 4$ for $t<12$ and $\eta_t = 2^{-7}$ for
$12 \le t < 2^{13}$. The next part corresponds to the training initialized at
the limit point from \cref{thm:gen-phase2}. At around $t = 2^{13}$ all features
reach constant scales and the cross-entropy loss starts to dump gradients. This
allows us to progressively increase the learning rate, in order to speed up the
simulation, in the third part of schedule, setting $\eta_t = 2^{-7} \Par*{1 +
2^5 \Par*{\frac{t}{2^{13}} - 1}}$ for $2^{13} \le t < 2^{14}$. Finally, at the
end of the third part, the cross-entropy loss causes gradients to decay
exponentially and the training process almost stops. To combat this and
simulate the later stages of training, we use an exponential learning rate
$\eta_t = 2^{-7} \Par[\Big]{1 + 2^{5 + \frac{t - 2^{14}}{2^9}}}$ for $t \ge
2^{14}$.

\cref{fig:evo} depicts the evolution of the 4-neuron network. Here, plots
titled ``Angles'' depict the signed angles between the network features and the
cluster directions. And plots titled ``Scales'' depict $\norm{\ve_k}$. (See
more experiments in \cref{sec:conv-graphs}.) As we can see, eventually the
scales of the neurons became almost identical and the network start to converge
to the desired local-max-margin direction $\prmmm$ (forth column), empirically
supporting our assumption.

%% file: 5b-spec-emp.tex
\section{Empirical Validation of Results}
\label{sec:exper-b}

\begin{figure}[!b]
    \centering
    \includegraphics{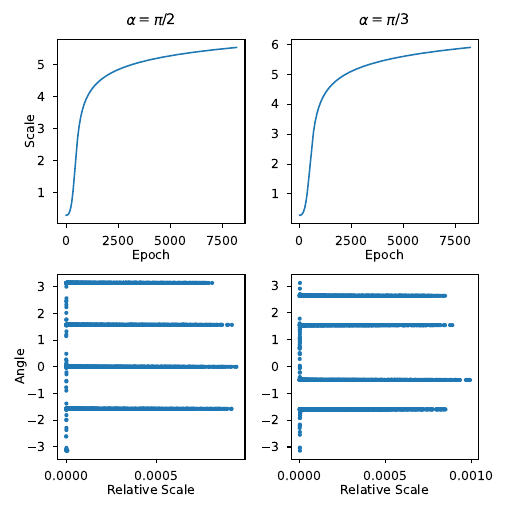}
    \caption{Empirical evidence of simplicity bias on XOR-like data. Relative
    scale in the second row defined as $\nicefrac{\norm{\v_i}^2}{\sum_{j=1}^m
    \norm{\v_j}^2}$.}
    \label{fig:align}
\end{figure}

In this section, we empirically validate the predictions of Theorems
\ref{thm:gen-phase1},
\ref{thm:gen-phase2}, and \cref{lem:spec-extreme}. We consider a skewed
XOR-like dataset in $\mathbb{R}^2$, similar to
the dataset covered by Assumption 5.1, but in which the angle between cluster
directions can be arbitrary. We consider two angles: $\alpha =
\nicefrac{\pi}{2}$ and $\alpha =
\nicefrac{\pi}{3}$. The experiments for $\alpha = \nicefrac{\pi}{2}$ seek to
verify our predictions from Section \ref{sec:spec}, while the experiments for
$\alpha = \nicefrac{\pi}{3}$ test if these predictions transfer to
non-orthogonal cases. We train a two-layer
neural network with $m = 2^{12}$ randomly initialized neurons with
initialization scale $\sigma = 2^{-7}$. We train this network for $2^{13}$
epochs using gradient descent with $\mathtt{lr} = 2^{-4}$. At that stage, we
observed that the network essentially converged.

When $\alpha = \nicefrac{\pi}{2}$, the scale of the first
layer grows from 0.3 to 5.5 during the training (\cref{fig:align}, first row,
left). At the
same time, almost all neurons end up aligned in four directions
(\cref{fig:align}, second row, left): $0, \nicefrac{\pi}{2}, \pi,
-\nicefrac{\pi}{2}$, which are
the global extrema of the function $G$. Similarly, when $\alpha =
\nicefrac{\pi}{3}$ the scale of the first layer grows from 0.3 to 5.9 during
the training
(\cref{fig:align}, first row, right). At the same time, almost all neurons end
up aligned in four directions (\cref{fig:align}, second row, right):
$-\nicefrac{\pi}{6}, \nicefrac{\pi}{2}, \nicefrac{5\pi}{6},
-\nicefrac{\pi}{2}$, which again are the global
extrema of the function $G$. In both cases, the few non-aligned neurons have
smaller
weights than aligned ones (\cref{fig:align}, second row). Therefore, our theory
indeed predicts the
alignment well in this setting.

%% file: 6-exper.tex
\section{Effects of Extreme Simplicity Bias}
\label{sec:exper}

Finally, we test the possible implications of our theoretical characterization
of simplicity bias on real-world datasets. Our previous results suggest that
simplicity bias disproportionally amplifies ``simple'' features that are very
informative about the target. Additionally, in the case of extreme simplicity
bias, if the ``simple'' features are enough to classify the training set
perfectly, the network effectively ``forgets'' all features except for the
``simple'' ones.

The latter fact suggests that if spurious features are enough to classify the
target in the train distribution, then the network should progressively lose
its ability to fine-tune out of distribution. We test this hypothesis on the
MNIST-CIFAR-10 domino dataset proposed by \citet{s20p}.

\begin{figure}[ht!]
  \centering
  \includegraphics[width=0.15\linewidth]{"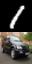"}
  \includegraphics[width=0.15\linewidth]{"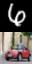"}
  \caption{Examples of domino with a car (class $1$ in CIFAR-10) in train
  (left) and test (right) dataset. Notice that the top MNIST image is an image
  of $1$ only for the train data.}
  \label{fig:exper-mnist-cifar10}
\end{figure}

\begin{figure}[ht!]
  \centering
  \input{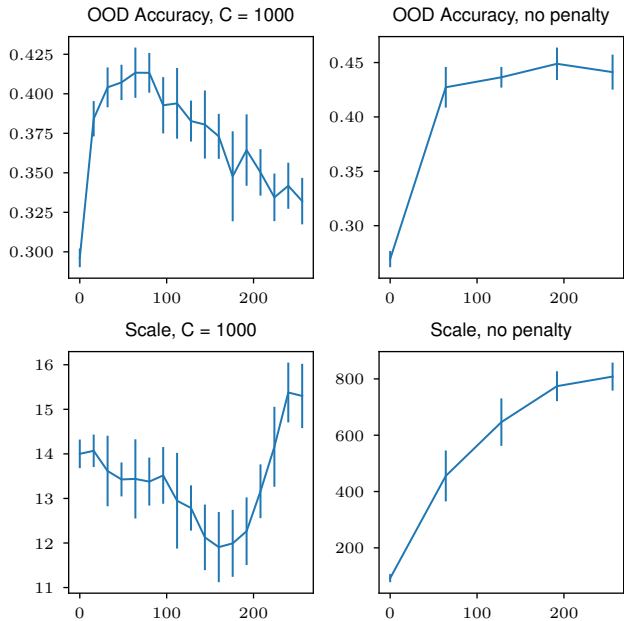}
  \caption{Accuracy and scale of the logistic regression on the validation part
  of the OOD test set ($y$-axis) vs. the training epoch at which the ResNet
  features are extracted ($x$-axis).}
  \label{fig:exper-ood}
\end{figure}

\paragraph{Setup} The dataset contains vertical concatenations of MNIST and
CIFAR-10 images and uses CIFAR-10 image labels. On the train distribution, the
MNIST and CIFAR-10 labels of the images are perfectly correlated: digit $i$
from MNIST will always be concatenated with the image from class $i$ in
CIFAR-10. On the test distribution, MNIST and CIFAR-10 labels are not
correlated, i.e., the concatenations of images from different classes are
random. Thus, MNIST images represent ``simple'' spurious features that can be
used to perfectly classify the data. (See \cref{fig:exper-mnist-cifar10}.)

We fit a ResNet-18 model \citep{h16d} on this domino dataset and track its
parameter trajectory. We use the standard PyTorch initialization,
multiplied by $2^{-5}$ to mimic the small initialization regime studied in the
previous sections. We train this model for $2^8$ epochs using the usual
SGD optimizer with Nesterov momentum and linear schedule with warm-up
\citep[similar training recipe was used by][]{j23d}. Periodically during
training, we apply the current model to our test set and extract its last layer
features, $X$, on test data. We normalize these last layer features (to make
feature scales comparable across different epochs), $X^\text{norm} =
\frac{X}{\sqrt{\frac{1}{n} \sum_{i=1}^n \sum_{j=1}^k X_{ij}^2}},$
where $n$ is the number of test samples, $k$ is the number of features. We then
use these normalized last layer features to train simple logistic regression
model. We train two types of linear models: with small regularization (with
inverse regularization strength equal to $1000$) and without regularization. We
plot the accuracy on the validation part of the test distribution  and the
quadratic mean of the regression coefficients on \cref{fig:exper-ood}. (See
training details in \cref{sec:exper-details}.)

\paragraph{Analysis} First, we can see that the model can not make reliable
predictions on the test set, peaking in accuracy at about $45\%$. This is in
contrast to training the model using the same recipe on plain CIFAR-10 data,
where it achieves around $90\%$ accuracy. At the same time, using the MNIST
labels on the test set, we get around $99\%$ accuracy \citep[see
\cref{fig:ood-rev} in \cref{sec:exper-add}; in line with the results
of][]{s20p,h20w}.
These facts suggest that the network indeed experiences
a simplicity bias and mainly relies on ``simple'' MNIST features for
prediction. Second, as we can see, the OOD accuracy increases fast at the
beginning of training. This may indicate that even simple MNIST features are
better for classification of CIFAR-10 data compared to random features.
Finally, we can see that the OOD accuracy does not increase in the latter
stages of training, indicating that the simplicity bias persists even if we
train longer.  In \cref{sec:exper-add}, we present additional setups, where we
do not scale the initialization (\cref{fig:ood1}) or break the perfect
correlation
between MNIST and CIFAR-10 labels on the train set (\cref{fig:ood-err5}). In
both
cases, we also observe the presence of simplicity bias.

Finally, we can see some evidence in favor of the extreme simplicity bias
mechanism described in \cref{sec:spec}. For the small regularization setup, we observe a significant drop in OOD accuracy when using features from later training stages (the drop
between epochs 64 and 256 is approximately $8.13\% \pm 0.80\%$, which gives
a p-value around $10^{-6}$ according to the t-test). This result suggests a
potential presence of extreme simplicity bias, which impedes the learning of
complex features and makes the network forget the initial random features. For the no regularization setup, we can see that the network does not lose OOD
accuracy, but the last-layer weights become much bigger. This result is again
consistent with the presence of extreme simplicity bias, as our proposed
mechanism does not force the network to forget ``non-simple'' features
directly. Instead, it makes ``simple'' features grow faster, so the
regression can only use ``non-simple'' features by applying
huge weights (approximately $50 \times$ bigger than the regularized
regression).

\paragraph{Discussion} We can draw three practical conclusions from these
experiments. Suppose the training data can be classified using a simple
heuristic. Then, it might be beneficial to train networks in a more ``lazy
regime'' \citep{c19o}, which forces the network to remember randomly
initialized features. Similarly, one could apply early stopping of training. In
this way, one could benefit from new features learned from the training data
without losing access to potentially beneficial random features. Finally, since
simplicity bias works by making ``simple'' features disproportionally large,
the natural countermeasure is to use additional normalization layers, which
have already been proven effective for few-shot transfer \citep[e.g.,][]{p18f}.
At the same time, the ``lazy regime'' or early stopping might harm
in-distribution generalization performance \citep{t23f,l24d}. Thus, a
practical method would need to trade off the benefits of the ``lazy
regime'' or early stopping for the OOD task to their potential
harm to generalization.

%% file: 7-conc.tex
\section{Conclusion}

We characterize simplicity bias beyond linearly separable datasets as the
tendency of features to cluster in several directions, which do not depend on
the network size. We also demonstrate that extreme simplicity bias may appear
for non-linearly-separable datasets and observe it experimentally on image
data. We see our results as an indication that the characterization of
simplicity bias is a crucial step toward improving out-of-distribution
generalization.

%% file: proof-gen-phase1.tex
\section{Proof of Theorem \ref{thm:gen-phase1}}
\label{sec:proof-gen-phase1}

We will prove the theorem in several stages. First, we will analyze
\cref{eq:gen-gf-lin}. Then, we couple Equations (\ref{eq:set-gf}) and
(\ref{eq:gen-gf-lin}). This will allow us to prove the desired result.

\subsection{Analysis of Equation (\ref{eq:gen-gf-lin})}

\subsubsection{Directional Convergence}

Now, we want to apply the results of \citet{j20d} about the directional
convergence to \cref{eq:gen-gf-lin}.

\paragraph{Dynamics of neuron direction} Notice that
\begin{equation}
  \label{eq:gen-direction-dynamics}
  \odv{\vu}{\v} = \odv{\nicefrac{\v}{\norm{\v}}}{\v} = \frac{\norm{\v}^2 -
  \v \v^\tran}{\norm{\v}^3}.
\end{equation}
Thus,
\[
  \odv{\vul_j}{t} = \odv{\vul_j}{\vl_j} \odv{\vl_j}{t} = s_j \P_{\vul}
  \g(\vul_j)
\]
and $\vul(0) = \frac{\v^0}{\norm{\v^0}}$ does not depend on $\sigma$.

Consider the auxiliary system
\begin{equation}
  \label{eq:gen-dir-gf-at-g}
  \odv{\r_j}{t} = s_j \nabla_{\r} G(\ru_j) = \frac{s_j}{\norm{\r_j}}
  \P_{\r_j} \g(\ru_j),
\end{equation}
where $\r_j(0) = \vul_j(0)$. Notice that
\[
  \r_j \odv{\r_j}{t} = 0 \land \norm{\r_j(0)} = 1 \implies \odv{\r_j}{t} = s_j
  \P_{\ru_j} \g(\ru_j).
\]
Thus, $\forall t \: \r_j(t) = \vul_j(t)$: we could use $\r_j$ instead of
$\vul_j$ in further derivations.

We want to show the convergence of $\r_j$ similarly to Theorem 3.1 of
\citet{j20d}. To do it, we want to use Lemma B.11 of \citet{j20d}.

\begin{lemma}[Lemma B.11 of \citealt{j20d}]
  Given a locally Lipschitz definable function $f\colon A \to \mathbb{R}$ with
  an open bounded domain $A$, there exists $\nu > 0$ and a definable
  desingularizing function $\psi$ on $[0, \nu)$ such that
  \[
    \forall \v \in f^{-1}((0, \nu)) \: \psi'(f(\v)) \norm{\nabla_{\v} f(\v)}
    \ge 1.
  \]
\end{lemma}

To apply it, we need to show that $G(\ru)$ is locally Lipschitz and definable.
And since
\[
  G(\vu) = \frac{-\ell'(0)}{n} \sum_{i=1}^n \phi(\vu, \x_i) y_i,
\]
we only need to show the desired properties for $\phi$.

\begin{proposition}
  Denote $c \defeq \max\Par*{-1, \min\Par*{1, \frac{\v^\tran \x}{\xi
  \norm{\v}}}}$ and the class of all polynomials of $x_1, \dots, x_p$ by
  $\mathrm{Poly}(x_1, \dots, x_p)$. We have
  \label{prop:phi-form}
  \[
    \phi(\v, \x) = \int_{-c}^{1} (\v^\tran \x + \xi
    \norm{\v} a) (1 - a^2)^{\frac{d-1}{2}} \frac{\Vol(\D^{d-1})}{\Vol(\D^d)} \d
    a \in \xi \norm{\v} \mathrm{Poly}(c, \sqrt{1 - c^2}).
  \]
\end{proposition}

\begin{proof}
    By the definition
    \[
        \begin{split}
            \phi(\v, \x) = & \int_{\norm{\z} \le 1} (\v^\tran (\x + \xi \z))_+
            \frac{\d \z}{\Vol(\D^d)} = \int_{-1}^1 \int_{\norm{\vec{b}}^2 \le 1
            - a^2} (\v^\tran \x + \xi \norm{\v} a)_+ \frac{\d \vec{b} \d
            a}{\Vol(\D^d)}\\
            = & \int_{-1}^1 (\v^\tran \x + \xi \norm{\v} a)_+ (1 -
            a^2)^{\frac{d-1}{2}} \frac{\Vol(\D^{d-1})}{\Vol(\D^d)} \d a =
            \int_{-c}^1 (\v^\tran \x + \xi \norm{\v} a) (1 -
            a^2)^{\frac{d-1}{2}} \frac{\Vol(\D^{d-1})}{\Vol(\D^d)} \d a\\
            = & \int_{-\arcsin(c)}^{\nicefrac{\uppi}{2}} (\v^\tran \x + \xi
            \norm{\v} \sin(\varphi)) \cos(\varphi)^d
            \frac{\Vol(\D^{d-1})}{\Vol(\D^d)} \d \varphi,
        \end{split}
    \]
    where $a \defeq \vu^\tran \z$ and $\vec{b} \defeq \P_{\v} \z$. To prove that
    the right hand part is a polynomial in $c$ and $\sqrt{1 - c^2}$, notice the
    following identities.
    \[
        \int_{-\arcsin(c)}^{\nicefrac{\uppi}{2}} \sin(\varphi) \cos(\varphi)^d \d
        \varphi = \frac{\cos(\arcsin(c))^{d+1}}{d+1} = \frac{(1 -
        c^2)^{\frac{d+1}{2}}}{d+1}.
    \]
    \[
        \begin{split}
            \int_{-\arcsin(c)}^{\nicefrac{\uppi}{2}} \cos(\varphi)^d \d \varphi
            = & \int_{-\arcsin(c)}^{\nicefrac{\uppi}{2}} \Par*{\frac{\e^{\i
            \varphi} + \e^{-\i \varphi}}{2}}^d \d \varphi =
            \sum_{k=0}^{\frac{d-1}{2}} \binom{d}{k} \frac{\e^{i (d - 2k)
            \varphi} - \e^{-i (d - 2k) \varphi}}{i (d - 2k) 2^d}
            \biggr\rvert_{-\arcsin(c)}^{\nicefrac{\uppi}{2}}\\
            = & \sum_{k=0}^{\frac{d-1}{2}} \binom{d}{k} \frac{\sin((d - 2k)
            \arcsin(c)) + \sin(\frac{(d - 2k) \uppi}{2})}{(d - 2k) 2^{d-1}}.
        \end{split}
    \]

  Finally, notice that
  \[
    \sin(k \arcsin(c)) = \Im(\e^{\i k \arcsin(c)}) = \Im((\sqrt{1 - c^2} + \i
    c)^k) \in \mathrm{Poly}(c, \sqrt{1 - c^2}).
  \]
  Thus, $\phi(\v, \x) \in \xi \norm{\v} \mathrm{Poly}(c, \sqrt{1 - c^2})$.
\end{proof}

\begin{remark}
  Notice that, for even $d$, the activation function will have a term
  proportional to $\arcsin(c)$ in addition to the polynomial in $c$. While
  $\arcsin(c)$ is not definable in the smallest structure on $(\mathbb{R}, +,
  \times)$ (see discussion below), this function is definable on the structure
  $(\mathbb{R}, +, \times, \mathcal{A})$, where $\mathcal{A}$ is the class of
  all restricted functions on $[-1, 1]^n$ \citep[Example 1.5 of][]{l10l}.
\end{remark}

\paragraph{Definability of $\phi$} To prove that $\phi$ is definable in the
o-minimal structure on $(\mathbb{R}, +, \times)$, we will employ the following
properties of definable functions \citep[Appendix B.1,][]{j20d}.
\begin{enumerate}
  \item Let $f, g\colon D \to \mathbb{R}$ be definable functions. Then
    $\forall \alpha, \beta \in \mathbb{R}$ $\alpha f + \beta g$ and $f g$ are
    definable. If $g \neq 0$, $\nicefrac{f}{g}$ is definable. If $f \ge 0$,
    $\forall l \in \mathbb{N} \: \sqrt[l]{f}$ is definable.

  \item Let $f\colon D \to \mathbb{R}^d$. Then $f$ is definable iff all
    coordinate projections of $f$ are definable.

  \item Composition of definable functions is definable.

  \item Any coordinate permutation of a definable set is definable.

  \item The image and pre-image of a definable set by a definable function is
    definable.

  \item Any combination of finitely many definable functions with disjoint
    domains is definable. For example, the point-wise maximum and minimum of
    definable functions are definable.

  \item Polynomial functions are definable.
\end{enumerate}

Now, notice that $\v^\tran \x$ is a linear function, $\norm{\v}$ is a square
root of a quadratic polynomial. Thus, their ratio $\frac{\v^\tran \x}{\xi
\norm{\v}}$ is definable. Therefore, $c$ is definable since it a composition of
the maxima and minima of definable functions. Finally, $\phi(\vu, \x)$ is
definable since it is a polynomial of two definable functions $c$ and $\sqrt{1
- c^2}$.

\paragraph{Local Lipschitzness} To prove that $\phi(\vu, \x)$ is locally
Lipschitz on $A$, we prove a stronger property that $\phi(\v, \x)$ is twice
continuously differentiable on $A$, which would be useful later. Notice that
\[
  \phi(\v, \x) = \int_{-c}^{1} (\v^\tran \x + \xi
  \norm{\v} a) (1 - a^2)^{\frac{d-1}{2}} \frac{\Vol(\D^{d-1})}{\Vol(\D^d)} \d
  a.
\]
Denote $H(z, \v, \x) \defeq \int_{-z}^1 (\v^\tran \x + \xi \norm{\v} a) (1 -
a^2)^{\frac{d-1}{2}} \frac{\Vol(\D^{d-1})}{\Vol(\D^d)} \d a$. Notice that
$\phi(\v, \x) = H(c, \v, \x)$.

By the chain rule for total derivative and the Leibniz integral rule (notice
that $\v^\tran \x + \xi \norm{\v} a$ is smooth for $\v \neq \vec{0}$), when
$\frac{\v^\tran \x}{\xi \norm{\v}} \notin \{-1, 1\}$, we get
\[
    \begin{split}
        \nabla_{\v} \phi(\v, \x) = & \pdv{H}{c} \nabla_{\v} c + \nabla_{\v} H\\
        = & (\v^\tran \x - \xi \norm{\v} c) (1 - c^2)^{\frac{d-1}{2}}
        \frac{\Vol(\D^{d-1})}{\Vol(\D^d)} \nabla_{\v} c + \int_{-c}^{1}
        \Par*{\x + \xi \frac{\v}{\norm{\v}} a} (1 - a^2)^{\frac{d-1}{2}}
        \frac{\Vol(\D^{d-1})}{\Vol(\D^d)} \d a\\
        = & \int_{-c}^{1} \Par*{\x + \xi \frac{\v}{\norm{\v}} a} (1 -
        a^2)^{\frac{d-1}{2}} \frac{\Vol(\D^{d-1})}{\Vol(\D^d)} \d a.
    \end{split}
\]
Thus, $\nabla_{\v} \phi(\v, \x)$ exists and is continuous when $\frac{\v^\tran
\x}{\xi \norm{\v}} \notin \{-1, 1\}$. When $\frac{\v^\tran \x}{\xi \norm{\v}}
\in \{-1, 1\}$ (which implies $c \in \{-1, 1\}$), we could find the derivative
by the definition. Notice that $c$ is locally Lipschitz in $\v$ when $\v \neq
\vec{0}$ since it is a clipping of a continuously differentiable function. Now,
using the Leibniz rule again, we get
\[
    \begin{split}
        \lim_{\d \v \to \vec{0}} \frac{\phi(\v + \d \v, \x) - \phi(\v,
        \x)}{\norm{\d \v}} & \\
        = \lim_{\d \v \to \vec{0}} & \frac{H(c(\v + \d \v), \v + \d \v, \x) -
        H(c(\v + \d \v), \v, \x) + H(c(\v + \d \v, \v, \x) - H(c(\v), \v,
        \x)}{\norm{\d \v}} \\
        = \lim_{\d \v \to \vec{0}} & \int_{-c(\v + \d \v)}^{1} \Par*{\x
        + \xi \frac{\v + \d \v}{\norm{\v + \d \v}} a} (1 - a^2)^{\frac{d-1}{2}}
        \frac{\Vol(\D^{d-1})}{\Vol(\D^d)} \d a + \o(1)\\
        & + \frac{\int_{-c(\v + \d \v)}^{-c(\v)} (\v^\tran \x + \xi \norm{\v}
        a) (1 - a^2)^{\frac{d-1}{2}} \frac{\Vol(\D^{d-1})}{\Vol(\D^d)} \d
        a}{\norm{\d \v}}.
    \end{split}
\]
Since $\abs{c(\v)} = 1$, the integrand in the last expression satisfy
\[
    \begin{split}
        \abs{(\v^\tran \x + \xi \norm{\v} a) (1 - a^2)^{\frac{d-1}{2}}} \le &
        (\abs{\v^\tran \x} + \xi \norm{\v}) (c(\v)^2 - c(\v + \d
        \v)^2)^{\frac{d-1}{2}}\\
        \le & (\abs{\v^\tran \x} + \xi \norm{\v}) 2^{\frac{d-1}{2}} \abs{c(\v)
        - c(\v + \d \v)}^{\frac{d-1}{2}}\\
        = & \O(\norm{\d \v}^{\frac{d-1}{2}}).
    \end{split}
\]
Thus, using Lebesgue's dominated convergence theorem, we get
\[
    \begin{split}
        \lim_{\d \v \to \vec{0}} \frac{\phi(\v + \d \v, \x) - \phi(\v,
        \x)}{\norm{\d \v}}\\
        = & \lim_{\d \v \to \vec{0}} \int_{-c(\v + \d \v)}^{1} \Par*{\x
        + \xi \frac{\v + \d \v}{\norm{\v + \d \v}} a} (1 - a^2)^{\frac{d-1}{2}}
        \frac{\Vol(\D^{d-1})}{\Vol(\D^d)} \d a + \o(1) + \O(\norm{\d
        \v}^{\frac{d-1}{2}})\\
        = & \int_{-c}^{1} \Par*{\x + \xi \frac{\v}{\norm{\v}} a} (1 -
        a^2)^{\frac{d-1}{2}} \frac{\Vol(\D^{d-1})}{\Vol(\D^d)} \d a.
    \end{split}
\]
Therefore, $\phi(\v, \x)$ is continuously differentiable in $\v$ when $\v \neq
\vec{0}$.

Similarly, we get the following Hessian
\[
    \begin{split}
        \nabla_{\v}^2 \phi(\v, \x)\\
        = & \int_{-c}^{1} \frac{\xi a}{\norm{\v}} \P_{\v} (1 -
        a^2)^{\frac{d-1}{2}} \frac{\Vol(\D^{d-1})}{\Vol(\D^d)} \d a + \Par*{\x
        - \xi \frac{\v}{\norm{\v}} c} (1 - c^2)^{\frac{d-1}{2}}
        \frac{\Vol(\D^{d-1})}{\Vol(\D^d)} (\nabla_{\v} c)^\tran\\
        = & \P_{\v} \frac{\xi (1 - c^2)^{\frac{d+1}{2}}}{(d+1) \norm{\v}}
        \frac{\Vol(\D^{d-1})}{\Vol(\D^d)} + \Par*{\x - \xi \frac{\v}{\norm{\v}}
        c} (1 - c^2)^{\frac{d-1}{2}} \frac{\Vol(\D^{d-1})}{\Vol(\D^d)}
        (\nabla_{\v} c)^\tran.
    \end{split}
\]
Notice that when $c \in (-1, 1)$
\[
  \nabla_{\v} c = \P_{\v} \frac{\x}{\xi}, \: \x - \xi \frac{\v}{\norm{\v}} c =
  \P_{\v} \x.
\]
Thus, the Hessian exists, is continuous, and has the form $\nabla^2_{\v}
\phi(\v, \x) = \P_{\v} \mat{A}(\v, \x) \P_{\v}$.

\paragraph{Convergence of $\r_j$} First, notice that $\exists \lim_{t \to
\infty} s_j G(\r_j) \eqdef G_j^*$ because $s_j G(\r_j)$ is monotonically
increasing, $\r_j \in \S^{d-1}$, and $G$ is continuous. Now, consider function
$\zeta_j(t) \defeq \int_0^t \norm*{\odv{\r_j}{t}} \d \tau$, function $f_j(\r_j)
\defeq G_j^* - s_j G(\ru_j)$, and desingularizing function $\psi_j$,
corresponding to $f_j$. Then for big enough $t$ we have
\[
  \odv{f_j}{\zeta_j} = -\norm{\nabla_{\r} G(\ru_j)} = -\norm{\nabla_{\r}
  f_j(\r_j)} \le -\frac{1}{\psi'_j(f_j(\r_j))}
  \implies -\odv{\psi_j}{t} \ge \odv{\zeta_j}{t} \implies \zeta_j(t) \le
  \zeta_j(t_0) + \psi_j(t_0) - \psi_j(t).
\]
Thus, $\zeta_j$ is bounded, and hence $\lim_{t \to \infty} \zeta_j$ exists.
Therefore, the trajectory of $\r_j$ has a finite length, and hence $\lim_{t \to
\infty} \r_j \eqdef \vu^*_j$ exists. Also, since
\[
  \odv{f_j}{t} = -\norm{\nabla_{\r} G(\ru_j)}^2,
\]
$\nabla_{\r} G(\ru)$ is differentiable, and $\ru_j$ converges. Moreover,
$\lim_{t \to \infty} \norm{\nabla_{\r} G(\ru_j)} = 0$ (otherwise $f$ would
increase to infinity). Thus, $\vu^*_j$ is critical: $\P_{\vu^*_j}
\g(\vu^*_j) = \vec{0}$.

\subsubsection{Dynamics around critical directions}

We divide all neurons into two categories: prominent and non-prominent.
Prominent neurons achieve the optimum of function $G$: $s_j G(\vu^*_j)
= \lambda \defeq \max_{\vu \in \S^{d-1}} \abs{G(\vu)}$. Denote $\lambda_j
\defeq s_j G(\vu^*_j)$, $P \defeq \Bc{j \given \lambda_j = \lambda}$ (prominent
neurons), and $R \defeq [m] \setminus P$ (non-prominent neurons).

\paragraph{Dynamics of small neurons} First, we describe the dynamics of
neurons from $R$. We have
\[
  \ul_j = u_j(0) \exp\Par*{\int_0^t s_j G(\vul_j) \d \tau} \le u_j(0)
  \e^{\lambda_j t}.
\]
Notice that the norm growth of these neurons is slower than $\e^{\lambda t}$.

\paragraph{Dynamics of the big neurons' directions} Now, we describe the
dynamics of neurons from $P$. Near critical directions, the Hessian of $G$
restricted on the unit sphere is non-positive if $s_j = 1$ and
non-negative if $s_j = -1$ due to the local characterization of extremum.
Consider the case $s_j = 1$ (the case $s_j = -1$ is similar). Near critical
direction, we have
\[
  (\vu^*_j)^\tran \odv{\vul_j}{t} = (\vu^*_j)^\tran \P_{\vul_j} \g(\vul_j) =
  (\vu^*_j)^\tran \P_{\vul_j} (\g(\vu^*_j) + \nabla^2 G(\vu^*_j) (\vul_j -
  \vu^*_j) + \O(\norm{\vul_j - \vu^*_j}^2))
\]
Denote $\vul_j = \cos(\alpha) \vu^*_j + \sin(\alpha) \veps$, where
$\cos(\alpha) = (\vu^*_j)^\tran \vul_j$ and $\veps \defeq \frac{\P_{\vu^*_j}
\vul_j}{\norm{\P_{\vu^*_j} \vul_j}}$. We get
\[
  \P_{\vul_j} \vu^*_j = (1 - \cos(\alpha)^2) \vu^*_j - \sin(\alpha)
  \cos(\alpha) \veps.
\]
Since $G$ is homogeneous, we get
\[
  \g(\vu^*_j) = \lambda_j \vu^*_j.
\]
Also, we know that
\[
  \nabla^2 G(\vu) = \P_{\vu} \nabla^2 G(\vu) \P_{\vu} \implies \nabla^2
  G(\vu^*_j) (\vul_j - \vu^*_j) = \nabla^2 G(\vu^*_j) \sin(\alpha) \veps.
\]
These equations give
\[
    \begin{split}
        (\vu^*_j)^\tran \odv{\vul_j}{t} = & ((1 - \cos(\alpha)^2) \vu^*_j -
        \sin(\alpha) \cos(\alpha) \veps)^\tran (\lambda \vu^*_j + \nabla^2
        G(\vu^*_j) \sin(\alpha) \veps + \O(\alpha^2))\\
        = & \lambda (1 - \cos(\alpha)^2) - \veps^\tran \nabla^2 G(\vu^*_j)
        \veps \sin(\alpha)^2 \cos(\alpha) + \O(\alpha^3)\\
        \ge & \lambda \norm{\vu^*_j - \vul_j}^2 + \O(\alpha^3).
    \end{split}
\]
This equation implies
\[
    \begin{split}
        & \odv{\norm{\vul_j - \vu^*_j}^2}{t} \le -2 \lambda \norm{\vul_j -
        \vu^*_j}^2 + 2 \delta \norm{\vul_j - \vu^*_j}^3
        \implies \frac{\norm{\vul_j - \vu^*_j}}{1 - \frac{\delta}{\lambda}
        \norm{\vul_j - \vu^*_j}} \le \frac{\norm{\vul_j(t_{0, j}) -
        \vu^*_j(t_{0, j})} \e^{-\lambda (t - t_{0, j})}}{1 -
        \frac{\delta}{\lambda} \norm{\vul_j(t_{0, j}) - \vu^*_j(t_{0, j})}}\\
        \implies & \norm{\vul_j - \vu^*_j} \le \frac{\norm{\vul_j(t_{0, j}) -
        \vu^*_j(t_{0, j})} \e^{-\lambda (t - t_{0, j})}}{1 -
        \frac{\delta}{\lambda} \norm{\vul_j(t_{0, j}) - \vu^*_j(t_{0, j})} (1 -
        \e^{-\lambda (t - t_{0, j})})},
    \end{split}
\]
where $2 \delta \norm{\vul_j - \vu^*_j}^3$ comes from the term $\O(\alpha^3)$
and $t_{0, j}$ is chosen such that $\norm{\vul_j - \vu^*_j}$ is sufficiently
small to bound our big-O term and, at the same time, $\frac{\delta}{\lambda}
\norm{\vul_j(t_0) - \vu^*_j(t_0)} \le \frac{1}{2}$. Thus, we get an
exponentially fast convergence near the critical direction.

Denote
\[
  t_0 \defeq \max_j t_{0, j}, \: c_0 \defeq \max_j 2 \norm{\vul_j(t_{0, j}) -
  \vu^*_j(t_{0, j})} \e^{\lambda t_{0, j}}.
\]
We get
\[
  \forall j \in P, t \ge t_0 \: \norm{\vul_j - \vu_j} \le c_0 \e^{-\lambda t}.
\]

\paragraph{Dynamics of the big neurons' scales} Now, we will describe the
dynamics of $\ul_j$. Notice
\[
  \odv{\ul_j}{t} = \ul_j G(\vul_j) \implies \ul_j = \ul_j(0) \exp\Par*{\int_0^t
  G(\vul_j) \d \tau}.
\]
This equality motivates us to consider a limit
\[
  u^*_j \defeq \lim_{t \to \infty} \ul_j \e^{-\lambda t}.
\]
This limit exists since the right hand part is monotonically decreasing. We get
\[
  \ul_j = \ul_j(0) \exp\Par*{\int_0^t G(\vul_j) \d \tau} = u^*_j
  \exp\Par*{\lambda t + \int_t^\infty (\lambda - G(\vul_j)) \d \tau}.
\]
Thus,
\[
  u^*_j \e^{\lambda t} \le \ul_j \le u^*_j \exp\Par*{\lambda t +
  \int_t^\infty c_0 b_0 \e^{-\lambda \tau} \d \tau} = u^*_j \exp\Par*{\lambda t
  + \frac{c_0 b_0}{\lambda} \e^{-\lambda t}},
\]
where $b_0 \defeq \sup_{\v \neq \v'} \frac{\abs{G(\v) - G(\v')}}{\norm{\v -
\v'}}$. Notice that this derivation also implies that neurons in $P$ grow as
$\Theta(\e^{\lambda t})$.

\begin{remark}
  It is easy to see that $b_0 \le (-\ell'(0)) (1 + \xi)$.
\end{remark}

\paragraph{Neuron capture} Notice that all global extrema are attractive.
Therefore, for isotropic initialization, with probability at least $1 -
(f_j)^m$, some point will be attracted to the global extremum $\vu^*_j$, where
$f_j \defeq 1 - \frac{\text{volume of attraction region}}{\text{volume of
shpere}}$.

\subsection{Coupling Equations (\ref{eq:set-gf}) and (\ref{eq:gen-gf-lin})}

\subsubsection{Norm Growth Rate}

First, we prove the following proposition and lemma.

\begin{proposition}
  Assuming $\norm{\x_i} \le 1$, the following identities hold
  \[
    \begin{split}
      \abs*{\frac{1}{n} \sum_{i=1}^n (-\ell'(f(\prm, \x_i) y_i) + \ell'(0))
      \phi(\vu, \x_i) y_i} & \le a \norm{\prm}_{[m]}^2,\\
      \abs*{\frac{1}{n} \sum_{i=1}^n (-\ell'(f(\prm, \x_i) y_i) + \ell'(0))
      \nabla_{\v} \phi(\z, \x_i)\rvert_{\z = \vu} y_i} & \le a
      \norm{\prm}_{[m]}^2,
    \end{split}
  \]
  where $a \defeq \frac{m (1 + \xi)^2}{4}$.
\end{proposition}

\begin{proof}
  Notice that
  \[
    \abs{f(\prm, \x_i)} \le \sum_{j=1}^m u_j^2 \abs{\phi(\vu_j, \x_i)} \le
    (1 + \xi) \sum_{j=1}^m u_j^2 \le m (1 + \xi) \norm{\prm}_{[m]}^2,
  \]
  and
  \[
    \abs{-\ell'(f(\prm, \x_i) y_i) + \ell'(0)} \le \abs{f(\prm, \x_i)} \sup_{z}
    \abs{\ell''(z)} \le \frac{m (1 + \xi)}{4} \norm{\prm}_{[m]}^2.
  \]
  Thus,
  \[
    \abs*{\frac{1}{n} \sum_{i=1}^n (-\ell'(f(\prm, \x_i) y_i) + \ell'(0))
    \phi(\vu, \x_i) y_i} \le \frac{m (1 + \xi)}{4} \norm{\prm}_{[m]}^2 \max_{i,
    \vu} \abs{\phi(\vu, \x_i)} \le a \norm{\prm}_{[m]}^2.
  \]
  Similarly for the gradients of activation.
\end{proof}

\begin{lemma}
  \label{lem:prm-ub}
  Assume that $\prm$ follows \cref{eq:set-gf} and $\abs{u_j(0)} =
  \norm{\v_j(0)}$. Then
  \[
    \forall t \le t_1 \: \norm{\prm}_{[m]}^2 \le 2 \norm{\prm(0)}_{[m]}^2 \e^{2
    \lambda t},
  \]
  where $t_1 \defeq \frac{1}{2 \lambda} \ln\Par*{\frac{\lambda}{2 a
  \norm{\prm(0)}_{[m]}^2}}$.
\end{lemma}

\begin{proof}
    Thus,
    \[
        \begin{split}
            &\odv{u_j}{t} = G(\v_j) + \frac{1}{n} \sum_{i=1}^n (p_i(\prm) +
            \ell'(0)) \phi(\v_j, \x_i) y_i \implies \abs*{\odv{u_j}{t}} \le
            \abs{u_j} \abs{G(\vu_j)} + a \norm{\prm}_{[m]}^2 \abs{u_j} \le
            \lambda \norm{\prm}_{[m]} + a \norm{\prm}_{[m]}^3\\
            &\implies \int \frac{\d \norm{\prm}_{[m]}^2}{\norm{\prm}_{[m]}^2 +
            \frac{a}{\lambda} \norm{\prm}_{[m]}^4} \le 2 \lambda t \implies
            \norm{\prm}_{[m]}^2 \le \frac{\norm{\prm(0)}_{[m]}^2 \e^{2 \lambda
            t}}{1 - \frac{a}{\lambda} \norm{\prm(0)}_{[m]}^2 (\e^{2 \lambda t}
            - 1)}.
        \end{split}
    \]
    Therefore,
    \[
        \norm{\prm}_{[m]}^2 \le 2 \norm{\prm(0)}_{[m]}^2 \e^{2 \lambda t} \:
        \forall t \le \frac{1}{2 \lambda} \ln\Par*{\frac{\lambda}{2 a
        \norm{\prm(0)}_{[m]}^2}}.
    \]
\end{proof}

\subsubsection{Coupling Directions}

We have
\[
  \begin{split}
    \odv{\vu_j}{t} &= s_j \P_{\v} \Par*{\g(\vu_j) + \frac{1}{n} \sum_{i=1}^n
    (-\ell'(f(\prm, \x_i) y_i) + \ell'(0)) \nabla_{\v} \phi(\v_j, \x_i) y_i},\\
    \odv{\vul_j}{t} &= s_j \P_{\vl} \g(\vul_j).
  \end{split}
\]

Denote $b_1 \defeq \sup_{\vu, \vu' \in \S^{d-1}, \vu \neq \vu'}
\frac{\norm{\P_{\vu} \g(\vu) - \P_{\vu'} \g(\vu')}}{\norm{\vu - \vu'}}$ (since
$\phi$ is twice continuously differentiable, this constant is defined).

\begin{remark}
  It is easy to see that $b_1 \le 2 \sup_{\vu} \norm{\g(\vu)} +
  (-\ell'(0)) \sup_{i, \vu} \norm{\nabla_{\v}^2 \phi(\v, \x_i)} \implies b_1 =
  \O\Par*{\nicefrac{1}{\xi}}$, when $\xi \to 0$.
\end{remark}

We get
\[
    \begin{split}
        \odv{\norm{\vu_j - \vul_j}}{t} \le & \norm*{\odv{\vu_j}{t} -
        \odv{\vul_j}{t}} \le \norm*{\frac{1}{n} \sum_{i=1}^n (-\ell'(f(\prm,
        \x_i) y_i) + \ell'(0)) \nabla_{\v} \phi(\v_j, \x_i) y_i} +
        \norm{\P_{\vu} \g(\vu) - \P_{\vul} \g(\vul)}\\
        \le & a \norm{\prm}_{[m]}^2 + b_1 \norm{\vu_j - \vul_j}.
    \end{split}
\]
Consider function $h \defeq \norm{\vu_j - \vul_j} \e^{-b_1 t}$, we get
\[
    \begin{split}
        & \forall t \le t_1 \: \odv{h}{t} \le a \norm{\prm}_{[m]}^2 \e^{-b_1 t}
        \implies \forall t \le t_1 \: h \le \int_0^t 2 a \norm{\prm(0)}_{[m]}^2
        \e^{(2 \lambda - b_1) \tau} \d \tau\\
        \implies & \forall t \le t_1 \: h \le
        \frac{2 a}{b_1 - 2 \lambda} \norm{\prm(0)}_{[m]}^2 (1 - \e^{(2 \lambda
        - b_1) t})
        \implies \forall t \le t_1 \: \norm{\vu_j - \vul_j} \le \frac{2
        a}{b_1 - 2 \lambda} \norm{\prm(0)}_{[m]}^2 \e^{b_1 t},
    \end{split}
\]
where we have assumed that $b_1 > 2 \lambda$ (which holds for small enough
$\xi$).

\subsubsection{Coupling Scales}

We have
\[
  \begin{split}
    \odv{u_j}{t} &= G(\v_j) + \frac{1}{n} \sum_{i=1}^n (-\ell'(f(\prm, \x_i)
    y_i) + \ell'(0)) \phi(\v_j, \x_i) y_i,\\
    \odv{\ul_j}{t} &= G(\vl_j).
  \end{split}
\]
It gives
\[
    \begin{split}
        \forall t \le t_1 \: \odv{\abs{u_j - \ul_j}}{t} \le &
        \abs*{\odv{u_j}{t} - \odv{\ul_j}{t}} \le a \norm{\prm}_{[m]}^2
        \abs{u_j} + \abs{\ul_j} \abs{G(\vu_j) - G(\vul_j)} + \abs{u_j - \ul_j}
        \abs{G(\vu_j)}\\
        \le & 2 \sqrt{2} a \norm{\prm(0)}_{[m]}^3 \e^{3 \lambda t} + \frac{2 a
        b_0 \norm{\prm(0)}_{[m]}^3}{b_1 - 2 \lambda} \e^{(\lambda_j + b_1) t} +
        \lambda \abs{u_j - \ul_j}.
    \end{split}
\]
As previously, consider function $h \defeq \abs{u_j - \ul_j} \e^{-\lambda t}$,
we get
\[
    \begin{split}
        & \forall t \le t_1 \: h \le 2 \sqrt{2} a \norm{\prm(0)}_{[m]}^3 \e^{2
        \lambda t} + \frac{2 a b_0 \norm{\prm(0)}_{[m]}^3}{b_1 - 2 \lambda}
        \e^{(\lambda_j + b_1 - \lambda) t}\\
        \implies & \forall t \le t_1 \: \abs{u_j - \ul_j} \le \frac{a
        \sqrt{2}}{\lambda} \norm{\prm(0)}_{[m]}^3 (\e^{3 \lambda t} -
        \e^{\lambda t}) + \frac{2 a b_0}{(b_1 - 2 \lambda) (\lambda_j + b_1 -
        \lambda)} \norm{\prm(0)}_{[m]}^3 (\e^{(\lambda_j + b_1) t} -
        \e^{\lambda t})\\
        \implies & \forall t \le t_1 \: \abs{u_j - \ul_j} \le c_1
        \norm{\prm(0)}_{[m]}^3 \e^{(\lambda_j + b_1) t},
    \end{split}
\]
where we have assumed that $b_1 \ge 4 \lambda$ and denoted $c_1 \defeq \frac{a
\sqrt{2}}{\lambda} + \frac{2 a b_0}{(b_1 - 2 \lambda)^2}$.

\subsection{Final Bound}

The result of the previous sections show
\[
    \begin{aligned}
        \forall j \in P, t \in [t_0, t_1] & & \norm{\vu_j - \vu^*_j} & \le
        \norm{\vu_j - \vul_j} + \norm{\vul_j - \vu^*_j} \le \frac{2 a}{b_1 - 2
        \lambda} \norm{\prm(0)}_{[m]}^2 \e^{c^{\g} t} + c_0 \e^{-\lambda t},\\
        \forall j \in P, t \in [t_0, t_1] & & \abs{u_j - u^*_j \e^{\lambda t}}
        & \le \abs{u_j - \ul_j} + \abs{\ul_j - u^*_j \e^{\lambda t}}\\
        & & & \le c_1 \norm{\prm(0)}_{[m]}^3 \e^{(\lambda + b_1) t} + \abs{u^*_j}
        \e^{\lambda t} \Par*{\exp\Par*{\frac{c_0 b_0}{\lambda} \e^{-\lambda t}} -
        1},\\
        \forall j \in R, t \in [0, t_1] & & \abs{u_j} & \le c_1
        \norm{\prm(0)}_{[m]}^3 \e^{(\lambda_j + c^{\g}) t} + \abs{u_j(0)}
        \e^{\lambda_j t}.
    \end{aligned}
\]

We want to control these errors until the point when the fastest growing
directions will have a predefined scale, $r$. To do this, we will choose $T_1 =
\frac{1}{\lambda} \ln\Par*{\frac{r}{\sigma}}$ and $\sigma = r^{\kappa + 1}$.
(Notice that $t_0$ does not depend on $r$, and $t_1 =
-\nicefrac{\ln(\sigma)}{\lambda} + \O(1)$. Thus, for sufficiently small $r$, we
have $t_0 < T_1 < t_1$.)

This functional form will give us the following errors:
\[
    \begin{aligned}
        \forall j \in P & & \norm{\vu_j - \vu^*_j} \le & \O(r^{2 - \kappa
        \Par*{\nicefrac{b_1}{\lambda} - 2}} + r^\kappa),\\
        \forall j \in P & & \abs{u_j - u^*_j \e^{\lambda T_1}} \le & \O(r^{3 -
        \kappa \Par*{\nicefrac{b_1}{\lambda} - 2}} + r^{1 + \kappa}),\\
        \forall j \in R & & \abs{u_j} \le & \O\Par*{r^{3 - \kappa
        \Par[\big]{\frac{b_1 + \lambda_j}{\lambda} - 3}} + r^{1 + \kappa
        \Par[\big]{1 - \frac{\lambda_j}{\lambda}}}}.
    \end{aligned}
\]
We choose $\kappa^* = \frac{2}{\frac{b_1}{\lambda} - 1}$. It will give
\[
    \begin{aligned}
        \forall j \in P & & \norm{\vu_j - \vu^*_j} \le & \O(r^{\kappa^*}),\\
        \forall j \in P & & \abs{u_j - u^*_j \e^{\lambda T_1}} \le & \O(r^{1 +
        \kappa^*}),\\
        \forall j \in R & & \abs{u_j} \le & \O(r^{1 + \kappa_j}),
    \end{aligned}
\]
where $\kappa_j \defeq \kappa^* (1 - \nicefrac{\lambda_j}{\lambda})$.

%% file: proof-gen-phase2.tex
\section{Proof of Theorem \ref{thm:gen-phase2}}
\label{sec:proof-gen-phase2}

By Lemma 5.3 of \citet{l21g}, we could write the following dynamics for
embedding (\ref{eq:gen-emb})
\begin{equation}
  \label{eq:gen-gf-at-g}
  \begin{split}
    \odv{\ut_j}{t} &= \frac{1}{n} \sum_{i=1}^n (-\ell'(f(\prmt, \x_i)
    y_i)) \phi(\vt_j, \x_i) y_i,\\
    \odv{\vt_j}{t} &= \frac{\ut_j}{n} \sum_{i=1}^n (-\ell'(f(\prmt, \x_i)
    y_i)) \nabla_{\v} \phi(\vt_j, \x_i) y_i.
  \end{split}
\end{equation}
Denote $\prm^* \defeq \prmt(T_1) / r$.

We will prove the theorem in two stages. First, we will couple Equations
(\ref{eq:set-gf}) and (\ref{eq:gen-gf-at-g}). Then we will investigate the
sensitivity of \cref{eq:gen-gf-at-g} to initialization scale. This analysis
will allow us to analyze the solution to \cref{eq:set-gf}.

\subsection{Coupling Equations (\ref{eq:set-gf}) and (\ref{eq:gen-gf-at-g})}

\subsubsection{Norm Growth}

Since $\norm{\prmt(T_1)}_{[m]} = \Theta(r)$ and $\norm{\prmt(T_1)
- \prm(T_1)}_{[m]} = \o(r)$, we have that $\norm{\prm(T_1)} \le
\sqrt{2} \norm{\prmt(T_1)}$ for sufficiently small $r$. Using this fact and
\cref{lem:prm-ub}, we get
\[
    \begin{aligned}
        \forall t \in [T_1, T_1 + t_3] & & \norm{\prm(T_1)}_{[m]}^2 \le & 2 q^2
        \e^{2 \lambda (t - T_1)},\\
        \forall t \in [T_1, T_1 + t_3] & & \norm{\prmt(T_1)}_{[m]}^2 \le & 2
        q^2 \e^{2 \lambda (t - T_1)},\\
  \end{aligned}
\]
where $t_3 \defeq \frac{1}{2 \lambda} \ln\Par*{\frac{\lambda}{2 a q^2}}$ and
$q^2 \defeq 2 \norm{\prmt(T_1)}_{[m]}^2 = 2 r^2 \norm{\prm^*}_{[m]}^2$.

\subsubsection{Bounding \texorpdfstring{$\norm{\vtu_j - \vu^*_j}$}{||\~v\_j
- v\_j||}}

W.l.o.g. assume that $s_j = 1$. Additionally, assume $\norm{\vtu_j - \vu^*_j}
\le \varepsilon$ and denote $\vtu_j = \cos(\alpha') \vu^*_j +
\sin(\alpha') \veps$. First, notice that $\norm{\vtu_j - \vu^*_j} \le
\varepsilon \le \varDelta$ implies
\[
  \phi(\vtu_j, \x_i) = ((\vtu_j)^\tran \x_i)_+ \implies \g(\vtu_j) =
  \g(\vu^*_j).
\]
We have
\[
    \begin{split}
        \forall j \in P \: (\vu^*_j)^\tran \odv{\vtu_j}{t} = & (\P_{\vtu_j}
        \vu^*_j)^\tran \Par*{\g(\vtu_j) + \frac{1}{n} \sum_{i=1}^n
        (-\ell'(f(\prmt, \x_i) y_i) + \ell'(0)) \nabla_{\v} \phi(\vt_j, \x_i)
        y_i}\\
        \ge & ((1 - \cos(\alpha')^2) \vu^*_j - \sin(\alpha') \cos(\alpha')
        \veps)^\tran \g(\vtu_j) - \sin(\alpha') a \norm{\prmt}_{[m]}^2\\
        = & \lambda \sin(\alpha')^2 - \sin(\alpha') a \norm{\prmt}_{[m]}^2
    \end{split}
\]
where $\z \in [\vu^*_j, \vtu_j]$. It implies
\[
  \forall t \in [T_1, T_1 + t_3] \: \odv{\alpha'}{t} \le 2 a q^2 \e^{2 \lambda
  (t - T_1)} - \lambda \sin(\alpha') \implies \alpha' \le \frac{a}{\lambda} q^2
  (\e^{2 \lambda (t - T_1)} - 1).
\]
So, to ensure $\norm{\vtu_j - \vu^*_j} \le \varepsilon$, it is
sufficient to have
\[
  \frac{a}{\lambda} q^2 \e^{2 \lambda (t - T_1)} \le \varepsilon
  \impliedby t - T_1 \le \frac{1}{2 \lambda} \ln\Par*{\frac{\lambda
  \varepsilon}{a q^2}} \eqdef t^\varepsilon_4.
\]

\subsubsection{Bounding \texorpdfstring{$\norm{\prm - \prmt}$}{||theta -
\~theta||}}

\paragraph{Coupling Directions} First, we want to bound $\norm{\vu_j -
\vtu_j}$. Assuming that $\norm{\vu_j - \vtu_j} \le \varepsilon$, we get
$\norm{\vu_j - \vu^*_j} \le 2 \varepsilon$ and
\[
    \begin{split}
        \odv{\norm{\vu_j - \vtu_j}^2}{t} = & -\vu_j^\tran \odv{\vtu_j}{t} -
        (\vtu_j)^\tran \odv{\vu_j}{t}\\
        = & -(\P_{\vtu_j} \vu_j)^\tran \Par*{\g(\vtu_j) + \frac{1}{n}
        \sum_{i=1}^n (-\ell'(f(\prmt, \x_i) y_i) + \ell'(0)) \nabla_{\v}
        \phi(\vt_j, \x_i) y_i}\\
        & - (\P_{\vu_j} \vtu_j)^\tran \Par*{\g(\vu_j) + \frac{1}{n}
        \sum_{i=1}^n (-\ell'(f(\prm, \x_i) y_i) + \ell'(0)) \nabla_{\v}
        \phi(\v_j, \x_i) y_i}\\
        = & -(\P_{\vtu_j} \vu_j)^\tran (\g(\vtu_j) - \g(\vu_j))\\
        & - (\P_{\vtu_j} \vu_j)^\tran \Par*{\frac{1}{n} \sum_{i=1}^n
        (-\ell'(f(\prmt, \x_i) y_i) + \ell'(0)) (\nabla_{\v} \phi(\vt_j, \x_i)
        - \nabla_{\v} \phi(\v_j, \x_i)) y_i}\\
        & - (\P_{\vtu_j} \vu_j)^\tran \Par*{\frac{1}{n} \sum_{i=1}^n
        (-\ell'(f(\prmt, \x_i) y_i) + \ell'(f(\prm, \x_i) y_i)) \nabla_{\v}
        \phi(\v_j, \x_i) y_i}\\
        & - (\P_{\vtu_j} \vu_j + \P_{\vu_j} \vtu_j)^\tran \Par*{\g(\vu_j) +
        \frac{1}{n} \sum_{i=1}^n (-\ell'(f(\prm, \x_i) y_i) + \ell'(0))
        \nabla_{\v} \phi(\v_j, \x_i) y_i}\\
        = & - (\P_{\vtu_j} \vu_j)^\tran \Par*{\frac{1}{n} \sum_{i=1}^n
        (-\ell'(f(\prmt, \x_i) y_i) + \ell'(f(\prm, \x_i) y_i)) \nabla_{\v}
        \phi(\v_j, \x_i) y_i}\\
        & - (\P_{\vtu_j} \vu_j + \P_{\vu_j} \vtu_j)^\tran \Par*{\g(\vu_j) +
        \frac{1}{n} \sum_{i=1}^n (-\ell'(f(\prm, \x_i) y_i) + \ell'(0))
        \nabla_{\v} \phi(\v_j, \x_i) y_i}.
    \end{split}
\]
Denote $\cos(\beta) \defeq \vu_j^\tran \vtu_j$. We get
\[
  \norm{\P_{\vu_j} \vtu_j}^2 = \norm{\vtu_j - \cos(\beta) \vu_j}^2 = 1 -
  \cos(\beta)^2 \le \norm{\vu_j - \vtu_j}^2.
\]
Now, notice
\[
    \begin{split}
        \abs{f(\prm, \x_i) - f(\prmt, \x_i)} = & \abs*{\sum_{j=1}^m (u_j^2
        \phi(\vu_j, \x_i) - (\ut_j)^2 \phi(\vtu_j, \x_i))}\\
        = & \abs*{\sum_{j=1}^m ((u_j^2 - (\ut_j)^2) \phi(\vu_j, \x_i) +
        (\ut_j)^2 (\phi(\vu_j, \x_i) - \phi(\vtu_j, \x_i)))}\\
        = & \abs*{\sum_{j=1}^m ((u_j^2 - (\ut_j)^2) \phi(\vu_j, \x_i) +
        (\ut_j)^2 [(\vu^*_j)^\tran \x_i \ge 0] (\vu_j - \vtu_j)^\tran \x_i)}\\
        \le & \abs{P} (1 + \xi) (\norm{\prm}_P + \norm{\prmt}_P) U + \abs{R} (1
        + \xi) \norm{\prm}_R^2 + \abs{P} \norm{\prm}_{[m]}^2 V,
    \end{split}
\]
where $U \defeq \max_{j \in P} \abs{u_j - \ut_j}$, $V \defeq \max_{j \in P}
\norm{\vu_j - \vtu_j}$. This formula implies
\begin{multline*}
  \norm*{(\P_{\vtu_j} \vu_j)^\tran \Par*{\frac{1}{n} \sum_{i=1}^n
  (-\ell'(f(\prmt, \x_i) y_i) + \ell'(f(\prm, \x_i) y_i)) \nabla_{\v}
  \phi(\v_j, \x_i) y_i}}\\
  \le a U (\norm{\prm}_{[m]} + \norm{\prmt}_{[m]}) \norm{\vu_j - \vtu_j} + a
  \norm{\prm}_{R}^2 \norm{\vu_j - \vtu_j} + a \norm{\prmt}_{[m]}^2 V
  \norm{\vu_j - \vtu_j}.
\end{multline*}
Finally,
\[
  \P_{\vtu_j} \vu_j + \P_{\vu_j} \vtu_j = (1 - \cos(\beta)) (\vu_j + \vtu_j) =
  \norm{\vu_j - \vtu_j}^2 \frac{\vu_j + \vtu_j}{2},
\]
which gives
\begin{multline*}
    (\P_{\vtu_j} \vu_j + \P_{\vu_j} \vtu_j)^\tran \Par*{\g(\vu_j) + \frac{1}{n}
    \sum_{i=1}^n (-\ell'(f(\prm, \x_i) y_i) + \ell'(0)) \nabla_{\v} \phi(\v_j,
    \x_i) y_i}\\
    \ge \lambda \norm{\vu_j - \vtu_j}^2 \frac{\cos(\alpha) + \cos(\alpha')}{2}
    - a \norm{\prm}_{[m]}^2 \norm{\vu_j - \vtu_j}^2,
\end{multline*}
where $\cos(\alpha) = \vu_j^\tran \vu^*_j$. Together, we get
\[
    \begin{alignedat}{2}
        & & \odv{V^2}{t} \le & a (\norm{\prm}_{[m]}^2 + \norm{\prmt}_{[m]}^2)
        V^2 + a \norm{\prm}_R^2 V + a (\norm{\prm}_{[m]} + \norm{\prmt}_{[m]})
        U V\\
        & & \le & 4 a q^2 \e^{2 \lambda (t - T_1)} V^2 + a \norm{\prm}_R^2 V + 2
        \sqrt{2} a q \e^{\lambda (t - T_1)} U V\\
        \implies & & \odv{V}{t} \le & 2 a q^2 \e^{2 \lambda (t - T_1)} V +
        \frac{a}{2} \norm{\prm}_R^2 + \sqrt{2} a q \e^{\lambda (t - T_1)} U.
    \end{alignedat}
\]

\paragraph{Bounding $\norm{\prm}_R$} We get
\[
  \odv{u_j}{t} = G(\v_j) + \frac{1}{n} \sum_{i=1}^n (-\ell'(f(\prm, \x_i) y_i)
  + \ell'(0)) \phi(\v_j, \x_i) y_i \implies \abs*{\odv{u_j}{t}} \le \abs{u_j}
  (\lambda + a \norm{\prm}_{[m]}^2).
\]
Thus,
\[
  \abs{u_j} \le \abs{u_j(T_1)} \exp\Par*{\int_{T_1}^t \lambda + a
  \norm{\prm}_{[m]}^2 \d \tau} \le \abs{u_j(T_1)} \exp\Par*{\lambda (t -
  T_1) + \frac{a}{\lambda} q^2 \e^{2 \lambda (t - T_1)}}.
\]
If $t - T_1 \le t^\varepsilon_4$, we get
\[
  \abs{u_j} \le \e^{\lambda (t - T_1) + \varepsilon} \abs{u_j(T_1)}
  \implies \norm{\prm}_R \le \e^{\lambda (t - T_1) + \varepsilon}
  \norm{\prm(T_1)}_R.
\]

\paragraph{Bounding $U$} Similarly to the previous cases, we have
\begin{multline*}
  \odv{\abs{u_j - \ut_j}}{t} \le \abs*{\odv{u_j}{t} - \odv{\ut_j}{t}}\\
    \le \abs*{u_j G(\vu_j) - \ut_j G(\vtu_j)}
  + \abs*{\frac{1}{n} \sum_{i=1}^n ((\ell'(0) - \ell'(f(\prm, \x_i) y_i))
  \phi(\v_j, \x_i) - (\ell'(0) -\ell'(f(\prmt, \x_i) y_i)) \phi(\vt_j, \x_i))
  y_i} \le \\
  \lambda U + \norm{\prmt}_{[m]} \abs{(\vu_j - \vtu_j)^\tran \g(\vu_j)}
  + a \norm{\prmt}_{[m]} (U (\norm{\prm}_{[m]} + \norm{\prmt}_{[m]}) +
  \norm{\prm}_{R}^2 + \norm{\prmt}_{[m]}^2 V) + a \norm{\prm}_{[m]}^2
  \norm{\v_j - \vt_j}.
\end{multline*}
Notice
\[
    \begin{split}
        \abs{(\vu_j - \vtu_j)^\tran \g(\vu_j)} = & \lambda \abs{\cos(\alpha) -
        \cos(\alpha')} = 2 \lambda \sin\Par*{\frac{\alpha + \alpha'}{2}}
        \sin\Par*{\frac{\abs{\alpha - \alpha'}}{2}}
        \le 2 \lambda \sin\Par*{\frac{\beta}{2}} \sin\Par*{\frac{\beta}{2} +
        \alpha'}\\
        \le & \lambda \norm{\vu_j - \vtu_j} \Par[\bigg]{\frac{\norm{\vu_j -
        \vtu_j}}{2} + \alpha'}.
    \end{split}
\]
Also, notice
\[
  \norm{\v_j - \vt_j} = \norm{u_j \vu_j - \ut_j \vtu_j} \le \abs{u_j - \ut_j} +
  \abs{\ut_j} \norm{\vu_j - \vtu_j}.
\]
It will give
\[
  \odv{U}{t} \le \lambda U + 6 a q^2 \e^{2 \lambda (t - T_1)} U +
  \lambda \sqrt{2} q \e^{\lambda (t - T_1)} \frac{V^2}{2} + \sqrt{2} a q
  \norm{\prm(T_1)}_R^2 \e^{3 \lambda (t - T_1) + \varepsilon} + 5 \sqrt{2} a q^3
  \e^{3 \lambda (t - T_1)} V
\]

\paragraph{Bounding $U$ and $V$} Consider $h(t) \defeq \exp\Par*{-\int_{T_1}^t
2 a q^2 \e^{2 \lambda (\tau - T_1)} \d \tau} \le 1$, $\tilde{U} \defeq
\frac{U}{q} \e^{-\lambda (t - T_1)} h(t)^3$ and $\tilde{V} \defeq V h(t)$.
Notice that
\[
  h(t) = \exp\Par*{-\frac{a q^2}{\lambda} \e^{2 \lambda (t - T_1)}} \ge
  \e^{-\varepsilon} \: \forall t \in [T_1, T^\varepsilon_2],
\]
where $T^\varepsilon_2 \defeq T_1 + t^\varepsilon_4$. For $t \in [T_1,
T^\varepsilon_2]$, we get
\[
  \begin{aligned}
    \odv{\tilde{U}}{t} &\le \frac{\sqrt{2} \lambda}{2} \tilde{V}^2 +
    \sqrt{2} a \e^{2 \lambda (t - T_1) + \varepsilon} \norm{\prm(T_1)}_R^2 + 5
    \sqrt{2} a q^2 \e^{2 \lambda (t - T_1)} \tilde{V},\\
    \odv{\tilde{V}}{t} &\le \frac{a}{2} \e^{2 \lambda (t - T_1) + \varepsilon}
    \norm{\prm(T_1)}_R^2 + \sqrt{2} a q^2 \e^{2 \lambda (t - T_1) + 2
    \varepsilon} \tilde{U}.
  \end{aligned}
\]
Consider $\tilde{W} \defeq \max\{\tilde{V}, \tilde{U}\}$. We get
\[
  \odv{\tilde{W}}{t} \le \lambda \tilde{W}^2 + 2 a \e^{2 \lambda (t - T_1) +
  \varepsilon} \norm{\prm(T_1)}_R^2 + 8 a q^2 \e^{2 \lambda (t - T_1) + 2
  \varepsilon} \tilde{W}.
\]
Again, consider $W \defeq \tilde{W} h(t)^4$, we get
\[
  \odv{W}{t} \le \lambda \e^{4 \varepsilon} W^2 + 2 a \e^{2 \lambda (t - T_1) +
  5 \varepsilon} \norm{\prm(T_1)}_R^2.
\]
Now, consider the following system
\[
  \odv{\bar{W}}{t} = \lambda \e^{4 \varepsilon} \bar{W}^2 + 2 a \e^{2 \lambda
  (t - T_1) + 5 \varepsilon} \norm{\prm(T_1)}_R^2.
\]
where $\bar{W}(T_1) = W(T_1)$. It is easy to see that $\bar{W} \ge W$ since the
right-hand part is the same function in both cases, this function is increasing
in $W$, and the initial conditions are the same. Also,
notice that $\bar{W}$ is increasing. It implies
\[
  \bar{W}(t) \le \bar{W}(T_1) + \lambda \e^{4 \varepsilon} \bar{W}(t)^2 (t -
  T_1) + \frac{a}{\lambda} \e^{2 \lambda (t - T_1) + 5 \varepsilon}
  \norm{\prm(T_1)}_R^2.
\]
Assuming that $\lambda \e^{4 \varepsilon} \bar{W}(t) (t - T_1) \le
\frac{1}{2}$, we get
\[
  \bar{W} \le 2 \bar{W}(T_1) + \frac{2 a}{\lambda} \e^{2 \lambda (t - T_1) + 5
  \varepsilon} \norm{\prm(T_1)}_R^2 \le 2 \bar{W}(T_1) + \frac{2 \varepsilon
  \e^{5 \varepsilon}}{q^2} \norm{\prm(T_1)}_R^2 = \O(r^{\kappa^*} +
  r^{\kappa_R}),
\]
where $\kappa_R \defeq \min_{j \in R} \kappa_j$. Thus, $\lambda \e^{4
\varepsilon} \bar{W}(t) (t - T_1) = \O(-\ln(r) (r^{\kappa^*} + r^{\kappa_R}))$.
So, for sufficiently small $r$, our assumption holds.

Therefore, we get the following bounds
\[
  U = \O(r^{\kappa^*} + r^{\kappa_R}), \: V = \O(r^{\kappa^*} + r^{\kappa_R}),
  \: \forall j \in R \: \abs{u_j} = \O(r^{\kappa_j}).
\]

\subsection{Finding a limit of Equation (\ref{eq:gen-gf-at-g})}

Now, we want to find $\lim_{r \to 0} \prmt(T^\varepsilon_2)$ to couple
$\prm(T^\varepsilon_2)$ with the vector that does not depend on $r$. To do
this, we want to shift time $t \mapsto t - T^\varepsilon_2$ and consider
\cref{eq:gen-gf-at-g} for different $r$.
\begin{equation}
  \label{eq:gen-gf-lim}
  \begin{aligned}
    \odv{\utr_j}{t} &= \frac{1}{n} \sum_{i=1}^n (-\ell'(f(\prmtr, \x_i) y_i))
    \phi(\vtr_j,
    \x_i) y_i,\\
    \odv{\vtr_j}{t} &= \frac{1}{n} \sum_{i=1}^n (-\ell'(f(\prmtr, \x_i) y_i))
    \utr_j \nabla_{\v} \phi(\vtr_j, \x_i) y_i,
  \end{aligned}
\end{equation}
where
\[
  \utr_j(-t^{\varepsilon, r}_4) =
  \begin{cases}
    r u^*_j, & j \in P,\\
    0, & j \in R,
  \end{cases}
  \vtr_j(-t^{\varepsilon, r}_4) =
  \begin{cases}
    r \abs{u^*_j} \vu^*_j, & j \in P,\\
    0, & j \in R,
  \end{cases}
\]
and $t^{\varepsilon, r}_4 \defeq \frac{1}{2 \lambda} \ln\Par*{\frac{\lambda
\varepsilon}{2 a r^2 \norm{\prm^*}_{[m]}^2}}$.

Similarly to the previous subsections, we get
\[
  \forall t \in [-t^{\varepsilon, r}_4, 0] \: \norm{\prmtr}_{[m]}^2 \le 2 r^2
  \norm{\prm^*}_{[m]}^2 \e^{2 \lambda (t + t^{\varepsilon, r}_4)},
\]
and
\[
  \alpha^r(t) \defeq \arccos((\vu^*_j)^\tran \vtru_j) \le \frac{a}{\lambda}
  r^2 \norm{\prm^*}_{[m]}^2 (\e^{2 \lambda (t + t^{\varepsilon, r}_4)} - 1).
\]
Therefore, for $r' \ge r$
\[
  \alpha^r(-t^{\varepsilon, r'}_4) \le \frac{a}{\lambda} r^2
  \norm{\prm^*}_{[m]}^2 (\e^{2 \lambda (-t^{\varepsilon, r'}_4 +
  t^{\varepsilon, r}_4)} - 1) = \frac{a}{\lambda} \norm{\prm^*}_{[m]}^2 ((r')^2
  - r^2) = \O((r')^2).
\]

For $\utr_j$, we get
\[
    \begin{alignedat}{2}
        && \odv{(s_j \utr_j \e^{-\lambda (t + t^{\varepsilon, r}_4)})}{t} = &
        \abs{\utr_j} \e^{-\lambda (t + t^{\varepsilon, r}_4)} \Par*{\lambda
        (\cos(\alpha^r) - 1) + \frac{1}{n} \sum_{i=1}^n (-\ell'(f(\prmtr, \x_i)
        y_i) + \ell'(0)) \abs{\ut_j} \phi(\vtu_j, \x_i) y_i}\\
        && \le & a \norm{\prmtr}_{[m]}^2 \abs{\utr_j} \e^{-\lambda (t +
        t^{\varepsilon, r}_4)}\\
        & \implies & \abs{\frac{\utr_j \e^{-\lambda (t + t^{\varepsilon,
        r}_4)}}{r u^*}} \le & \exp\Par*{a \int_{-t^{\varepsilon, r}_4}^t
        \norm{\prmtr}_{[m]}^2} \le \exp\Par*{\frac{a}{\lambda} r^2
        \norm{\prm^*}_{[m]}^2 (\e^{2 \lambda (t + t^{\varepsilon, r}_4)} - 1)}.
    \end{alignedat}
\]

Notice that since $\varepsilon \le \nicefrac{1}{2}$, we get
$\nicefrac{\alpha^r}{4} \le \nicefrac{\varepsilon}{4} \le 1$. Thus,
\[
  1 - \cos(\alpha^r) \le \frac{(\alpha^r)^2}{2} \le 2 \alpha^r.
\]
It gives
\[
    \begin{split}
        & \odv{(s_j \utr_j \e^{-\lambda (t + t^{\varepsilon, r}_4)})}{t} \ge
        \abs{\utr_j} \e^{-\lambda (t + t^{\varepsilon, r}_4)} (\lambda
        (\cos(\alpha^r) - 1) - a \norm{\prmtr}_{[m]}^2) \ge \abs{\utr_j}
        \e^{-\lambda (t + t^{\varepsilon, r}_4)} (-2 \lambda \alpha^r - a
        \norm{\prmtr}_{[m]}^2)\\
        & \implies \abs{\frac{\utr_j \e^{-\lambda (t + t^{\varepsilon,
        r}_4)}}{r u^*}} \ge \exp\Par*{\frac{-2 a}{\lambda} r^2
        \norm{\prm^*}_{[m]}^2 (\e^{2 \lambda (t + t^{\varepsilon, r}_4)} - 1)}.
    \end{split}
\]
Therefore,
\[
  \abs*{\ln\Par*{\frac{\utr_j(-t^{\varepsilon,
  r'}_4)}{u^{\chi,r'}_j(-t^{\varepsilon, r'}_4)}}} \le \frac{2 a}{\lambda}
  \norm{\prm^*}_{[m]}^2 ((r')^2 - r^2) \implies \abs{\utr_j(-t^{\varepsilon,
  r'}_4) - u^{\chi,r'}_j(-t^{\varepsilon, r'}_4)} \le \O((r')^3).
\]

Notice that the above derivations also imply that
\[
  \utr_j(0) = \Theta(r u^*_j \e^{\lambda t_4^{\varepsilon, r}}) =
  \Theta(\sqrt{\varepsilon}).
\]

Thus, we can use the results of previous subsection with $\kappa = 2$ and get
that
\[
  \norm{\vtru_j(0) - \vu^{\chi,r'}_j(0)} = \O((r')^2),
  \abs{\utr_j(0) - u^{\chi,r'}_j(0)} = \O((r')^2).
\]
This property ensures that the sequences $\vtru_j(0)$ and $\utr_j(0)$ are
fundamental. Since $\utr_j(0)$ is also bounded, $\exists \lim_{r \to 0}
\prmtr(0) = \prm^{\chi,\varepsilon,*}$. Since $\prmtr(0)$ is an image of
embedding (\ref{eq:gen-emb}), we could find $\prm^{\varepsilon,*}$ such that
$\prm^{\chi,\varepsilon,*} = \chi(\prm^{\varepsilon,*})$. Also, since we have
$\forall j \in P \: \norm{\vu^{\chi,\varepsilon,r}_j(0) - \vu^*} \le
\varepsilon$, this property will also hold for the limit parameters.

\subsection{Final Bound}

Finally, using the results of both subsections, we have
\[
  \begin{aligned}
      \forall j \in P && \abs{u_j(T^\varepsilon_2) - u^{\chi,\varepsilon,*}_j}
      = & \O(r^2 + r^{\kappa^*} + r^{\kappa_R}),\\
      \forall j \in P && \norm{\vu_j(T^\varepsilon_2) -
      \vu^{\chi,\varepsilon,*}_j} = & \O(r^2 + r^{\kappa^*} + r^{\kappa_R}),\\
      \forall j \in R && \abs{u_j(T^\varepsilon_2)} = & \O(r^{\kappa_j}).
  \end{aligned}
\]

%% file: proof-spec-phase12.tex
\section{Proofs for Section \ref{sec:spec}}
\label{sec:proof-spec}

\subsection{Proof of Lemma \ref{lem:spec-g-extr}}
\label{sec:proof-spec-g-extr}

We only proof that $\ee_1$ and $-\ee_1$ are the maxima of function $G$. The
proof that $\ee_2$ and $-\ee_2$ are the minima is similar.

First, we have
\[
  G(\ee_1) = \frac{-\ell'(0)}{n} \Par*{\sum_{i \in S_1} \phi(\ee_1, \x_i)
  + \sum_{i \in S_3} \phi(\ee_1, \x_i) - \sum_{i \in S_2} \phi(\ee_1,
  \x_i) - \sum_{i \in S_4} \phi(\ee_1, \x_i)}.
\]
Notice
\[
    \begin{split}
        \sum_{i \in S_1} \phi(\ee_1, \x_i) + \sum_{i \in S_3} \phi(\ee_1,
        \x_i) = & \sum_{i \in S_1} \x^1_i \ge \frac{n}{4} (1 - \delta),\\
        \sum_{i \in S_2} \phi(\ee_1, \x_i) + \sum_{i \in S_4} \phi(\ee_1,
        \x_i) \le & \sum_{i \in S_2} (\x^1_i + \xi)_+ + \sum_{i \in S_4}
        (\x^1_i + \xi)_+ \le \frac{n}{2} (\delta + \xi).
    \end{split}
\]
Thus,
\[
  G(\ee_1) \ge \frac{-\ell'(0)}{4} (1 - 3 \delta - 2 \xi).
\]
Let $\vu^*$ be a maximum direction of $G$. W.l.o.g., assume that
$\ee_1^\tran \vu^* \ge 0$. Then we have
\[
    \begin{split}
        \sum_{i \in S_1} \phi(\vu^*, \x_i) + \sum_{i \in S_3} \phi(\vu^*,
        \x_i) \le & \frac{n}{4} (\ee_1^\tran \vu^* + 2 \xi + 2 \delta),\\
        \sum_{i \in S_2} \phi(\vu^*, \x_i) + \sum_{i \in S_4} \phi(\vu^*,
        \x_i) \ge & 0.
    \end{split}
\]
Therefore,
\[
  G(\vu^*) \le \frac{-\ell'(0)}{4} (\ee_1^\tran \vu^* + 2 \xi + 2 \delta).
\]
These properties imply
\[
  \ee_1^\tran \vu^* \ge 1 - 5 \delta - 4 \xi > \delta + \xi.
\]
Hence, $\vu^*$ should satisfy
\[
    \begin{aligned}
        \forall i \in S_3 && (\vu^*)^\tran \x_i \le & -\xi & \implies &
        \phi(\vu^*, \x_i) = 0,\\
        \forall i \in S_1 && (\vu^*)^\tran \x_i \ge & \xi & \implies &
        \phi(\vu^*, \x_i) = (\vu^*)^\tran \x_i.
    \end{aligned}
\]
Thus,
\[
  \sum_{i \in S_3} \phi(\vu^*, \x_i) + \sum_{i \in S_1} \phi(\vu^*,
  \x_i) = (\vu^*)^\tran \Par[\Bigg]{\sum_{i \in S_1} \x_i}.
\]
Similarly,
\[
  \sum_{i \in S_3} \phi(\ee_1, \x_i) + \sum_{i \in S_1} \phi(\ee_1,
  \x_i) = \ee_1^\tran \Par[\Bigg]{\sum_{i \in S_1} \x_i} \ge \frac{n}{4}
  (1 - \delta).
\]
Notice that, due to symmetry $\R_2 \R_r D_{\x} = D_{\x}$, $\sum_{i \in
S_1} \x_i$ is the multiple of $\ee_1$. Thus,
\[
  (\vu^*)^\tran \Par[\Bigg]{\sum_{i \in S_1} \x_i} = \vu^{*, 1}
  \ee_1^\tran \Par[\Bigg]{\sum_{i \in S_1} \x_i}.
\]

Now, consider the points from $S_2$ and $S_4$. Define $\x^r_i$ by the following
equality $(x^1_i, (\x^r_i)^\tran)^\tran = \x_i$. Since $\R_2 \R_r D_{\x} =
D_{\x}$, all points $i \in S_2$ have corresponding point $i^- \in
S_4$ such that $\x_{i^-} = \R_2 \R_r \x_i$. Since $\R_1 D_{\x} = D_{\x}$, all
$i \in S_2$ have a corresponding point $i^+ \in S_2$ such that $\x_{i^+} =
\R_1 \x_i$. Thus,
\[
    \begin{split}
        \sum_{i \in S_2} \phi(\vu, \x_i) + \sum_{i \in S_4} \phi(\vu,
        \x_i) = & \sum_{i \in S_2} \phi(\vu, \x_i) + \phi(\vu, \x_{i^-})\\
        = & \sum_{\substack{i \in S_2\\ x^1_i > 0}} \phi(\vu, \x_i) +
        \phi(\vu, \x_{i^-}) + \phi(\vu, \x_{i^+}) + \phi(\vu, \x_{(i^+)^-}) +
        \sum_{\substack{i \in S_2\\ x^1_i = 0}} \phi(\vu, \x_i) +
        \phi(\vu, \x_{i^-}).
    \end{split}
\]
Now,
\[
  \phi(\vu, \x_i) + \phi(\vu, \x_{i^-}) = \int_{\z \in \D^d}
  ((\vu)^\tran (\x_i + \xi \z))_+ + ((\vu)^\tran \R_2 \R_r (\x_i + \xi
  \z))_+ Q(\d \z).
\]
Denote $\x \defeq \x_i + \xi \z$ and consider integrand
\[
  ((\vu)^\tran \x)_+ + ((\vu)^\tran \R_2 \R_r \x)_+ =
  (\vu^1 x^1 + (\vu^r)^\tran \x^r)_+ + (\vu^1 x^1 - (\vu^r)^\tran \x^r)_+ \ge
  (\vu^1 x^1)_+.
\]
Therefore,
\[
  \phi(\vu, \x_i) + \phi(\vu, \x_{i^-}) \ge \int_{\z \in \D^d} (\vu^1 (x^1_i
  + \xi z^1))_+ Q(\d \z).
\]
So,
\[
  \sum_{i \in S_2} \phi(\vu, \x_i) + \sum_{i \in S_4} \phi(\vu,
  \x_i) \ge
  \sum_{\substack{i \in S_2\\ x^1_i > 0}} \int_{\z \in \D^d}
  \abs{\vu^1} \abs{x^1_i + \xi z^1} Q(\d \z) + \sum_{\substack{i \in
  S_2\\ x^1_i = 0}} \int_{\z \in \D^d} \abs{\vu^1} (\xi z^1)_+ Q(\d \z).
\]
Similarly,
\[
  \sum_{i \in S_2} \phi(\ee_1, \x_i) + \sum_{i \in S_4} \phi(\ee_1,
  \x_i) =
  \sum_{\substack{i \in S_2\\ x^1_i > 0}} \int_{\z \in \D^d}
  \abs{x^1_i + \xi z^1} Q(\d \z) + \sum_{\substack{i \in S_2\\ x^1_i = 0}}
  \int_{\z \in \D^d} (\xi z^1)_+ Q(\d \z) \le \frac{n}{2} (\delta + \xi).
\]
Thus,
\[
    \begin{split}
        \sum_{i \in S_2} \phi(\vu, \x_i) + \sum_{i \in S_4} \phi(\vu, \x_i) - &
        \sum_{i \in S_2} \phi(\ee_1, \x_i) - \sum_{i \in S_4} \phi(\ee_1,
        \x_i)\\
        \ge & (\abs{\vu^1} - 1) \Par[\Bigg]{\sum_{\substack{i \in S_2\\ x^1_i >
        0}} \int_{\z \in \D^d} \abs{x^1_i + \xi z^1} Q(\d \z) +
        \sum_{\substack{i \in S_2\\ x^1_i = 0}} \int_{\z \in \D^d} (\xi z^1)_+
        Q(\d \z)}\\
        \ge & \frac{n}{2} (\abs{\vu^1} - 1) (\delta + \xi).
    \end{split}
\]

So, we have
\[
  0 \ge G(\ee_1) - G(\vu^*) \ge \frac{(-\ell'(0))}{4} (1 - \vu^{*, 1}) (1 -
  \delta) - \frac{(-\ell'(0))}{2} (1 - \vu^{*, 1}) (\delta + \xi)
  \ge \frac{(-\ell'(0))}{4} (1 - \vu^{*, 1}) (1 - 3 \delta - 2 \xi).
\]
Therefore, $\vu^{*, 1} = 1 \implies \vu^* = \ee_1$.

\subsection{Proof of Lemma \ref{lem:spec-init}}
\label{sec:proof-spec-init}

W.l.o.g., consider direction $\ee_1$. We will show that this direction is
attractive for positive neurons ($s_j = 1$) at the beginning of training in
the region $\ee_1^\tran \vu^* \ge \delta + \xi$. Thus, the direction $\ee_1$
captures the random neuron with probability greater than $h \defeq
\frac{1}{2} \frac{\Vol(A)}{\Vol(\S^{d-1})} \ge \frac{1}{2} \Par*{\frac{1}{2}
- \frac{\Vol(\D^{d-2})}{\Vol(\S^{d-1})} \arcsin(\delta + \xi)} = \frac{1}{4}
(1 - \O(\delta + \xi))$, where $A = \Bc*{\x \in \S^{d-1} \given \ee_1^\tran
\x \ge \delta + \xi}$. This bound implies the following probability of
success
\[
  \Pr(\text{success}) \ge (1 - (1 - h)^m)^4 = 1 - 4 (1 - h)^m - \O((1 - h)^{2
  m}) =
  1 - 4 \Par*{\frac{3}{4}}^m (1 + \O(\delta + \xi)) -
  \O\Par*{\Par*{\frac{9}{16}}^m}.
\]

Now, we will proof that $A$ is attractive region for $\ee_1$. Consider
positive neuron with $\vu \in \S^{d-1}$ such that $\ee_1^\tran \vu \ge \delta
+ \xi$. \cref{eq:gen-direction-dynamics} gives
\[
  \odv{\vu}{t} = \P_{\vu} \g(\vu).
\]

We have
\[
  \odv{\norm{\vu - \ee_1}^2}{t} = -2 \ee_1^\tran \odv{\vu}{t}.
\]
So, we only need to show that $\ee_1 \odv{\vu}{t}$ is positive.

Denote $\vu \eqdef \ee_1 \cos(\alpha) + \veps \sin(\alpha)$. Notice
\[
  \P_{\vu} \ee_1 = \ee_1 - \vu \cos(\alpha) = \ee_1 \sin(\alpha)^2 -
  \veps \sin(\alpha) \cos(\alpha).
\]

Similarly to the proof of \cref{lem:spec-g-extr}, we have
\[
  \g(\vu) = \frac{-\ell'(0)}{n} \Par*{\sum_{i \in S_1} \x_i - \sum_{i \in
  S_2} \nabla_{\v} (\phi(\z, \x_i) + \phi(\z, \R_2 \R_r \x_i))\rvert_{\z =
  \vu}}.
\]
Notice
\begin{multline*}
  \nabla_{\v} (\phi(\z, \x_i) + \phi(\z, \R_2 \R_r \x_i))\rvert_{\z = \vu}\\
  = \int_{\z \in \D^d} ([\vu^\tran (\x_i + \xi \z) \ge 0] (\x_i + \xi \z) +
  [\vu^\tran \R_2 \R_r (\x_i + \xi \z) \ge 0] \R_2 \R_r (\x_i + \xi \z)) Q(\d
  \z).
\end{multline*}
Consider the integrand. Denote $\x \defeq \x_i + \xi \z$. We have four
potential cases:
\begin{enumerate}
  \item $\vu^\tran \x \le 0 \land \vu^\tran \R_2 \R_r \x \le 0$,
  \item $\vu^\tran \x > 0 \land \vu^\tran \R_2 \R_r \x \le 0$,
  \item $\vu^\tran \x \le 0 \land \vu^\tran \R_2 \R_r \x > 0$,
  \item $\vu^\tran \x > 0 \land \vu^\tran \R_2 \R_r \x > 0$.
\end{enumerate}
In the first case, the contribution of the integrand to expression $\ee_1
\odv{\vu}{t}$ is zero. In the fourth case, the contribution of the integrand
is
\[
  -(\P_{\vu} \ee_1)^\tran (\x + \R_2 \R_r \x) = -2 x^1 \sin(\alpha)^2 \ge -2
  (\delta + \xi) \sin(\alpha)^2.
\]
The second and third cases are symmetric. So, we consider only the second
case. In the second case, we have
\[
  \vu^\tran \x = x^1 \cos(\alpha) + \veps^\tran \x^r \sin(\alpha) \ge 0.
\]
Thus, $\veps^\tran \x^r \ge 0$ (otherwise $\vu^\tran \R_2 \R_r \x \ge
\vu^\tran \x > 0$). It implies
\[
  -(\P_{\vu} \ee_1)^\tran \x = -x^1 \sin(\alpha)^2 + \veps^\tran \x
  \cos(\alpha) \sin(\alpha) \ge -x^1 \sin(\alpha)^2.
\]
Therefore, we get
\[
  \ee_1 \odv{\vu}{t} \ge \frac{-\ell'(0)}{n} \Par*{\frac{n}{4} (1 - \delta) -
  \frac{n}{4} (2 (\delta + \xi))} \sin(\alpha)^2 \ge \frac{-\ell'(0)}{8}
  \sin(\alpha)^2.
\]
It gives
\[
  \odv{\alpha}{t} \le \frac{-\ell'(0)}{8} \sin(\alpha) \le \frac{-\ell'(0)}{4
  \uppi} \alpha,
\]
which implies exponentially fast convergnce to $\ee_1$.

%% file: proof-spec-phase3.tex
\subsection{Proof of Proposition \ref{prop:spec-conv}}
\label{sec:proof-spec-conv}

First, notice that, due to symmetry $\R_r D_{\x} = D_{\x}$, the features of
4-neuron network belong to a sub-space generated by vectors $\ee_1$ and
$\ee_2$ and clustered around these directions.

\begin{lemma}
    \label{lem:spec-2d-features}
    $\v^{*, \varepsilon}_j$ form \cref{thm:gen-phase2} belong to a sub-space
    generated by vectors $\ee_1$ and $\ee_2$ ($\R_r \v^{*, \varepsilon}_j =
    \v^{*, \varepsilon}_j$). Moreover, this symmetry is preserved under
    gradient flow dynamics (\ref{eq:set-gf}).
\end{lemma}

\begin{proof}
  Consider \cref{eq:gen-gf-at-g}. Notice that at the start of function
  $h(\x) \defeq f(\prmt(T_1), \x)$ depends only on $x^1$ and $x^2$. Due to
  symmetry $\R_r D_{\x} = D_{\x}$, this property will hold for the right-hand
  part of \cref{eq:gen-gf-at-g}. Therefore, features will not be able to escape
  the sub-space generated by $\ee_1$ and $\ee_2$. Thus, limit features $\v^{*,
  \varepsilon}_j$ will also lie in the sub-space generated by $\ee_1$ and
  $\ee_2$. Since the limit and dataset respect symmetry $\R_r$, it will
  propagate to further stages of training.
\end{proof}

Consider dynamics
\[
    \begin{split}
        \odv{\ue_k}{t} = & \sum_{i=1}^n (-\ell'(f(\prme, \x_i) y_i))
        \phi(\ve_k, \x_i) y_i,\\
        \odv{\ve_k}{t} = & \sum_{i=1}^n (-\ell'(f(\prme, \x_i) y_i)) \ue_k
        \nabla_{\v} \phi(\ve_k, \x_i) y_i,
    \end{split}
\]
where $\ue_k(0) = s_{P_k} \sqrt{\sum_{j \in P_k} (u^{\varepsilon, *}_j)^2}$ and
$\ve_k(0) = \abs{\ue_k(0)} \vu^{\varepsilon, *}_{P_k}$. Notice that, by
\cref{lem:spec-2d-features}, $\ve_k \in \vspan\{\ee_1, \ee_2\}$.

\begin{proposition}
  \label{prop:spec-ang-max}
  Denote a signed angle between $\R_1^{\Floor*{\frac{k-1}{2}}} \P^{a(k)} \ve_k$
  and $\ee_{a(k) + 1}$ by $\alpha_k$. Then we have $\abs{\sin(\alpha_k)} \le
  \delta + \xi$.
\end{proposition}

\begin{proof}
  W.l.o.g., consider $\ve_1$ and assume $\alpha_1 \ge 0$. We have
  \[
    \odv{\vue_1}{t} = \sum_{i=1}^n (-\ell'(f(\prme, \x_i) y_i)) \P_{\ve_1}
    \nabla_{\v} \phi(\ve_1, \x_i) y_i.
  \]
  Notice
  \[
    \ee_1^\tran \P_{\ve_1} \nabla_{\v} \phi(\ve_1, \x_i) y_i = \int_{\z \in
    \D^d} [(\ve_1)^\tran (\x_i + \xi \z) \ge 0] \ee_1^\tran \P_{\ve_1} (\x_i +
    \xi \z) y_i Q(\d \z).
  \]
  Consider the integrand and denote $\z_i \defeq \x_i + \xi \z$. Assuming that
  $\alpha_1 \le \nicefrac{\uppi}{4}$, we get
  \[
    \begin{split}
      \forall i \in S_1 \: [(\ve_1)^\tran \z_i \ge 0] \ee_1^\tran \P_{\ve_1}
      \z_i y_i & =
      (\ee_1 \sin(\alpha_1) - \ee_2 \cos(\alpha_1))^\tran \sin(\alpha_1)
      (\ee_1 + \z_i - \ee_1) \ge (\sin(\alpha_1) - (\delta + \xi))
      \sin(\alpha_1),\\
      \forall i \in S_2 \: [(\ve_1)^\tran \z_i \ge 0] \ee_1^\tran \P_{\ve_1}
      \z_i y_i & =
      -[(\ve_1)^\tran \z_i \ge 0] (\ee_1 \sin(\alpha_1) - \ee_2
      \cos(\alpha_1))^\tran \sin(\alpha_1) (\ee_2 + \z_i - \ee_2)\\
      & \ge (\cos(\alpha_1) - (\delta + \xi)) \sin(\alpha_1)\\
      & \ge 0,\\
      \forall i \in S_3 \: [(\ve_1)^\tran \z_i \ge 0] \ee_1^\tran \P_{\ve_1}
      \z_i y_i & = 0,\\
      \forall i \in S_4 \: [(\ve_1)^\tran \z_i \ge 0] \ee_1^\tran
      \P_{\ve_1} \z_i y_i & \ge -[\sin(\alpha_1) \le \delta + \xi]
      (1 + \delta + \xi).
    \end{split}
  \]
  Thus, the function $h(\alpha_1) = \max(\alpha_1, \arcsin(\delta + \xi))$ is
  always decreasing (when $\alpha_1$ is less than $\arcsin(\delta + \xi)$, the
  time derivative of $h(\alpha_1)$ is zero, when $\alpha_1$ is greater than
  $\arcsin(\delta + \xi)$, the time derivative of $h(\alpha_1)$ is determined
  only by the points from $S_1$ and $S_2$ and hence negative). Therefore,
  $\alpha_1$ never exceeds $\arcsin(\delta + \xi)$.
\end{proof}

\begin{proof}[Proof of \cref{prop:spec-conv}]
  Denote $a \defeq 0.01$, $b \defeq 0.001$, $c = 1000$, $q \defeq
  \nicefrac{8}{3}$, $T \defeq \inf\Bc*{t > 0 \given \max_k \abs{\ue_k(t)} \ge
  q}$.

  From the previous proposition, we know that $\abs{\sin(\alpha_k)} \le \delta
  + \xi \le a$. Thus, for $t \in [0, T]$ and $i \in S_1$, we get
  \[
    f(\prme, x_i) = (\ue_1)^2 (\vue_1)^\tran \x_i - (\ue_2)^2 \phi(\vue_2,
    \x_i) - (\ue_4)^2 \phi(\vue_4, \x_i).
  \]
  Therefore,
    \[
        \begin{split}
            f(\prme, x_i) &\le (\ue_1)^2 (1 + \delta) \le (\ue_1)^2 (1 + a),\\
            f(\prme, x_i) &\ge (\ue_1)^2 (\cos(\alpha_k) - \delta) - q^2
            (\phi(\vue_2, \x_i) + \phi(\vue_4, \x_i)) \ge (\ue_1)^2 (1 -
            (\delta + \xi)^2 - \delta) - q^2 (4 \delta + 2 \xi)\\
            & \ge (\ue_1)^2 (1 - 2 a) - 4 q^2 a, 
        \end{split}
    \]
  where the last inequality follows from the following property:
    \[
        \begin{split}
            \forall j \in \{2, 4\} \: \phi(\vue_j, \x_i) = & \int_{-1}^1
            ((\vue_j)^\tran \x_i + \xi z)_+ (1 - z^2)^{\frac{d-1}{2}}
            \frac{\Vol(\D^{d-1})}{\Vol(\D^d)} \d z\\
            \le & \int_{-1}^1 (\abs{\sin(\alpha_j)} + \delta + \xi z)_+ (1 -
            z^2)^{\frac{d-1}{2}} \frac{\Vol(\D^{d-1})}{\Vol(\D^d)} \d z\\
            \le & \int_{-1}^1 (2 \delta + \xi + \xi z) (1 -
            z^2)^{\frac{d-1}{2}} \frac{\Vol(\D^{d-1})}{\Vol(\D^d)} \d z\\
            = & 2 \delta + \xi.
        \end{split}
    \]
  Similarly, we could derive the same inequalities for $S_2, \dots, S_4$.

  These inequalities imply
    \[
        \begin{split}
            \odv{\ue_1}{t} = & \frac{1}{n} \sum_{i=1}^n (-\ell'(f(\prme, \x_i)
            y_i)) \phi(\ve_1, \x_i) y_i\\
            \ge & \frac{\ue_1}{4} \biggl(\frac{1 - 2 a}{1 + \exp((\ue_1)^2 (1 +
            a))} - \frac{2 a}{1 + \exp((\ue_2)^2 (1 - 2 a) - 4 q^2 a)} -
            \frac{2 a}{1 + \exp((\ue_4)^2 (1 - 2 a) - 4 q^2 a)}\biggr)
        \end{split}
    \]
  and
  \[
    \odv{\ue_1}{t} \le \frac{\ue_1 (1 + a)}{4 (1 + \exp((\ue_1)^2 (1 - 2 a) -
    4 q^2 a))}.
  \]

  Denote $x \defeq \min_k \abs{\ue_k}$ and $y \defeq \max_k \abs{\ue_k}$. The
  property above implies
  \[
    \begin{aligned}
      \odv{x}{t} &\ge \frac{x}{4} \Par*{\frac{1 - 2 a}{1 + \exp(x^2 (1 + a))} -
      \frac{4 a}{1 + \exp(x^2 (1 - 2 a) - 4 q^2 a)}}
      \ge \frac{x ((1 - 2 a) e^{-7 q^2 a} - 4 a)}{4 (1 + \exp(x^2 (1 - 2 a) -
      4 q^2 a))},\\
      \odv{y}{t} &\le \frac{y (1 + a)}{4 (1 + \exp(y^2 (1 - 2 a) - 4 q^2 a)}.
    \end{aligned}
  \]
  Now, denote $X \defeq x^2 (1 - 2 a)$, $Y \defeq y^2 (1 - 2 a)$, $A \defeq
  \frac{(1 - 2 a) e^{-7 q^2 a} - 4 a}{2}$, $B \defeq \frac{1 + a}{2}$, and $C
  \defeq e^{-4 q^2 a}$. We get
  \[
    \begin{aligned}
        \odv{X}{t} \ge & \frac{A X}{1 + C \e^X} & \implies &&
        \int_{X(0)}^{X(t)} \frac{\d X (1 + C \e^X)}{X} \ge & A t,\\
        \odv{Y}{t} \le & \frac{B Y}{1 + C \e^Y} & \implies &&
        \int_{Y(0)}^{Y(t)} \frac{\d Y (1 + C \e^Y)}{Y} \le & B t.
    \end{aligned}
  \]
  Denote $h(x) \defeq \int_1^x \frac{\d z (1 + C \e^z)}{z}$. We get
  \[
    h(X(t)) - h(X(0)) \ge A t, \: h(Y(t)) - h(Y(0)) \le B t.
  \]

  We have two possible cases: $T < \infty$ and $T = \infty$. In the second
  case, we notice that our lower bound for the derivative of $x$ holds on the
  whole timeline. In this case, $\lim_{t \to \infty} x(t) = \infty$. Thus, at
  some time point we would get $x > q$, which contradicts definition of $T$.
  Therefore, $T$ is finite.

  Now, notice that $Y(T) = (1 - 2a) q^2$ and we have
    \[
        \begin{split}
            & B (h(X(T)) - h(X(0))) \ge A B T \ge A (h(Y(T)) - h(Y(0)))\\
            \implies & h(X(T)) \ge \frac{A}{B} h(Y(T)) + \frac{B - A}{B}
            h(Y(0)) - h(Y(0)) + h(X(0)).
        \end{split}
    \]
  By the definition of $h$,
  \[
    \begin{aligned}
      h(Y(T)) &= \ln(Y(T)) + C \int_1^{Y(T)} \frac{\e^z}{z} \d z,\\
      h(Y(0)) &= -\int_{Y(0)}^1 \frac{1 + C \e^z}{z} \d z \ge (1 + C \e)
      \ln(Y(0)) = (1 + C \e) \ln((1 - 2 a) b^2),\\
      h(Y(0)) - h(X(0)) &= \int_{X(0)}^{Y(0)} \frac{1 + C \e^z}{z} \d z \le (1
      + C \e^{Y(0)}) \ln\Par*{\frac{Y(0)}{X(0)}} \le (1 + C \e^{\frac{1 -
      2a}{4}}) \ln(c^2).
    \end{aligned}
  \]
  Using numerical integration, we get $h(X(T)) \ge 31.52$, which implies that
  $x(T) \ge \nicefrac{9}{4}$.

  Therefore, at time $T$, we have
  \[
    f(\prme(T), \x_i) y_i \ge (1 - 2 a) x(T)^2 - 4 q^2 a > 4.67 > 0,
  \]
  the network classifies all points correctly.
\end{proof}

\subsection{Proof of Proposition \ref{prop:spec-loc-max-marg}}
\label{sec:proof-spec-loc-max-marg}

First, consider direction $\prmmm$ and w.l.o.g. assume that $\norm{\prmmm}^2 =
8$, which implies $\forall k \: \abs{\umm_k} = 1$. Consider some orbit, $M$, of
data points under the group generated by $\P, \R_1, \R_2, \R_r$.

Choose $i^* \in M \cap S_1 : \ee_2^\tran \x_{i^*} \ge 0$. For this point, we
have
\[
  f(\prmmm, \x_{i^*}) = \sum_{k=1}^4 (\umm_k)^2 \phi(\vmmu_k, \x_{i^*}) =
  \phi(\vmmu_1, \x_{i^*}) - \phi(\vmmu_2, \x_{i^*}).
\]
Since $\forall a \: \abs{\ee_a^\tran \x_{i^*}} > \xi$, we get
\[
  f(\prmmm, \x_{i^*}) = x^1_{i^*} - x^2_{i^*}.
\]
Due to symmetry of the network, all points in $M$ will have the same margin.

Now, consider a small perturbation, $\prm$, of $\chi(\prmmm)$, which does not
change activation patterns for points in $M$. Notice
\[
  \min_{i \in M} \frac{f(\prm, \x_i) y_i}{\norm{\prm}^2} \le \frac{1}{16
  \norm{\prm}^2} \sum_{i \in M} f(\prm, \x_i) y_i.
\]
However,
\[
    \begin{split}
        \sum_{i \in M} f(\prm, \x_i) y_i = & \sum_{k=1}^4 \sum_{j \in P_k}
        \sum_{l=1}^4 \sum_{i \in S_l} u_j \phi(\v_j, \x_i) y_i = \sum_k \sum_{j
        \in P_k} 4 \abs{u_j} \abs{\v_j^{a(k)+1}} (x_{i^*}^1 - x_{i^*}^2)\\
        \le & \sum_k \sum_{j \in P_k} 2 (\abs{u_j}^2 + \norm{\v_j}^2)
        (x_{i^*}^1 - x_{i^*}^2).
    \end{split}
\]
Thus,
\[
  \min_{i \in M} \frac{f(\prm, \x_i) y_i}{\norm{\prm}^2} \le \frac{x^1_i -
  x^2_i}{8} = \min_{i \in M} \frac{f(\prmmm, \x_i) y_i}{\norm{\prmmm}^2}.
\]
It implies that, for sufficiently small perturbation,
\[
  \min_i \frac{f(\prm, \x_i) y_i}{\norm{\prm}^2} = \min_M \min_{i \in M}
  \frac{f(\prm, \x_i) y_i}{\norm{\prm}^2} \le \min_i \frac{f(\prmmm, \x_i)
  y_i}{\norm{\prmmm}^2},
\]
where the equality is achieved only if $\abs{u_j} = \norm{\v_j}$ and $\v_j$
is aligned with $b(k) \ee_{a(k)+1}$. However, in this case,
\[
  f(\prm, \x_{i^*}) = U_1^2 x_{i^*}^1 - U_2^2 x_{i^*}^2,
\]
where $U_k = s_{P_k} \sqrt{\sum_{j \in P_k} u_j^2}$. W.l.o.g., we could assume that $U_1^2 = \min_k U_k^2$, then, we get
\[
  \frac{f(\prm, \x_{i^*})}{\norm{\prm}^2} \le \frac{U_1^2 x_{i^*}^1 - U_1^2
  x_{i^*}^2}{2 \sum_k U_k^2}.
\]
But this margin should be equal to $\frac{x^1_{i^*} - x^2_{i^*}}{8}$. Thus,
$4 U_1^2 \ge \sum_k U_k^2$. However, this implies that $U_1^2 = U_2^2 = U_3^2 =
U_4^2$, i.e., $\prm$ is proportional to $\chi(\prmmm)$. Therefore,
$\chi(\prmmm)$ is indeed local extremum. Similarly, we can show that $\prmmm$
is an isolated local extremum.

\subsection{Proof of Lemma \ref{lem:spec-extreme}}
\label{sec:proof-spec-extreme}

Denote 4-neuron network initialized at $\prm^{\varepsilon,*}$ as $\prmf$. We
get
\[
  \begin{aligned}
    \odv{\uf_j}{t} &= \frac{1}{n} \sum_{i=1}^n (-\ell'(f(\prmf, \x_i) y_i))
    \phi(\vf_j, \x_i) y_i,\\
    \odv{\vf_j}{t} &= \frac{1}{n} \sum_{i=1}^n (-\ell'(f(\prmf, \x_i) y_i))
    \uf_j \nabla_{\v} \phi(\vf_j, \x_i) y_i,
  \end{aligned}
\]
where $\prmf(0) = \prm^{\varepsilon,*}$.

First, we want to show that $\norm{\prmf} \xrightarrow[]{t \to \infty} \infty$.
Similarly to the proof of \cref{prop:spec-conv}, using
\cref{prop:spec-ang-max}, we get that for all $i \in S_1$
\begin{align*}
  f(\prmf, \x_i) &\le (\ue_1)^2 (1 + a),\\
  f(\prmf, \x_i) &\ge (\ue_1)^2 (1 - 2 a) - 2 a ((\ue_2)^2 + (\ue_4)^2),
\end{align*}
where $a \defeq \delta + \xi$. Since $\prmf$ converges in direction to
$\prmmm$, we could find a moment when ratio $\frac{8
(\ue_k)^2}{\norm{\prmf}^2}$ will lie in interval $(1 - \epsilon, 1 +
\epsilon)$, where $\frac{3 \epsilon}{1 - \epsilon} < \frac{1 - 6 a}{4 a}$.
Then,
\[
  f(\prmf, \x_i) \ge b \norm{\prmf}^2,
\]
where $b \defeq \frac{(1 - \epsilon) (1 - 2 a) - (1 + \epsilon) 2 a}{8} > 0$.
Similarly, for other clusters. Also define $c \defeq \frac{(1 + \epsilon) (1 +
a)}{8}$. Now, notice that
\[
    \begin{split}
        \frac{1}{4} \odv{\norm{\prm}^2}{t} = & \sum_{k=1}^4 \ue_k
        \odv{\ue_k}{t} = \frac{1}{n} \sum_{i=1}^n \sum_{k=1}^4 \ue_k
        (-\ell'(f(\prmf, \x_i) y_i)) \phi(\ve_k, \x_i) y_i = \frac{1}{n}
        \sum_{i=1}^n \frac{f(\prmf, \x_i) y_i}{1 + \exp(f(\prmf, \x_i) y_i)}\\
        \ge & \frac{b \norm{\prmf}^2}{1 + \exp(c \norm{\prmf})}.
    \end{split}
\]
Similarly to the proof of \cref{prop:spec-conv}, this differential inequality
means that $\norm{\prmf} \xrightarrow[]{t \to \infty} \infty$.

Now, we want to apply \cref{thm:spec-extr-bias}. To do it, choose $\zeta$ small
enough so that $\chi(\prmmm)$ becomes the biggest local-max-margin direction in
$\zeta$-neighborhood around its image in the original weight space. Then, apply
\cref{thm:spec-extr-bias} for $\chi(\prmmm)$, which gives us parameters
$\omega$ and $\rho$. Consider time $T$ when $\chi(\prmf)$ converged to the
desired local-max-margin direction closer than $\nicefrac{\omega}{2}$ and its
scale became bigger than $2 \rho$. After that, apply Theorem 2.1, Chapter 5
from \citet{h02o} (notice that our activation function is twice continuously
differentiable) and choose the initial scale $\sigma$ to be sufficiently small
so that original system at time $\prm(T + T_2^\varepsilon)$ is bigger than
$\rho$ and $\nicefrac{\omega}{2}$-close in direction to $\prmf$. Then, $\prm$
will converge to some direction with normalized margin bigger than that of
$\chi(\prmmm)$. However, since $\chi(\prmmm)$ is a strict local-max-margin
direction, this would mean that $\prm$ will converge in direction to
$\chi(\prmmm)$.

%% file: conv-graphs.tex
\newpage

\section{Additional Experiments for Section \ref{sec:spec}}
\label{sec:conv-graphs}

\begin{figure}[!hb]
  \centering
  \input{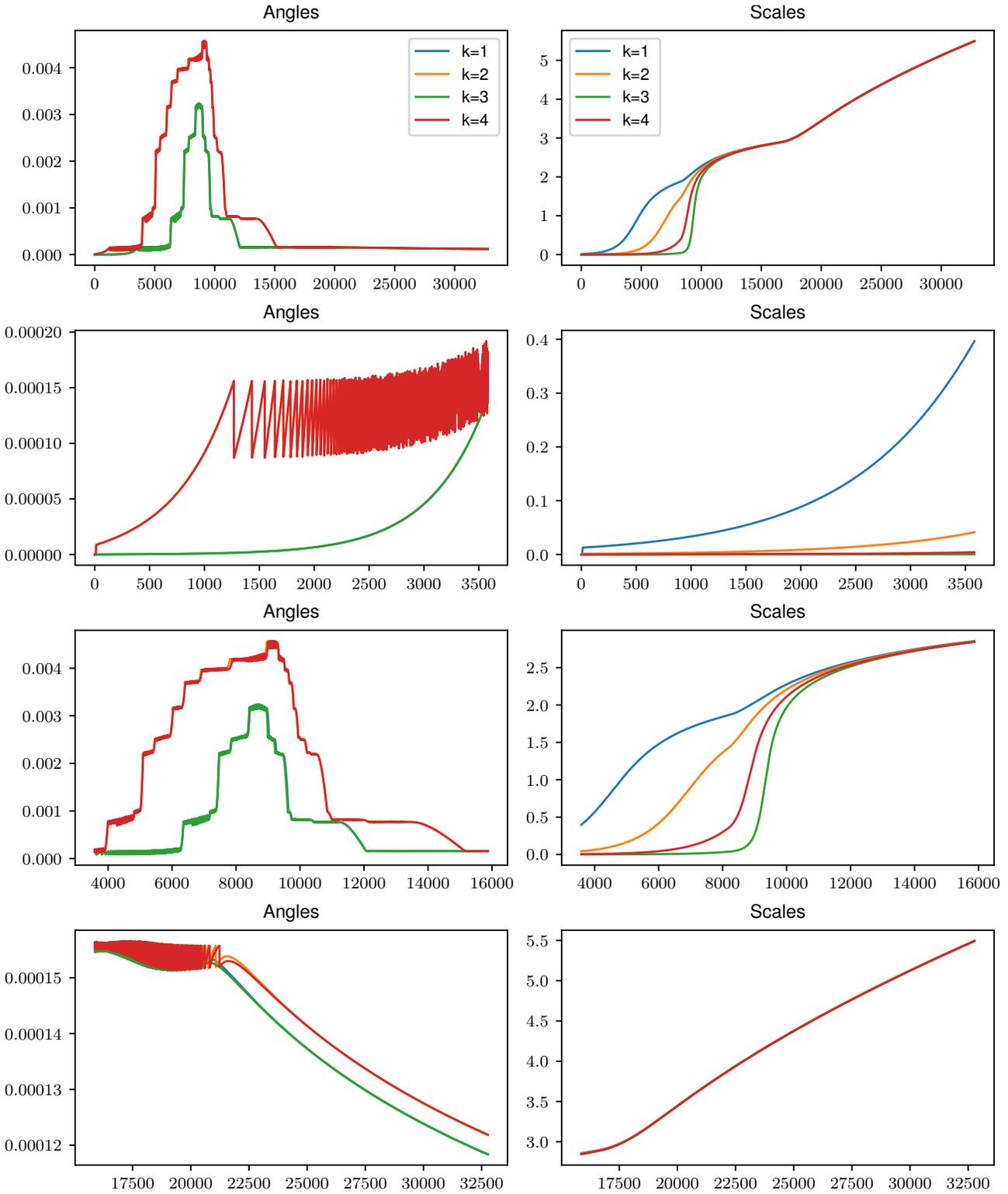}
  \caption{Evolution of 4-neuron network initialized at $(\ue_1(0),
  \ue_2(0), \ue_3(0), \ue_4(0)) = (10^{-4}, -10^{-5}, 10^{-7}, -10^{-6})$. The
  first row depicts the whole training process; the second row depicts
  the first $3584$ training epochs; the third row depicts the epochs from
  $3584$ to $15872$; the last row depicts training after the $15872$th
  epoch. Notice that $\alpha_1 \approx \alpha_3$ and $\alpha_2 \approx
  \alpha_4$.}
  \label{fig:evo1}
\end{figure}

\newpage

\begin{figure}[!ht]
  \centering
  \input{evo2.pgf}
  \caption{Evolution of 4-neuron network initialized at $(\ue_1(0),
  \ue_2(0), \ue_3(0), \ue_4(0)) = (10^{-4}, -10^{-5}, 10^{-6}, -10^{-7})$. The
  first row depicts the whole training process; the second row depicts
  the first $3584$ training epochs; the third row depicts the epochs from
  $3584$ to $15872$; the last row depicts training after the $15872$th
  epoch. Notice that $\alpha_1 \approx \alpha_3$ and $\alpha_2 \approx
  \alpha_4$.}
  \label{fig:evo2}
\end{figure}

\newpage

\begin{figure}[!ht]
  \centering
  \input{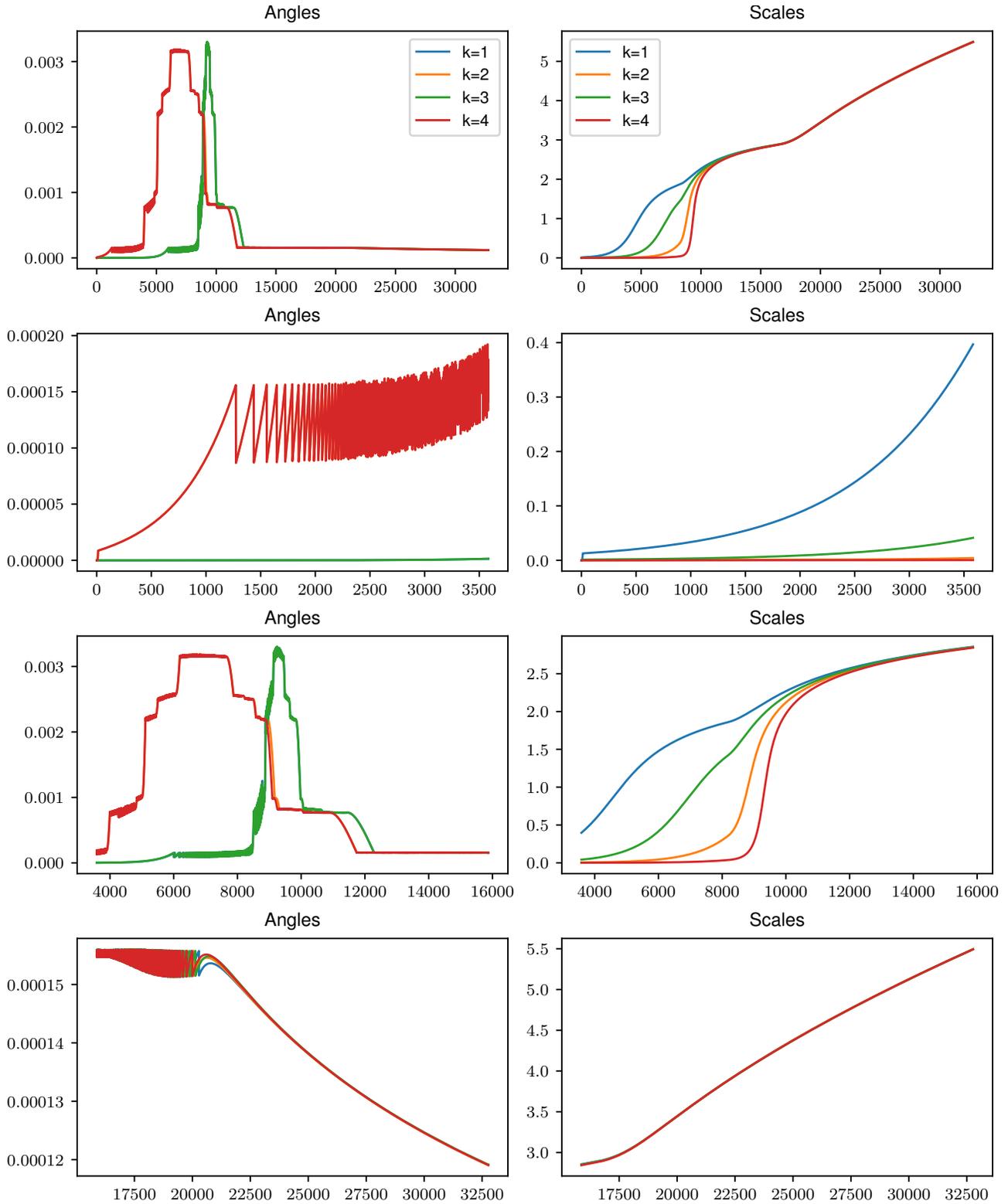}
  \caption{Evolution of 4-neuron network initialized at $(\ue_1(0),
  \ue_2(0), \ue_3(0), \ue_4(0)) = (10^{-4}, -10^{-6}, 10^{-5}, -10^{-7})$. The
  first row depicts the whole training process; the second row depicts
  the first $3584$ training epochs; the third row depicts the epochs from
  $3584$ to $15872$; the last row depicts training after the $15872$th
  epoch. Notice that $\alpha_1 \approx \alpha_3$ and $\alpha_2 \approx
  \alpha_4$.}
  \label{fig:evo3}
\end{figure}

\newpage

%% file: exper-details.tex
\section{Additional Experimental Results}

\subsection{Experimental Details}
\label{sec:exper-details}

\paragraph{Data} We used the usual MNIST and CIFAR-10 datasets for creation of
dominos. For creating train and test data, we used the default train-test split
of these datasets, resulting in $50000$ images in train set and $10000$ images
in test set. We further devoted $25\%$ train and test data for validation,
giving us four datasets: train-train, train-validation, test-train, and
test-validation. We also normalize images in these datasets using the default
values for MNIST and CIFAR-10. During training, we also apply random horizontal
flip augmentation.

\paragraph{Model} We used the standard model from Torchvision library, but
changed the first layer to $3 \times 3$ convolutions instead of the default $7
\times 7$ convolutions. Additionally, after initialization we multiplied all
parameters of the model by a factor $2^{-5}$ to capture the desired simplicity
bias mechanism.

\paragraph{Optimization procedure} We use the standard SGD optimizer
from PyTorch and linear learning scheduler with warm-up from Transformers
library. The parameters of data and optimizer are listed below.

\begin{table}[ht]
  \centering
  \begin{tabular}{@{}lr@{}}
    \toprule
    \texttt{batch\_size} & \texttt{128}\\
    \texttt{lr} & \texttt{0.125}\\
    \texttt{momentum} & \texttt{0.9}\\
    \texttt{nesterov} & \texttt{True}\\
    \texttt{weight\_decay} & \texttt{0.0005}\\
    Share of warm-up steps & $12.5\%$\\
    \bottomrule
  \end{tabular}
\end{table}

\paragraph{Parameters of logistic regression} We used the standard
implementation of the logistic regression from scikit-learn library. By
default, we use the following parameters.

\begin{table}[ht]
  \centering
  \begin{tabular}{@{}lr@{}}
    \toprule
    \texttt{penalty} & \texttt{l2}\\
    \texttt{C} & \texttt{1000}\\
    \texttt{max\_iter} & \texttt{20000}\\
    \bottomrule
  \end{tabular}
\end{table}

Notice that effectively the current version of scikit-learn library does not
allow to change the parameter \texttt{maxfun} parameter of the \texttt{lbfgs}
optimizer. Thus, to ensure convergence we rerun fitting procedure $50$ times
using warm start.

\subsection{Additional Experiments}
\label{sec:exper-add}

\begin{figure}[!ht] 
  \centering
  \input{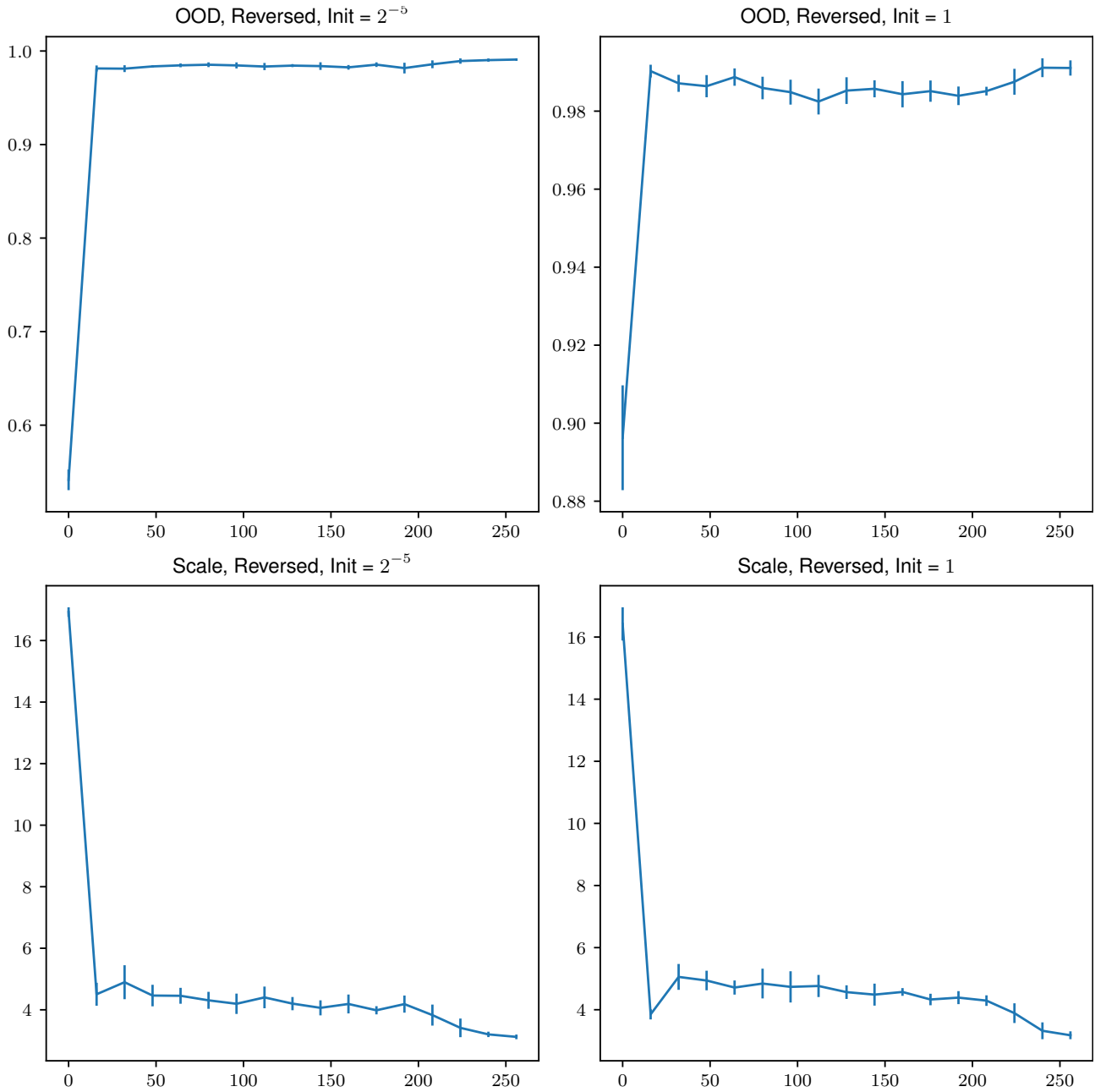}
  \caption{Accuracy and scale of the logistic regression on the validation part
  of the OOD test set ($y$-axis) vs. the training epoch at which the ResNet
  features are extracted ($x$-axis).}
  \label{fig:ood-rev}
\end{figure}

\begin{figure}[!ht] 
  \centering
  \input{ood-err5.pgf}
  \caption{Accuracy and scale of the logistic regression on the validation part
  of the OOD test set ($y$-axis) vs. the training epoch at which the ResNet
  features are extracted ($x$-axis).}
  \label{fig:ood-err5}
\end{figure}
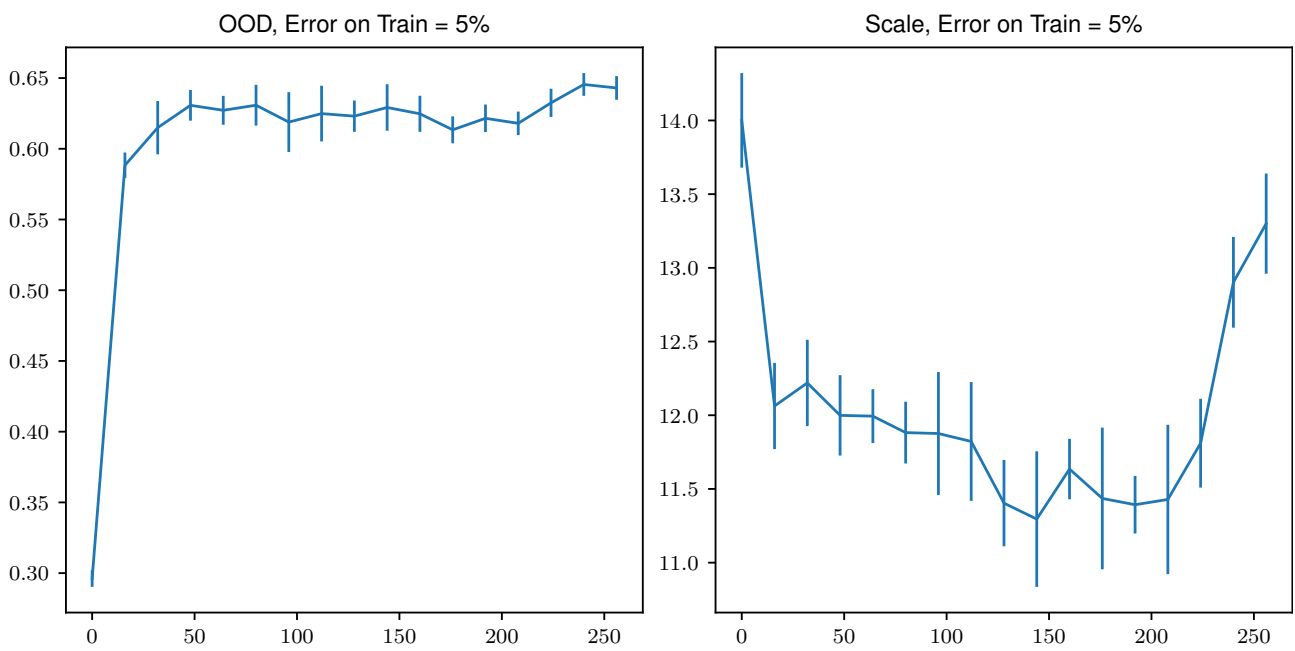

Figures \ref{fig:ood-rev}, \ref{fig:ood1}, and \ref{fig:ood-err5} depict
additional results for \cref{sec:exper}.

For \cref{fig:ood-rev}, we repeated the experiment but used MNIST labels on
the test set. As we can see, the learned features are sufficient to achieve
almost perfect accuracy on MNIST label, indicating that the network learned
``simple'' MNIST features.

For \cref{fig:ood1}, we repeated the experiment but did not scale the model at
initialization. As we can see, the drop in OOD accuracy is less pronounced for
the model initialized from the normal scale (approximately $6.43\% \pm 2.43\%$,
implying p-value around $0.00062$). This experiment indicates that closeness to
more lazy training regime is indeed beneficial for OOD generalization in
presence of simplicity bias.

For \cref{fig:ood-err5}, we repeated the experiment on different
train data, on which the correlation between MNIST and CIFAR-10 classes is not
perfect and with $5\%$ probability MNIST class might not match CIFAR-10 class.
As we can see, the model still experience simplicity bias. However, the drop in
OOD accuracy at the end of training disappears.

\begin{figure}[!ht] 
  \centering
  \input{ood1.pgf}
  \caption{Accuracy and scale of the logistic regression on the validation part
  of the OOD test set ($y$-axis) vs. the training epoch at which the ResNet
  features are extracted ($x$-axis).}
  \label{fig:ood1}
\end{figure}

\clearpage

%% file: ood-err5.pgf
\begingroup%
\makeatletter%
\begin{pgfpicture}%
\pgfpathrectangle{\pgfpointorigin}{\pgfqpoint{6.716660in}{3.316660in}}%
\pgfusepath{use as bounding box, clip}%
\begin{pgfscope}%
\pgfsetbuttcap%
\pgfsetmiterjoin%
\definecolor{currentfill}{rgb}{1.000000,1.000000,1.000000}%
\pgfsetfillcolor{currentfill}%
\pgfsetlinewidth{0.000000pt}%
\definecolor{currentstroke}{rgb}{1.000000,1.000000,1.000000}%
\pgfsetstrokecolor{currentstroke}%
\pgfsetdash{}{0pt}%
\pgfpathmoveto{\pgfqpoint{0.000000in}{0.000000in}}%
\pgfpathlineto{\pgfqpoint{6.716660in}{0.000000in}}%
\pgfpathlineto{\pgfqpoint{6.716660in}{3.316660in}}%
\pgfpathlineto{\pgfqpoint{0.000000in}{3.316660in}}%
\pgfpathlineto{\pgfqpoint{0.000000in}{0.000000in}}%
\pgfpathclose%
\pgfusepath{fill}%
\end{pgfscope}%
\begin{pgfscope}%
\pgfsetbuttcap%
\pgfsetmiterjoin%
\definecolor{currentfill}{rgb}{1.000000,1.000000,1.000000}%
\pgfsetfillcolor{currentfill}%
\pgfsetlinewidth{0.000000pt}%
\definecolor{currentstroke}{rgb}{0.000000,0.000000,0.000000}%
\pgfsetstrokecolor{currentstroke}%
\pgfsetstrokeopacity{0.000000}%
\pgfsetdash{}{0pt}%
\pgfpathmoveto{\pgfqpoint{0.297380in}{0.186266in}}%
\pgfpathlineto{\pgfqpoint{3.316660in}{0.186266in}}%
\pgfpathlineto{\pgfqpoint{3.316660in}{3.146521in}}%
\pgfpathlineto{\pgfqpoint{0.297380in}{3.146521in}}%
\pgfpathlineto{\pgfqpoint{0.297380in}{0.186266in}}%
\pgfpathclose%
\pgfusepath{fill}%
\end{pgfscope}%
\begin{pgfscope}%
\pgfsetbuttcap%
\pgfsetroundjoin%
\definecolor{currentfill}{rgb}{0.000000,0.000000,0.000000}%
\pgfsetfillcolor{currentfill}%
\pgfsetlinewidth{0.702625pt}%
\definecolor{currentstroke}{rgb}{0.000000,0.000000,0.000000}%
\pgfsetstrokecolor{currentstroke}%
\pgfsetdash{}{0pt}%
\pgfsys@defobject{currentmarker}{\pgfqpoint{0.000000in}{-0.038889in}}{\pgfqpoint{0.000000in}{0.000000in}}{%
\pgfpathmoveto{\pgfqpoint{0.000000in}{0.000000in}}%
\pgfpathlineto{\pgfqpoint{0.000000in}{-0.038889in}}%
\pgfusepath{stroke,fill}%
}%
\begin{pgfscope}%
\pgfsys@transformshift{0.434620in}{0.186266in}%
\pgfsys@useobject{currentmarker}{}%
\end{pgfscope}%
\end{pgfscope}%
\begin{pgfscope}%
\definecolor{textcolor}{rgb}{0.000000,0.000000,0.000000}%
\pgfsetstrokecolor{textcolor}%
\pgfsetfillcolor{textcolor}%
\pgftext[x=0.434620in,y=0.098766in,,top]{\color{textcolor}\sffamily\fontsize{8.000000}{9.600000}\selectfont \(\displaystyle {0}\)}%
\end{pgfscope}%
\begin{pgfscope}%
\pgfsetbuttcap%
\pgfsetroundjoin%
\definecolor{currentfill}{rgb}{0.000000,0.000000,0.000000}%
\pgfsetfillcolor{currentfill}%
\pgfsetlinewidth{0.702625pt}%
\definecolor{currentstroke}{rgb}{0.000000,0.000000,0.000000}%
\pgfsetstrokecolor{currentstroke}%
\pgfsetdash{}{0pt}%
\pgfsys@defobject{currentmarker}{\pgfqpoint{0.000000in}{-0.038889in}}{\pgfqpoint{0.000000in}{0.000000in}}{%
\pgfpathmoveto{\pgfqpoint{0.000000in}{0.000000in}}%
\pgfpathlineto{\pgfqpoint{0.000000in}{-0.038889in}}%
\pgfusepath{stroke,fill}%
}%
\begin{pgfscope}%
\pgfsys@transformshift{0.970713in}{0.186266in}%
\pgfsys@useobject{currentmarker}{}%
\end{pgfscope}%
\end{pgfscope}%
\begin{pgfscope}%
\definecolor{textcolor}{rgb}{0.000000,0.000000,0.000000}%
\pgfsetstrokecolor{textcolor}%
\pgfsetfillcolor{textcolor}%
\pgftext[x=0.970713in,y=0.098766in,,top]{\color{textcolor}\sffamily\fontsize{8.000000}{9.600000}\selectfont \(\displaystyle {50}\)}%
\end{pgfscope}%
\begin{pgfscope}%
\pgfsetbuttcap%
\pgfsetroundjoin%
\definecolor{currentfill}{rgb}{0.000000,0.000000,0.000000}%
\pgfsetfillcolor{currentfill}%
\pgfsetlinewidth{0.702625pt}%
\definecolor{currentstroke}{rgb}{0.000000,0.000000,0.000000}%
\pgfsetstrokecolor{currentstroke}%
\pgfsetdash{}{0pt}%
\pgfsys@defobject{currentmarker}{\pgfqpoint{0.000000in}{-0.038889in}}{\pgfqpoint{0.000000in}{0.000000in}}{%
\pgfpathmoveto{\pgfqpoint{0.000000in}{0.000000in}}%
\pgfpathlineto{\pgfqpoint{0.000000in}{-0.038889in}}%
\pgfusepath{stroke,fill}%
}%
\begin{pgfscope}%
\pgfsys@transformshift{1.506807in}{0.186266in}%
\pgfsys@useobject{currentmarker}{}%
\end{pgfscope}%
\end{pgfscope}%
\begin{pgfscope}%
\definecolor{textcolor}{rgb}{0.000000,0.000000,0.000000}%
\pgfsetstrokecolor{textcolor}%
\pgfsetfillcolor{textcolor}%
\pgftext[x=1.506807in,y=0.098766in,,top]{\color{textcolor}\sffamily\fontsize{8.000000}{9.600000}\selectfont \(\displaystyle {100}\)}%
\end{pgfscope}%
\begin{pgfscope}%
\pgfsetbuttcap%
\pgfsetroundjoin%
\definecolor{currentfill}{rgb}{0.000000,0.000000,0.000000}%
\pgfsetfillcolor{currentfill}%
\pgfsetlinewidth{0.702625pt}%
\definecolor{currentstroke}{rgb}{0.000000,0.000000,0.000000}%
\pgfsetstrokecolor{currentstroke}%
\pgfsetdash{}{0pt}%
\pgfsys@defobject{currentmarker}{\pgfqpoint{0.000000in}{-0.038889in}}{\pgfqpoint{0.000000in}{0.000000in}}{%
\pgfpathmoveto{\pgfqpoint{0.000000in}{0.000000in}}%
\pgfpathlineto{\pgfqpoint{0.000000in}{-0.038889in}}%
\pgfusepath{stroke,fill}%
}%
\begin{pgfscope}%
\pgfsys@transformshift{2.042901in}{0.186266in}%
\pgfsys@useobject{currentmarker}{}%
\end{pgfscope}%
\end{pgfscope}%
\begin{pgfscope}%
\definecolor{textcolor}{rgb}{0.000000,0.000000,0.000000}%
\pgfsetstrokecolor{textcolor}%
\pgfsetfillcolor{textcolor}%
\pgftext[x=2.042901in,y=0.098766in,,top]{\color{textcolor}\sffamily\fontsize{8.000000}{9.600000}\selectfont \(\displaystyle {150}\)}%
\end{pgfscope}%
\begin{pgfscope}%
\pgfsetbuttcap%
\pgfsetroundjoin%
\definecolor{currentfill}{rgb}{0.000000,0.000000,0.000000}%
\pgfsetfillcolor{currentfill}%
\pgfsetlinewidth{0.702625pt}%
\definecolor{currentstroke}{rgb}{0.000000,0.000000,0.000000}%
\pgfsetstrokecolor{currentstroke}%
\pgfsetdash{}{0pt}%
\pgfsys@defobject{currentmarker}{\pgfqpoint{0.000000in}{-0.038889in}}{\pgfqpoint{0.000000in}{0.000000in}}{%
\pgfpathmoveto{\pgfqpoint{0.000000in}{0.000000in}}%
\pgfpathlineto{\pgfqpoint{0.000000in}{-0.038889in}}%
\pgfusepath{stroke,fill}%
}%
\begin{pgfscope}%
\pgfsys@transformshift{2.578995in}{0.186266in}%
\pgfsys@useobject{currentmarker}{}%
\end{pgfscope}%
\end{pgfscope}%
\begin{pgfscope}%
\definecolor{textcolor}{rgb}{0.000000,0.000000,0.000000}%
\pgfsetstrokecolor{textcolor}%
\pgfsetfillcolor{textcolor}%
\pgftext[x=2.578995in,y=0.098766in,,top]{\color{textcolor}\sffamily\fontsize{8.000000}{9.600000}\selectfont \(\displaystyle {200}\)}%
\end{pgfscope}%
\begin{pgfscope}%
\pgfsetbuttcap%
\pgfsetroundjoin%
\definecolor{currentfill}{rgb}{0.000000,0.000000,0.000000}%
\pgfsetfillcolor{currentfill}%
\pgfsetlinewidth{0.702625pt}%
\definecolor{currentstroke}{rgb}{0.000000,0.000000,0.000000}%
\pgfsetstrokecolor{currentstroke}%
\pgfsetdash{}{0pt}%
\pgfsys@defobject{currentmarker}{\pgfqpoint{0.000000in}{-0.038889in}}{\pgfqpoint{0.000000in}{0.000000in}}{%
\pgfpathmoveto{\pgfqpoint{0.000000in}{0.000000in}}%
\pgfpathlineto{\pgfqpoint{0.000000in}{-0.038889in}}%
\pgfusepath{stroke,fill}%
}%
\begin{pgfscope}%
\pgfsys@transformshift{3.115089in}{0.186266in}%
\pgfsys@useobject{currentmarker}{}%
\end{pgfscope}%
\end{pgfscope}%
\begin{pgfscope}%
\definecolor{textcolor}{rgb}{0.000000,0.000000,0.000000}%
\pgfsetstrokecolor{textcolor}%
\pgfsetfillcolor{textcolor}%
\pgftext[x=3.115089in,y=0.098766in,,top]{\color{textcolor}\sffamily\fontsize{8.000000}{9.600000}\selectfont \(\displaystyle {250}\)}%
\end{pgfscope}%
\begin{pgfscope}%
\pgfsetbuttcap%
\pgfsetroundjoin%
\definecolor{currentfill}{rgb}{0.000000,0.000000,0.000000}%
\pgfsetfillcolor{currentfill}%
\pgfsetlinewidth{0.702625pt}%
\definecolor{currentstroke}{rgb}{0.000000,0.000000,0.000000}%
\pgfsetstrokecolor{currentstroke}%
\pgfsetdash{}{0pt}%
\pgfsys@defobject{currentmarker}{\pgfqpoint{-0.038889in}{0.000000in}}{\pgfqpoint{-0.000000in}{0.000000in}}{%
\pgfpathmoveto{\pgfqpoint{-0.000000in}{0.000000in}}%
\pgfpathlineto{\pgfqpoint{-0.038889in}{0.000000in}}%
\pgfusepath{stroke,fill}%
}%
\begin{pgfscope}%
\pgfsys@transformshift{0.297380in}{0.392805in}%
\pgfsys@useobject{currentmarker}{}%
\end{pgfscope}%
\end{pgfscope}%
\begin{pgfscope}%
\definecolor{textcolor}{rgb}{0.000000,0.000000,0.000000}%
\pgfsetstrokecolor{textcolor}%
\pgfsetfillcolor{textcolor}%
\pgftext[x=0.000000in, y=0.354225in, left, base]{\color{textcolor}\sffamily\fontsize{8.000000}{9.600000}\selectfont \(\displaystyle {0.30}\)}%
\end{pgfscope}%
\begin{pgfscope}%
\pgfsetbuttcap%
\pgfsetroundjoin%
\definecolor{currentfill}{rgb}{0.000000,0.000000,0.000000}%
\pgfsetfillcolor{currentfill}%
\pgfsetlinewidth{0.702625pt}%
\definecolor{currentstroke}{rgb}{0.000000,0.000000,0.000000}%
\pgfsetstrokecolor{currentstroke}%
\pgfsetdash{}{0pt}%
\pgfsys@defobject{currentmarker}{\pgfqpoint{-0.038889in}{0.000000in}}{\pgfqpoint{-0.000000in}{0.000000in}}{%
\pgfpathmoveto{\pgfqpoint{-0.000000in}{0.000000in}}%
\pgfpathlineto{\pgfqpoint{-0.038889in}{0.000000in}}%
\pgfusepath{stroke,fill}%
}%
\begin{pgfscope}%
\pgfsys@transformshift{0.297380in}{0.763261in}%
\pgfsys@useobject{currentmarker}{}%
\end{pgfscope}%
\end{pgfscope}%
\begin{pgfscope}%
\definecolor{textcolor}{rgb}{0.000000,0.000000,0.000000}%
\pgfsetstrokecolor{textcolor}%
\pgfsetfillcolor{textcolor}%
\pgftext[x=0.000000in, y=0.724680in, left, base]{\color{textcolor}\sffamily\fontsize{8.000000}{9.600000}\selectfont \(\displaystyle {0.35}\)}%
\end{pgfscope}%
\begin{pgfscope}%
\pgfsetbuttcap%
\pgfsetroundjoin%
\definecolor{currentfill}{rgb}{0.000000,0.000000,0.000000}%
\pgfsetfillcolor{currentfill}%
\pgfsetlinewidth{0.702625pt}%
\definecolor{currentstroke}{rgb}{0.000000,0.000000,0.000000}%
\pgfsetstrokecolor{currentstroke}%
\pgfsetdash{}{0pt}%
\pgfsys@defobject{currentmarker}{\pgfqpoint{-0.038889in}{0.000000in}}{\pgfqpoint{-0.000000in}{0.000000in}}{%
\pgfpathmoveto{\pgfqpoint{-0.000000in}{0.000000in}}%
\pgfpathlineto{\pgfqpoint{-0.038889in}{0.000000in}}%
\pgfusepath{stroke,fill}%
}%
\begin{pgfscope}%
\pgfsys@transformshift{0.297380in}{1.133716in}%
\pgfsys@useobject{currentmarker}{}%
\end{pgfscope}%
\end{pgfscope}%
\begin{pgfscope}%
\definecolor{textcolor}{rgb}{0.000000,0.000000,0.000000}%
\pgfsetstrokecolor{textcolor}%
\pgfsetfillcolor{textcolor}%
\pgftext[x=0.000000in, y=1.095136in, left, base]{\color{textcolor}\sffamily\fontsize{8.000000}{9.600000}\selectfont \(\displaystyle {0.40}\)}%
\end{pgfscope}%
\begin{pgfscope}%
\pgfsetbuttcap%
\pgfsetroundjoin%
\definecolor{currentfill}{rgb}{0.000000,0.000000,0.000000}%
\pgfsetfillcolor{currentfill}%
\pgfsetlinewidth{0.702625pt}%
\definecolor{currentstroke}{rgb}{0.000000,0.000000,0.000000}%
\pgfsetstrokecolor{currentstroke}%
\pgfsetdash{}{0pt}%
\pgfsys@defobject{currentmarker}{\pgfqpoint{-0.038889in}{0.000000in}}{\pgfqpoint{-0.000000in}{0.000000in}}{%
\pgfpathmoveto{\pgfqpoint{-0.000000in}{0.000000in}}%
\pgfpathlineto{\pgfqpoint{-0.038889in}{0.000000in}}%
\pgfusepath{stroke,fill}%
}%
\begin{pgfscope}%
\pgfsys@transformshift{0.297380in}{1.504171in}%
\pgfsys@useobject{currentmarker}{}%
\end{pgfscope}%
\end{pgfscope}%
\begin{pgfscope}%
\definecolor{textcolor}{rgb}{0.000000,0.000000,0.000000}%
\pgfsetstrokecolor{textcolor}%
\pgfsetfillcolor{textcolor}%
\pgftext[x=0.000000in, y=1.465591in, left, base]{\color{textcolor}\sffamily\fontsize{8.000000}{9.600000}\selectfont \(\displaystyle {0.45}\)}%
\end{pgfscope}%
\begin{pgfscope}%
\pgfsetbuttcap%
\pgfsetroundjoin%
\definecolor{currentfill}{rgb}{0.000000,0.000000,0.000000}%
\pgfsetfillcolor{currentfill}%
\pgfsetlinewidth{0.702625pt}%
\definecolor{currentstroke}{rgb}{0.000000,0.000000,0.000000}%
\pgfsetstrokecolor{currentstroke}%
\pgfsetdash{}{0pt}%
\pgfsys@defobject{currentmarker}{\pgfqpoint{-0.038889in}{0.000000in}}{\pgfqpoint{-0.000000in}{0.000000in}}{%
\pgfpathmoveto{\pgfqpoint{-0.000000in}{0.000000in}}%
\pgfpathlineto{\pgfqpoint{-0.038889in}{0.000000in}}%
\pgfusepath{stroke,fill}%
}%
\begin{pgfscope}%
\pgfsys@transformshift{0.297380in}{1.874627in}%
\pgfsys@useobject{currentmarker}{}%
\end{pgfscope}%
\end{pgfscope}%
\begin{pgfscope}%
\definecolor{textcolor}{rgb}{0.000000,0.000000,0.000000}%
\pgfsetstrokecolor{textcolor}%
\pgfsetfillcolor{textcolor}%
\pgftext[x=0.000000in, y=1.836047in, left, base]{\color{textcolor}\sffamily\fontsize{8.000000}{9.600000}\selectfont \(\displaystyle {0.50}\)}%
\end{pgfscope}%
\begin{pgfscope}%
\pgfsetbuttcap%
\pgfsetroundjoin%
\definecolor{currentfill}{rgb}{0.000000,0.000000,0.000000}%
\pgfsetfillcolor{currentfill}%
\pgfsetlinewidth{0.702625pt}%
\definecolor{currentstroke}{rgb}{0.000000,0.000000,0.000000}%
\pgfsetstrokecolor{currentstroke}%
\pgfsetdash{}{0pt}%
\pgfsys@defobject{currentmarker}{\pgfqpoint{-0.038889in}{0.000000in}}{\pgfqpoint{-0.000000in}{0.000000in}}{%
\pgfpathmoveto{\pgfqpoint{-0.000000in}{0.000000in}}%
\pgfpathlineto{\pgfqpoint{-0.038889in}{0.000000in}}%
\pgfusepath{stroke,fill}%
}%
\begin{pgfscope}%
\pgfsys@transformshift{0.297380in}{2.245082in}%
\pgfsys@useobject{currentmarker}{}%
\end{pgfscope}%
\end{pgfscope}%
\begin{pgfscope}%
\definecolor{textcolor}{rgb}{0.000000,0.000000,0.000000}%
\pgfsetstrokecolor{textcolor}%
\pgfsetfillcolor{textcolor}%
\pgftext[x=0.000000in, y=2.206502in, left, base]{\color{textcolor}\sffamily\fontsize{8.000000}{9.600000}\selectfont \(\displaystyle {0.55}\)}%
\end{pgfscope}%
\begin{pgfscope}%
\pgfsetbuttcap%
\pgfsetroundjoin%
\definecolor{currentfill}{rgb}{0.000000,0.000000,0.000000}%
\pgfsetfillcolor{currentfill}%
\pgfsetlinewidth{0.702625pt}%
\definecolor{currentstroke}{rgb}{0.000000,0.000000,0.000000}%
\pgfsetstrokecolor{currentstroke}%
\pgfsetdash{}{0pt}%
\pgfsys@defobject{currentmarker}{\pgfqpoint{-0.038889in}{0.000000in}}{\pgfqpoint{-0.000000in}{0.000000in}}{%
\pgfpathmoveto{\pgfqpoint{-0.000000in}{0.000000in}}%
\pgfpathlineto{\pgfqpoint{-0.038889in}{0.000000in}}%
\pgfusepath{stroke,fill}%
}%
\begin{pgfscope}%
\pgfsys@transformshift{0.297380in}{2.615538in}%
\pgfsys@useobject{currentmarker}{}%
\end{pgfscope}%
\end{pgfscope}%
\begin{pgfscope}%
\definecolor{textcolor}{rgb}{0.000000,0.000000,0.000000}%
\pgfsetstrokecolor{textcolor}%
\pgfsetfillcolor{textcolor}%
\pgftext[x=0.000000in, y=2.576958in, left, base]{\color{textcolor}\sffamily\fontsize{8.000000}{9.600000}\selectfont \(\displaystyle {0.60}\)}%
\end{pgfscope}%
\begin{pgfscope}%
\pgfsetbuttcap%
\pgfsetroundjoin%
\definecolor{currentfill}{rgb}{0.000000,0.000000,0.000000}%
\pgfsetfillcolor{currentfill}%
\pgfsetlinewidth{0.702625pt}%
\definecolor{currentstroke}{rgb}{0.000000,0.000000,0.000000}%
\pgfsetstrokecolor{currentstroke}%
\pgfsetdash{}{0pt}%
\pgfsys@defobject{currentmarker}{\pgfqpoint{-0.038889in}{0.000000in}}{\pgfqpoint{-0.000000in}{0.000000in}}{%
\pgfpathmoveto{\pgfqpoint{-0.000000in}{0.000000in}}%
\pgfpathlineto{\pgfqpoint{-0.038889in}{0.000000in}}%
\pgfusepath{stroke,fill}%
}%
\begin{pgfscope}%
\pgfsys@transformshift{0.297380in}{2.985993in}%
\pgfsys@useobject{currentmarker}{}%
\end{pgfscope}%
\end{pgfscope}%
\begin{pgfscope}%
\definecolor{textcolor}{rgb}{0.000000,0.000000,0.000000}%
\pgfsetstrokecolor{textcolor}%
\pgfsetfillcolor{textcolor}%
\pgftext[x=0.000000in, y=2.947413in, left, base]{\color{textcolor}\sffamily\fontsize{8.000000}{9.600000}\selectfont \(\displaystyle {0.65}\)}%
\end{pgfscope}%
\begin{pgfscope}%
\pgfpathrectangle{\pgfqpoint{0.297380in}{0.186266in}}{\pgfqpoint{3.019280in}{2.960256in}}%
\pgfusepath{clip}%
\pgfsetbuttcap%
\pgfsetroundjoin%
\pgfsetlinewidth{1.053937pt}%
\definecolor{currentstroke}{rgb}{0.121569,0.466667,0.705882}%
\pgfsetstrokecolor{currentstroke}%
\pgfsetdash{}{0pt}%
\pgfpathmoveto{\pgfqpoint{0.434620in}{0.320823in}}%
\pgfpathlineto{\pgfqpoint{0.434620in}{0.408478in}}%
\pgfusepath{stroke}%
\end{pgfscope}%
\begin{pgfscope}%
\pgfpathrectangle{\pgfqpoint{0.297380in}{0.186266in}}{\pgfqpoint{3.019280in}{2.960256in}}%
\pgfusepath{clip}%
\pgfsetbuttcap%
\pgfsetroundjoin%
\pgfsetlinewidth{1.053937pt}%
\definecolor{currentstroke}{rgb}{0.121569,0.466667,0.705882}%
\pgfsetstrokecolor{currentstroke}%
\pgfsetdash{}{0pt}%
\pgfpathmoveto{\pgfqpoint{0.606170in}{2.462054in}}%
\pgfpathlineto{\pgfqpoint{0.606170in}{2.596143in}}%
\pgfusepath{stroke}%
\end{pgfscope}%
\begin{pgfscope}%
\pgfpathrectangle{\pgfqpoint{0.297380in}{0.186266in}}{\pgfqpoint{3.019280in}{2.960256in}}%
\pgfusepath{clip}%
\pgfsetbuttcap%
\pgfsetroundjoin%
\pgfsetlinewidth{1.053937pt}%
\definecolor{currentstroke}{rgb}{0.121569,0.466667,0.705882}%
\pgfsetstrokecolor{currentstroke}%
\pgfsetdash{}{0pt}%
\pgfpathmoveto{\pgfqpoint{0.777720in}{2.586560in}}%
\pgfpathlineto{\pgfqpoint{0.777720in}{2.865801in}}%
\pgfusepath{stroke}%
\end{pgfscope}%
\begin{pgfscope}%
\pgfpathrectangle{\pgfqpoint{0.297380in}{0.186266in}}{\pgfqpoint{3.019280in}{2.960256in}}%
\pgfusepath{clip}%
\pgfsetbuttcap%
\pgfsetroundjoin%
\pgfsetlinewidth{1.053937pt}%
\definecolor{currentstroke}{rgb}{0.121569,0.466667,0.705882}%
\pgfsetstrokecolor{currentstroke}%
\pgfsetdash{}{0pt}%
\pgfpathmoveto{\pgfqpoint{0.949270in}{2.762681in}}%
\pgfpathlineto{\pgfqpoint{0.949270in}{2.923808in}}%
\pgfusepath{stroke}%
\end{pgfscope}%
\begin{pgfscope}%
\pgfpathrectangle{\pgfqpoint{0.297380in}{0.186266in}}{\pgfqpoint{3.019280in}{2.960256in}}%
\pgfusepath{clip}%
\pgfsetbuttcap%
\pgfsetroundjoin%
\pgfsetlinewidth{1.053937pt}%
\definecolor{currentstroke}{rgb}{0.121569,0.466667,0.705882}%
\pgfsetstrokecolor{currentstroke}%
\pgfsetdash{}{0pt}%
\pgfpathmoveto{\pgfqpoint{1.120820in}{2.741391in}}%
\pgfpathlineto{\pgfqpoint{1.120820in}{2.892740in}}%
\pgfusepath{stroke}%
\end{pgfscope}%
\begin{pgfscope}%
\pgfpathrectangle{\pgfqpoint{0.297380in}{0.186266in}}{\pgfqpoint{3.019280in}{2.960256in}}%
\pgfusepath{clip}%
\pgfsetbuttcap%
\pgfsetroundjoin%
\pgfsetlinewidth{1.053937pt}%
\definecolor{currentstroke}{rgb}{0.121569,0.466667,0.705882}%
\pgfsetstrokecolor{currentstroke}%
\pgfsetdash{}{0pt}%
\pgfpathmoveto{\pgfqpoint{1.292370in}{2.736998in}}%
\pgfpathlineto{\pgfqpoint{1.292370in}{2.950479in}}%
\pgfusepath{stroke}%
\end{pgfscope}%
\begin{pgfscope}%
\pgfpathrectangle{\pgfqpoint{0.297380in}{0.186266in}}{\pgfqpoint{3.019280in}{2.960256in}}%
\pgfusepath{clip}%
\pgfsetbuttcap%
\pgfsetroundjoin%
\pgfsetlinewidth{1.053937pt}%
\definecolor{currentstroke}{rgb}{0.121569,0.466667,0.705882}%
\pgfsetstrokecolor{currentstroke}%
\pgfsetdash{}{0pt}%
\pgfpathmoveto{\pgfqpoint{1.463920in}{2.598632in}}%
\pgfpathlineto{\pgfqpoint{1.463920in}{2.912015in}}%
\pgfusepath{stroke}%
\end{pgfscope}%
\begin{pgfscope}%
\pgfpathrectangle{\pgfqpoint{0.297380in}{0.186266in}}{\pgfqpoint{3.019280in}{2.960256in}}%
\pgfusepath{clip}%
\pgfsetbuttcap%
\pgfsetroundjoin%
\pgfsetlinewidth{1.053937pt}%
\definecolor{currentstroke}{rgb}{0.121569,0.466667,0.705882}%
\pgfsetstrokecolor{currentstroke}%
\pgfsetdash{}{0pt}%
\pgfpathmoveto{\pgfqpoint{1.635470in}{2.653933in}}%
\pgfpathlineto{\pgfqpoint{1.635470in}{2.945623in}}%
\pgfusepath{stroke}%
\end{pgfscope}%
\begin{pgfscope}%
\pgfpathrectangle{\pgfqpoint{0.297380in}{0.186266in}}{\pgfqpoint{3.019280in}{2.960256in}}%
\pgfusepath{clip}%
\pgfsetbuttcap%
\pgfsetroundjoin%
\pgfsetlinewidth{1.053937pt}%
\definecolor{currentstroke}{rgb}{0.121569,0.466667,0.705882}%
\pgfsetstrokecolor{currentstroke}%
\pgfsetdash{}{0pt}%
\pgfpathmoveto{\pgfqpoint{1.807020in}{2.704382in}}%
\pgfpathlineto{\pgfqpoint{1.807020in}{2.868501in}}%
\pgfusepath{stroke}%
\end{pgfscope}%
\begin{pgfscope}%
\pgfpathrectangle{\pgfqpoint{0.297380in}{0.186266in}}{\pgfqpoint{3.019280in}{2.960256in}}%
\pgfusepath{clip}%
\pgfsetbuttcap%
\pgfsetroundjoin%
\pgfsetlinewidth{1.053937pt}%
\definecolor{currentstroke}{rgb}{0.121569,0.466667,0.705882}%
\pgfsetstrokecolor{currentstroke}%
\pgfsetdash{}{0pt}%
\pgfpathmoveto{\pgfqpoint{1.978570in}{2.710050in}}%
\pgfpathlineto{\pgfqpoint{1.978570in}{2.953718in}}%
\pgfusepath{stroke}%
\end{pgfscope}%
\begin{pgfscope}%
\pgfpathrectangle{\pgfqpoint{0.297380in}{0.186266in}}{\pgfqpoint{3.019280in}{2.960256in}}%
\pgfusepath{clip}%
\pgfsetbuttcap%
\pgfsetroundjoin%
\pgfsetlinewidth{1.053937pt}%
\definecolor{currentstroke}{rgb}{0.121569,0.466667,0.705882}%
\pgfsetstrokecolor{currentstroke}%
\pgfsetdash{}{0pt}%
\pgfpathmoveto{\pgfqpoint{2.150120in}{2.704335in}}%
\pgfpathlineto{\pgfqpoint{2.150120in}{2.893244in}}%
\pgfusepath{stroke}%
\end{pgfscope}%
\begin{pgfscope}%
\pgfpathrectangle{\pgfqpoint{0.297380in}{0.186266in}}{\pgfqpoint{3.019280in}{2.960256in}}%
\pgfusepath{clip}%
\pgfsetbuttcap%
\pgfsetroundjoin%
\pgfsetlinewidth{1.053937pt}%
\definecolor{currentstroke}{rgb}{0.121569,0.466667,0.705882}%
\pgfsetstrokecolor{currentstroke}%
\pgfsetdash{}{0pt}%
\pgfpathmoveto{\pgfqpoint{2.321670in}{2.644294in}}%
\pgfpathlineto{\pgfqpoint{2.321670in}{2.785346in}}%
\pgfusepath{stroke}%
\end{pgfscope}%
\begin{pgfscope}%
\pgfpathrectangle{\pgfqpoint{0.297380in}{0.186266in}}{\pgfqpoint{3.019280in}{2.960256in}}%
\pgfusepath{clip}%
\pgfsetbuttcap%
\pgfsetroundjoin%
\pgfsetlinewidth{1.053937pt}%
\definecolor{currentstroke}{rgb}{0.121569,0.466667,0.705882}%
\pgfsetstrokecolor{currentstroke}%
\pgfsetdash{}{0pt}%
\pgfpathmoveto{\pgfqpoint{2.493220in}{2.702938in}}%
\pgfpathlineto{\pgfqpoint{2.493220in}{2.847224in}}%
\pgfusepath{stroke}%
\end{pgfscope}%
\begin{pgfscope}%
\pgfpathrectangle{\pgfqpoint{0.297380in}{0.186266in}}{\pgfqpoint{3.019280in}{2.960256in}}%
\pgfusepath{clip}%
\pgfsetbuttcap%
\pgfsetroundjoin%
\pgfsetlinewidth{1.053937pt}%
\definecolor{currentstroke}{rgb}{0.121569,0.466667,0.705882}%
\pgfsetstrokecolor{currentstroke}%
\pgfsetdash{}{0pt}%
\pgfpathmoveto{\pgfqpoint{2.664770in}{2.687332in}}%
\pgfpathlineto{\pgfqpoint{2.664770in}{2.810471in}}%
\pgfusepath{stroke}%
\end{pgfscope}%
\begin{pgfscope}%
\pgfpathrectangle{\pgfqpoint{0.297380in}{0.186266in}}{\pgfqpoint{3.019280in}{2.960256in}}%
\pgfusepath{clip}%
\pgfsetbuttcap%
\pgfsetroundjoin%
\pgfsetlinewidth{1.053937pt}%
\definecolor{currentstroke}{rgb}{0.121569,0.466667,0.705882}%
\pgfsetstrokecolor{currentstroke}%
\pgfsetdash{}{0pt}%
\pgfpathmoveto{\pgfqpoint{2.836320in}{2.782188in}}%
\pgfpathlineto{\pgfqpoint{2.836320in}{2.929986in}}%
\pgfusepath{stroke}%
\end{pgfscope}%
\begin{pgfscope}%
\pgfpathrectangle{\pgfqpoint{0.297380in}{0.186266in}}{\pgfqpoint{3.019280in}{2.960256in}}%
\pgfusepath{clip}%
\pgfsetbuttcap%
\pgfsetroundjoin%
\pgfsetlinewidth{1.053937pt}%
\definecolor{currentstroke}{rgb}{0.121569,0.466667,0.705882}%
\pgfsetstrokecolor{currentstroke}%
\pgfsetdash{}{0pt}%
\pgfpathmoveto{\pgfqpoint{3.007870in}{2.892847in}}%
\pgfpathlineto{\pgfqpoint{3.007870in}{3.011964in}}%
\pgfusepath{stroke}%
\end{pgfscope}%
\begin{pgfscope}%
\pgfpathrectangle{\pgfqpoint{0.297380in}{0.186266in}}{\pgfqpoint{3.019280in}{2.960256in}}%
\pgfusepath{clip}%
\pgfsetbuttcap%
\pgfsetroundjoin%
\pgfsetlinewidth{1.053937pt}%
\definecolor{currentstroke}{rgb}{0.121569,0.466667,0.705882}%
\pgfsetstrokecolor{currentstroke}%
\pgfsetdash{}{0pt}%
\pgfpathmoveto{\pgfqpoint{3.179420in}{2.872169in}}%
\pgfpathlineto{\pgfqpoint{3.179420in}{2.996090in}}%
\pgfusepath{stroke}%
\end{pgfscope}%
\begin{pgfscope}%
\pgfpathrectangle{\pgfqpoint{0.297380in}{0.186266in}}{\pgfqpoint{3.019280in}{2.960256in}}%
\pgfusepath{clip}%
\pgfsetrectcap%
\pgfsetroundjoin%
\pgfsetlinewidth{1.053937pt}%
\definecolor{currentstroke}{rgb}{0.121569,0.466667,0.705882}%
\pgfsetstrokecolor{currentstroke}%
\pgfsetdash{}{0pt}%
\pgfpathmoveto{\pgfqpoint{0.434620in}{0.364650in}}%
\pgfpathlineto{\pgfqpoint{0.606170in}{2.529098in}}%
\pgfpathlineto{\pgfqpoint{0.777720in}{2.726181in}}%
\pgfpathlineto{\pgfqpoint{0.949270in}{2.843244in}}%
\pgfpathlineto{\pgfqpoint{1.120820in}{2.817066in}}%
\pgfpathlineto{\pgfqpoint{1.292370in}{2.843738in}}%
\pgfpathlineto{\pgfqpoint{1.463920in}{2.755323in}}%
\pgfpathlineto{\pgfqpoint{1.635470in}{2.799778in}}%
\pgfpathlineto{\pgfqpoint{1.807020in}{2.786441in}}%
\pgfpathlineto{\pgfqpoint{1.978570in}{2.831884in}}%
\pgfpathlineto{\pgfqpoint{2.150120in}{2.798790in}}%
\pgfpathlineto{\pgfqpoint{2.321670in}{2.714820in}}%
\pgfpathlineto{\pgfqpoint{2.493220in}{2.775081in}}%
\pgfpathlineto{\pgfqpoint{2.664770in}{2.748902in}}%
\pgfpathlineto{\pgfqpoint{2.836320in}{2.856087in}}%
\pgfpathlineto{\pgfqpoint{3.007870in}{2.952405in}}%
\pgfpathlineto{\pgfqpoint{3.179420in}{2.934130in}}%
\pgfusepath{stroke}%
\end{pgfscope}%
\begin{pgfscope}%
\pgfsetrectcap%
\pgfsetmiterjoin%
\pgfsetlinewidth{0.702625pt}%
\definecolor{currentstroke}{rgb}{0.000000,0.000000,0.000000}%
\pgfsetstrokecolor{currentstroke}%
\pgfsetdash{}{0pt}%
\pgfpathmoveto{\pgfqpoint{0.297380in}{0.186266in}}%
\pgfpathlineto{\pgfqpoint{0.297380in}{3.146521in}}%
\pgfusepath{stroke}%
\end{pgfscope}%
\begin{pgfscope}%
\pgfsetrectcap%
\pgfsetmiterjoin%
\pgfsetlinewidth{0.702625pt}%
\definecolor{currentstroke}{rgb}{0.000000,0.000000,0.000000}%
\pgfsetstrokecolor{currentstroke}%
\pgfsetdash{}{0pt}%
\pgfpathmoveto{\pgfqpoint{3.316660in}{0.186266in}}%
\pgfpathlineto{\pgfqpoint{3.316660in}{3.146521in}}%
\pgfusepath{stroke}%
\end{pgfscope}%
\begin{pgfscope}%
\pgfsetrectcap%
\pgfsetmiterjoin%
\pgfsetlinewidth{0.702625pt}%
\definecolor{currentstroke}{rgb}{0.000000,0.000000,0.000000}%
\pgfsetstrokecolor{currentstroke}%
\pgfsetdash{}{0pt}%
\pgfpathmoveto{\pgfqpoint{0.297380in}{0.186266in}}%
\pgfpathlineto{\pgfqpoint{3.316660in}{0.186266in}}%
\pgfusepath{stroke}%
\end{pgfscope}%
\begin{pgfscope}%
\pgfsetrectcap%
\pgfsetmiterjoin%
\pgfsetlinewidth{0.702625pt}%
\definecolor{currentstroke}{rgb}{0.000000,0.000000,0.000000}%
\pgfsetstrokecolor{currentstroke}%
\pgfsetdash{}{0pt}%
\pgfpathmoveto{\pgfqpoint{0.297380in}{3.146521in}}%
\pgfpathlineto{\pgfqpoint{3.316660in}{3.146521in}}%
\pgfusepath{stroke}%
\end{pgfscope}%
\begin{pgfscope}%
\definecolor{textcolor}{rgb}{0.000000,0.000000,0.000000}%
\pgfsetstrokecolor{textcolor}%
\pgfsetfillcolor{textcolor}%
\pgftext[x=1.807020in,y=3.229854in,,base]{\color{textcolor}\sffamily\fontsize{9.000000}{10.800000}\selectfont OOD, Error on Train = 5\%}%
\end{pgfscope}%
\begin{pgfscope}%
\pgfsetbuttcap%
\pgfsetmiterjoin%
\definecolor{currentfill}{rgb}{1.000000,1.000000,1.000000}%
\pgfsetfillcolor{currentfill}%
\pgfsetlinewidth{0.000000pt}%
\definecolor{currentstroke}{rgb}{0.000000,0.000000,0.000000}%
\pgfsetstrokecolor{currentstroke}%
\pgfsetstrokeopacity{0.000000}%
\pgfsetdash{}{0pt}%
\pgfpathmoveto{\pgfqpoint{3.697380in}{0.186266in}}%
\pgfpathlineto{\pgfqpoint{6.716660in}{0.186266in}}%
\pgfpathlineto{\pgfqpoint{6.716660in}{3.146521in}}%
\pgfpathlineto{\pgfqpoint{3.697380in}{3.146521in}}%
\pgfpathlineto{\pgfqpoint{3.697380in}{0.186266in}}%
\pgfpathclose%
\pgfusepath{fill}%
\end{pgfscope}%
\begin{pgfscope}%
\pgfsetbuttcap%
\pgfsetroundjoin%
\definecolor{currentfill}{rgb}{0.000000,0.000000,0.000000}%
\pgfsetfillcolor{currentfill}%
\pgfsetlinewidth{0.702625pt}%
\definecolor{currentstroke}{rgb}{0.000000,0.000000,0.000000}%
\pgfsetstrokecolor{currentstroke}%
\pgfsetdash{}{0pt}%
\pgfsys@defobject{currentmarker}{\pgfqpoint{0.000000in}{-0.038889in}}{\pgfqpoint{0.000000in}{0.000000in}}{%
\pgfpathmoveto{\pgfqpoint{0.000000in}{0.000000in}}%
\pgfpathlineto{\pgfqpoint{0.000000in}{-0.038889in}}%
\pgfusepath{stroke,fill}%
}%
\begin{pgfscope}%
\pgfsys@transformshift{3.834620in}{0.186266in}%
\pgfsys@useobject{currentmarker}{}%
\end{pgfscope}%
\end{pgfscope}%
\begin{pgfscope}%
\definecolor{textcolor}{rgb}{0.000000,0.000000,0.000000}%
\pgfsetstrokecolor{textcolor}%
\pgfsetfillcolor{textcolor}%
\pgftext[x=3.834620in,y=0.098766in,,top]{\color{textcolor}\sffamily\fontsize{8.000000}{9.600000}\selectfont \(\displaystyle {0}\)}%
\end{pgfscope}%
\begin{pgfscope}%
\pgfsetbuttcap%
\pgfsetroundjoin%
\definecolor{currentfill}{rgb}{0.000000,0.000000,0.000000}%
\pgfsetfillcolor{currentfill}%
\pgfsetlinewidth{0.702625pt}%
\definecolor{currentstroke}{rgb}{0.000000,0.000000,0.000000}%
\pgfsetstrokecolor{currentstroke}%
\pgfsetdash{}{0pt}%
\pgfsys@defobject{currentmarker}{\pgfqpoint{0.000000in}{-0.038889in}}{\pgfqpoint{0.000000in}{0.000000in}}{%
\pgfpathmoveto{\pgfqpoint{0.000000in}{0.000000in}}%
\pgfpathlineto{\pgfqpoint{0.000000in}{-0.038889in}}%
\pgfusepath{stroke,fill}%
}%
\begin{pgfscope}%
\pgfsys@transformshift{4.370713in}{0.186266in}%
\pgfsys@useobject{currentmarker}{}%
\end{pgfscope}%
\end{pgfscope}%
\begin{pgfscope}%
\definecolor{textcolor}{rgb}{0.000000,0.000000,0.000000}%
\pgfsetstrokecolor{textcolor}%
\pgfsetfillcolor{textcolor}%
\pgftext[x=4.370713in,y=0.098766in,,top]{\color{textcolor}\sffamily\fontsize{8.000000}{9.600000}\selectfont \(\displaystyle {50}\)}%
\end{pgfscope}%
\begin{pgfscope}%
\pgfsetbuttcap%
\pgfsetroundjoin%
\definecolor{currentfill}{rgb}{0.000000,0.000000,0.000000}%
\pgfsetfillcolor{currentfill}%
\pgfsetlinewidth{0.702625pt}%
\definecolor{currentstroke}{rgb}{0.000000,0.000000,0.000000}%
\pgfsetstrokecolor{currentstroke}%
\pgfsetdash{}{0pt}%
\pgfsys@defobject{currentmarker}{\pgfqpoint{0.000000in}{-0.038889in}}{\pgfqpoint{0.000000in}{0.000000in}}{%
\pgfpathmoveto{\pgfqpoint{0.000000in}{0.000000in}}%
\pgfpathlineto{\pgfqpoint{0.000000in}{-0.038889in}}%
\pgfusepath{stroke,fill}%
}%
\begin{pgfscope}%
\pgfsys@transformshift{4.906807in}{0.186266in}%
\pgfsys@useobject{currentmarker}{}%
\end{pgfscope}%
\end{pgfscope}%
\begin{pgfscope}%
\definecolor{textcolor}{rgb}{0.000000,0.000000,0.000000}%
\pgfsetstrokecolor{textcolor}%
\pgfsetfillcolor{textcolor}%
\pgftext[x=4.906807in,y=0.098766in,,top]{\color{textcolor}\sffamily\fontsize{8.000000}{9.600000}\selectfont \(\displaystyle {100}\)}%
\end{pgfscope}%
\begin{pgfscope}%
\pgfsetbuttcap%
\pgfsetroundjoin%
\definecolor{currentfill}{rgb}{0.000000,0.000000,0.000000}%
\pgfsetfillcolor{currentfill}%
\pgfsetlinewidth{0.702625pt}%
\definecolor{currentstroke}{rgb}{0.000000,0.000000,0.000000}%
\pgfsetstrokecolor{currentstroke}%
\pgfsetdash{}{0pt}%
\pgfsys@defobject{currentmarker}{\pgfqpoint{0.000000in}{-0.038889in}}{\pgfqpoint{0.000000in}{0.000000in}}{%
\pgfpathmoveto{\pgfqpoint{0.000000in}{0.000000in}}%
\pgfpathlineto{\pgfqpoint{0.000000in}{-0.038889in}}%
\pgfusepath{stroke,fill}%
}%
\begin{pgfscope}%
\pgfsys@transformshift{5.442901in}{0.186266in}%
\pgfsys@useobject{currentmarker}{}%
\end{pgfscope}%
\end{pgfscope}%
\begin{pgfscope}%
\definecolor{textcolor}{rgb}{0.000000,0.000000,0.000000}%
\pgfsetstrokecolor{textcolor}%
\pgfsetfillcolor{textcolor}%
\pgftext[x=5.442901in,y=0.098766in,,top]{\color{textcolor}\sffamily\fontsize{8.000000}{9.600000}\selectfont \(\displaystyle {150}\)}%
\end{pgfscope}%
\begin{pgfscope}%
\pgfsetbuttcap%
\pgfsetroundjoin%
\definecolor{currentfill}{rgb}{0.000000,0.000000,0.000000}%
\pgfsetfillcolor{currentfill}%
\pgfsetlinewidth{0.702625pt}%
\definecolor{currentstroke}{rgb}{0.000000,0.000000,0.000000}%
\pgfsetstrokecolor{currentstroke}%
\pgfsetdash{}{0pt}%
\pgfsys@defobject{currentmarker}{\pgfqpoint{0.000000in}{-0.038889in}}{\pgfqpoint{0.000000in}{0.000000in}}{%
\pgfpathmoveto{\pgfqpoint{0.000000in}{0.000000in}}%
\pgfpathlineto{\pgfqpoint{0.000000in}{-0.038889in}}%
\pgfusepath{stroke,fill}%
}%
\begin{pgfscope}%
\pgfsys@transformshift{5.978995in}{0.186266in}%
\pgfsys@useobject{currentmarker}{}%
\end{pgfscope}%
\end{pgfscope}%
\begin{pgfscope}%
\definecolor{textcolor}{rgb}{0.000000,0.000000,0.000000}%
\pgfsetstrokecolor{textcolor}%
\pgfsetfillcolor{textcolor}%
\pgftext[x=5.978995in,y=0.098766in,,top]{\color{textcolor}\sffamily\fontsize{8.000000}{9.600000}\selectfont \(\displaystyle {200}\)}%
\end{pgfscope}%
\begin{pgfscope}%
\pgfsetbuttcap%
\pgfsetroundjoin%
\definecolor{currentfill}{rgb}{0.000000,0.000000,0.000000}%
\pgfsetfillcolor{currentfill}%
\pgfsetlinewidth{0.702625pt}%
\definecolor{currentstroke}{rgb}{0.000000,0.000000,0.000000}%
\pgfsetstrokecolor{currentstroke}%
\pgfsetdash{}{0pt}%
\pgfsys@defobject{currentmarker}{\pgfqpoint{0.000000in}{-0.038889in}}{\pgfqpoint{0.000000in}{0.000000in}}{%
\pgfpathmoveto{\pgfqpoint{0.000000in}{0.000000in}}%
\pgfpathlineto{\pgfqpoint{0.000000in}{-0.038889in}}%
\pgfusepath{stroke,fill}%
}%
\begin{pgfscope}%
\pgfsys@transformshift{6.515089in}{0.186266in}%
\pgfsys@useobject{currentmarker}{}%
\end{pgfscope}%
\end{pgfscope}%
\begin{pgfscope}%
\definecolor{textcolor}{rgb}{0.000000,0.000000,0.000000}%
\pgfsetstrokecolor{textcolor}%
\pgfsetfillcolor{textcolor}%
\pgftext[x=6.515089in,y=0.098766in,,top]{\color{textcolor}\sffamily\fontsize{8.000000}{9.600000}\selectfont \(\displaystyle {250}\)}%
\end{pgfscope}%
\begin{pgfscope}%
\pgfsetbuttcap%
\pgfsetroundjoin%
\definecolor{currentfill}{rgb}{0.000000,0.000000,0.000000}%
\pgfsetfillcolor{currentfill}%
\pgfsetlinewidth{0.702625pt}%
\definecolor{currentstroke}{rgb}{0.000000,0.000000,0.000000}%
\pgfsetstrokecolor{currentstroke}%
\pgfsetdash{}{0pt}%
\pgfsys@defobject{currentmarker}{\pgfqpoint{-0.038889in}{0.000000in}}{\pgfqpoint{-0.000000in}{0.000000in}}{%
\pgfpathmoveto{\pgfqpoint{-0.000000in}{0.000000in}}%
\pgfpathlineto{\pgfqpoint{-0.038889in}{0.000000in}}%
\pgfusepath{stroke,fill}%
}%
\begin{pgfscope}%
\pgfsys@transformshift{3.697380in}{0.447694in}%
\pgfsys@useobject{currentmarker}{}%
\end{pgfscope}%
\end{pgfscope}%
\begin{pgfscope}%
\definecolor{textcolor}{rgb}{0.000000,0.000000,0.000000}%
\pgfsetstrokecolor{textcolor}%
\pgfsetfillcolor{textcolor}%
\pgftext[x=3.400000in, y=0.409113in, left, base]{\color{textcolor}\sffamily\fontsize{8.000000}{9.600000}\selectfont \(\displaystyle {11.0}\)}%
\end{pgfscope}%
\begin{pgfscope}%
\pgfsetbuttcap%
\pgfsetroundjoin%
\definecolor{currentfill}{rgb}{0.000000,0.000000,0.000000}%
\pgfsetfillcolor{currentfill}%
\pgfsetlinewidth{0.702625pt}%
\definecolor{currentstroke}{rgb}{0.000000,0.000000,0.000000}%
\pgfsetstrokecolor{currentstroke}%
\pgfsetdash{}{0pt}%
\pgfsys@defobject{currentmarker}{\pgfqpoint{-0.038889in}{0.000000in}}{\pgfqpoint{-0.000000in}{0.000000in}}{%
\pgfpathmoveto{\pgfqpoint{-0.000000in}{0.000000in}}%
\pgfpathlineto{\pgfqpoint{-0.038889in}{0.000000in}}%
\pgfusepath{stroke,fill}%
}%
\begin{pgfscope}%
\pgfsys@transformshift{3.697380in}{0.833728in}%
\pgfsys@useobject{currentmarker}{}%
\end{pgfscope}%
\end{pgfscope}%
\begin{pgfscope}%
\definecolor{textcolor}{rgb}{0.000000,0.000000,0.000000}%
\pgfsetstrokecolor{textcolor}%
\pgfsetfillcolor{textcolor}%
\pgftext[x=3.400000in, y=0.795148in, left, base]{\color{textcolor}\sffamily\fontsize{8.000000}{9.600000}\selectfont \(\displaystyle {11.5}\)}%
\end{pgfscope}%
\begin{pgfscope}%
\pgfsetbuttcap%
\pgfsetroundjoin%
\definecolor{currentfill}{rgb}{0.000000,0.000000,0.000000}%
\pgfsetfillcolor{currentfill}%
\pgfsetlinewidth{0.702625pt}%
\definecolor{currentstroke}{rgb}{0.000000,0.000000,0.000000}%
\pgfsetstrokecolor{currentstroke}%
\pgfsetdash{}{0pt}%
\pgfsys@defobject{currentmarker}{\pgfqpoint{-0.038889in}{0.000000in}}{\pgfqpoint{-0.000000in}{0.000000in}}{%
\pgfpathmoveto{\pgfqpoint{-0.000000in}{0.000000in}}%
\pgfpathlineto{\pgfqpoint{-0.038889in}{0.000000in}}%
\pgfusepath{stroke,fill}%
}%
\begin{pgfscope}%
\pgfsys@transformshift{3.697380in}{1.219762in}%
\pgfsys@useobject{currentmarker}{}%
\end{pgfscope}%
\end{pgfscope}%
\begin{pgfscope}%
\definecolor{textcolor}{rgb}{0.000000,0.000000,0.000000}%
\pgfsetstrokecolor{textcolor}%
\pgfsetfillcolor{textcolor}%
\pgftext[x=3.400000in, y=1.181182in, left, base]{\color{textcolor}\sffamily\fontsize{8.000000}{9.600000}\selectfont \(\displaystyle {12.0}\)}%
\end{pgfscope}%
\begin{pgfscope}%
\pgfsetbuttcap%
\pgfsetroundjoin%
\definecolor{currentfill}{rgb}{0.000000,0.000000,0.000000}%
\pgfsetfillcolor{currentfill}%
\pgfsetlinewidth{0.702625pt}%
\definecolor{currentstroke}{rgb}{0.000000,0.000000,0.000000}%
\pgfsetstrokecolor{currentstroke}%
\pgfsetdash{}{0pt}%
\pgfsys@defobject{currentmarker}{\pgfqpoint{-0.038889in}{0.000000in}}{\pgfqpoint{-0.000000in}{0.000000in}}{%
\pgfpathmoveto{\pgfqpoint{-0.000000in}{0.000000in}}%
\pgfpathlineto{\pgfqpoint{-0.038889in}{0.000000in}}%
\pgfusepath{stroke,fill}%
}%
\begin{pgfscope}%
\pgfsys@transformshift{3.697380in}{1.605797in}%
\pgfsys@useobject{currentmarker}{}%
\end{pgfscope}%
\end{pgfscope}%
\begin{pgfscope}%
\definecolor{textcolor}{rgb}{0.000000,0.000000,0.000000}%
\pgfsetstrokecolor{textcolor}%
\pgfsetfillcolor{textcolor}%
\pgftext[x=3.400000in, y=1.567216in, left, base]{\color{textcolor}\sffamily\fontsize{8.000000}{9.600000}\selectfont \(\displaystyle {12.5}\)}%
\end{pgfscope}%
\begin{pgfscope}%
\pgfsetbuttcap%
\pgfsetroundjoin%
\definecolor{currentfill}{rgb}{0.000000,0.000000,0.000000}%
\pgfsetfillcolor{currentfill}%
\pgfsetlinewidth{0.702625pt}%
\definecolor{currentstroke}{rgb}{0.000000,0.000000,0.000000}%
\pgfsetstrokecolor{currentstroke}%
\pgfsetdash{}{0pt}%
\pgfsys@defobject{currentmarker}{\pgfqpoint{-0.038889in}{0.000000in}}{\pgfqpoint{-0.000000in}{0.000000in}}{%
\pgfpathmoveto{\pgfqpoint{-0.000000in}{0.000000in}}%
\pgfpathlineto{\pgfqpoint{-0.038889in}{0.000000in}}%
\pgfusepath{stroke,fill}%
}%
\begin{pgfscope}%
\pgfsys@transformshift{3.697380in}{1.991831in}%
\pgfsys@useobject{currentmarker}{}%
\end{pgfscope}%
\end{pgfscope}%
\begin{pgfscope}%
\definecolor{textcolor}{rgb}{0.000000,0.000000,0.000000}%
\pgfsetstrokecolor{textcolor}%
\pgfsetfillcolor{textcolor}%
\pgftext[x=3.400000in, y=1.953251in, left, base]{\color{textcolor}\sffamily\fontsize{8.000000}{9.600000}\selectfont \(\displaystyle {13.0}\)}%
\end{pgfscope}%
\begin{pgfscope}%
\pgfsetbuttcap%
\pgfsetroundjoin%
\definecolor{currentfill}{rgb}{0.000000,0.000000,0.000000}%
\pgfsetfillcolor{currentfill}%
\pgfsetlinewidth{0.702625pt}%
\definecolor{currentstroke}{rgb}{0.000000,0.000000,0.000000}%
\pgfsetstrokecolor{currentstroke}%
\pgfsetdash{}{0pt}%
\pgfsys@defobject{currentmarker}{\pgfqpoint{-0.038889in}{0.000000in}}{\pgfqpoint{-0.000000in}{0.000000in}}{%
\pgfpathmoveto{\pgfqpoint{-0.000000in}{0.000000in}}%
\pgfpathlineto{\pgfqpoint{-0.038889in}{0.000000in}}%
\pgfusepath{stroke,fill}%
}%
\begin{pgfscope}%
\pgfsys@transformshift{3.697380in}{2.377865in}%
\pgfsys@useobject{currentmarker}{}%
\end{pgfscope}%
\end{pgfscope}%
\begin{pgfscope}%
\definecolor{textcolor}{rgb}{0.000000,0.000000,0.000000}%
\pgfsetstrokecolor{textcolor}%
\pgfsetfillcolor{textcolor}%
\pgftext[x=3.400000in, y=2.339285in, left, base]{\color{textcolor}\sffamily\fontsize{8.000000}{9.600000}\selectfont \(\displaystyle {13.5}\)}%
\end{pgfscope}%
\begin{pgfscope}%
\pgfsetbuttcap%
\pgfsetroundjoin%
\definecolor{currentfill}{rgb}{0.000000,0.000000,0.000000}%
\pgfsetfillcolor{currentfill}%
\pgfsetlinewidth{0.702625pt}%
\definecolor{currentstroke}{rgb}{0.000000,0.000000,0.000000}%
\pgfsetstrokecolor{currentstroke}%
\pgfsetdash{}{0pt}%
\pgfsys@defobject{currentmarker}{\pgfqpoint{-0.038889in}{0.000000in}}{\pgfqpoint{-0.000000in}{0.000000in}}{%
\pgfpathmoveto{\pgfqpoint{-0.000000in}{0.000000in}}%
\pgfpathlineto{\pgfqpoint{-0.038889in}{0.000000in}}%
\pgfusepath{stroke,fill}%
}%
\begin{pgfscope}%
\pgfsys@transformshift{3.697380in}{2.763900in}%
\pgfsys@useobject{currentmarker}{}%
\end{pgfscope}%
\end{pgfscope}%
\begin{pgfscope}%
\definecolor{textcolor}{rgb}{0.000000,0.000000,0.000000}%
\pgfsetstrokecolor{textcolor}%
\pgfsetfillcolor{textcolor}%
\pgftext[x=3.400000in, y=2.725319in, left, base]{\color{textcolor}\sffamily\fontsize{8.000000}{9.600000}\selectfont \(\displaystyle {14.0}\)}%
\end{pgfscope}%
\begin{pgfscope}%
\pgfpathrectangle{\pgfqpoint{3.697380in}{0.186266in}}{\pgfqpoint{3.019280in}{2.960256in}}%
\pgfusepath{clip}%
\pgfsetbuttcap%
\pgfsetroundjoin%
\pgfsetlinewidth{1.053937pt}%
\definecolor{currentstroke}{rgb}{0.121569,0.466667,0.705882}%
\pgfsetstrokecolor{currentstroke}%
\pgfsetdash{}{0pt}%
\pgfpathmoveto{\pgfqpoint{3.834620in}{2.516653in}}%
\pgfpathlineto{\pgfqpoint{3.834620in}{3.011964in}}%
\pgfusepath{stroke}%
\end{pgfscope}%
\begin{pgfscope}%
\pgfpathrectangle{\pgfqpoint{3.697380in}{0.186266in}}{\pgfqpoint{3.019280in}{2.960256in}}%
\pgfusepath{clip}%
\pgfsetbuttcap%
\pgfsetroundjoin%
\pgfsetlinewidth{1.053937pt}%
\definecolor{currentstroke}{rgb}{0.121569,0.466667,0.705882}%
\pgfsetstrokecolor{currentstroke}%
\pgfsetdash{}{0pt}%
\pgfpathmoveto{\pgfqpoint{4.006170in}{1.042735in}}%
\pgfpathlineto{\pgfqpoint{4.006170in}{1.493936in}}%
\pgfusepath{stroke}%
\end{pgfscope}%
\begin{pgfscope}%
\pgfpathrectangle{\pgfqpoint{3.697380in}{0.186266in}}{\pgfqpoint{3.019280in}{2.960256in}}%
\pgfusepath{clip}%
\pgfsetbuttcap%
\pgfsetroundjoin%
\pgfsetlinewidth{1.053937pt}%
\definecolor{currentstroke}{rgb}{0.121569,0.466667,0.705882}%
\pgfsetstrokecolor{currentstroke}%
\pgfsetdash{}{0pt}%
\pgfpathmoveto{\pgfqpoint{4.177720in}{1.162993in}}%
\pgfpathlineto{\pgfqpoint{4.177720in}{1.615050in}}%
\pgfusepath{stroke}%
\end{pgfscope}%
\begin{pgfscope}%
\pgfpathrectangle{\pgfqpoint{3.697380in}{0.186266in}}{\pgfqpoint{3.019280in}{2.960256in}}%
\pgfusepath{clip}%
\pgfsetbuttcap%
\pgfsetroundjoin%
\pgfsetlinewidth{1.053937pt}%
\definecolor{currentstroke}{rgb}{0.121569,0.466667,0.705882}%
\pgfsetstrokecolor{currentstroke}%
\pgfsetdash{}{0pt}%
\pgfpathmoveto{\pgfqpoint{4.349270in}{1.008660in}}%
\pgfpathlineto{\pgfqpoint{4.349270in}{1.429656in}}%
\pgfusepath{stroke}%
\end{pgfscope}%
\begin{pgfscope}%
\pgfpathrectangle{\pgfqpoint{3.697380in}{0.186266in}}{\pgfqpoint{3.019280in}{2.960256in}}%
\pgfusepath{clip}%
\pgfsetbuttcap%
\pgfsetroundjoin%
\pgfsetlinewidth{1.053937pt}%
\definecolor{currentstroke}{rgb}{0.121569,0.466667,0.705882}%
\pgfsetstrokecolor{currentstroke}%
\pgfsetdash{}{0pt}%
\pgfpathmoveto{\pgfqpoint{4.520820in}{1.074053in}}%
\pgfpathlineto{\pgfqpoint{4.520820in}{1.356099in}}%
\pgfusepath{stroke}%
\end{pgfscope}%
\begin{pgfscope}%
\pgfpathrectangle{\pgfqpoint{3.697380in}{0.186266in}}{\pgfqpoint{3.019280in}{2.960256in}}%
\pgfusepath{clip}%
\pgfsetbuttcap%
\pgfsetroundjoin%
\pgfsetlinewidth{1.053937pt}%
\definecolor{currentstroke}{rgb}{0.121569,0.466667,0.705882}%
\pgfsetstrokecolor{currentstroke}%
\pgfsetdash{}{0pt}%
\pgfpathmoveto{\pgfqpoint{4.692370in}{0.967104in}}%
\pgfpathlineto{\pgfqpoint{4.692370in}{1.290541in}}%
\pgfusepath{stroke}%
\end{pgfscope}%
\begin{pgfscope}%
\pgfpathrectangle{\pgfqpoint{3.697380in}{0.186266in}}{\pgfqpoint{3.019280in}{2.960256in}}%
\pgfusepath{clip}%
\pgfsetbuttcap%
\pgfsetroundjoin%
\pgfsetlinewidth{1.053937pt}%
\definecolor{currentstroke}{rgb}{0.121569,0.466667,0.705882}%
\pgfsetstrokecolor{currentstroke}%
\pgfsetdash{}{0pt}%
\pgfpathmoveto{\pgfqpoint{4.863920in}{0.802015in}}%
\pgfpathlineto{\pgfqpoint{4.863920in}{1.446116in}}%
\pgfusepath{stroke}%
\end{pgfscope}%
\begin{pgfscope}%
\pgfpathrectangle{\pgfqpoint{3.697380in}{0.186266in}}{\pgfqpoint{3.019280in}{2.960256in}}%
\pgfusepath{clip}%
\pgfsetbuttcap%
\pgfsetroundjoin%
\pgfsetlinewidth{1.053937pt}%
\definecolor{currentstroke}{rgb}{0.121569,0.466667,0.705882}%
\pgfsetstrokecolor{currentstroke}%
\pgfsetdash{}{0pt}%
\pgfpathmoveto{\pgfqpoint{5.035470in}{0.771156in}}%
\pgfpathlineto{\pgfqpoint{5.035470in}{1.393780in}}%
\pgfusepath{stroke}%
\end{pgfscope}%
\begin{pgfscope}%
\pgfpathrectangle{\pgfqpoint{3.697380in}{0.186266in}}{\pgfqpoint{3.019280in}{2.960256in}}%
\pgfusepath{clip}%
\pgfsetbuttcap%
\pgfsetroundjoin%
\pgfsetlinewidth{1.053937pt}%
\definecolor{currentstroke}{rgb}{0.121569,0.466667,0.705882}%
\pgfsetstrokecolor{currentstroke}%
\pgfsetdash{}{0pt}%
\pgfpathmoveto{\pgfqpoint{5.207020in}{0.533382in}}%
\pgfpathlineto{\pgfqpoint{5.207020in}{0.985566in}}%
\pgfusepath{stroke}%
\end{pgfscope}%
\begin{pgfscope}%
\pgfpathrectangle{\pgfqpoint{3.697380in}{0.186266in}}{\pgfqpoint{3.019280in}{2.960256in}}%
\pgfusepath{clip}%
\pgfsetbuttcap%
\pgfsetroundjoin%
\pgfsetlinewidth{1.053937pt}%
\definecolor{currentstroke}{rgb}{0.121569,0.466667,0.705882}%
\pgfsetstrokecolor{currentstroke}%
\pgfsetdash{}{0pt}%
\pgfpathmoveto{\pgfqpoint{5.378570in}{0.320823in}}%
\pgfpathlineto{\pgfqpoint{5.378570in}{1.030891in}}%
\pgfusepath{stroke}%
\end{pgfscope}%
\begin{pgfscope}%
\pgfpathrectangle{\pgfqpoint{3.697380in}{0.186266in}}{\pgfqpoint{3.019280in}{2.960256in}}%
\pgfusepath{clip}%
\pgfsetbuttcap%
\pgfsetroundjoin%
\pgfsetlinewidth{1.053937pt}%
\definecolor{currentstroke}{rgb}{0.121569,0.466667,0.705882}%
\pgfsetstrokecolor{currentstroke}%
\pgfsetdash{}{0pt}%
\pgfpathmoveto{\pgfqpoint{5.550120in}{0.779759in}}%
\pgfpathlineto{\pgfqpoint{5.550120in}{1.096239in}}%
\pgfusepath{stroke}%
\end{pgfscope}%
\begin{pgfscope}%
\pgfpathrectangle{\pgfqpoint{3.697380in}{0.186266in}}{\pgfqpoint{3.019280in}{2.960256in}}%
\pgfusepath{clip}%
\pgfsetbuttcap%
\pgfsetroundjoin%
\pgfsetlinewidth{1.053937pt}%
\definecolor{currentstroke}{rgb}{0.121569,0.466667,0.705882}%
\pgfsetstrokecolor{currentstroke}%
\pgfsetdash{}{0pt}%
\pgfpathmoveto{\pgfqpoint{5.721670in}{0.413015in}}%
\pgfpathlineto{\pgfqpoint{5.721670in}{1.154926in}}%
\pgfusepath{stroke}%
\end{pgfscope}%
\begin{pgfscope}%
\pgfpathrectangle{\pgfqpoint{3.697380in}{0.186266in}}{\pgfqpoint{3.019280in}{2.960256in}}%
\pgfusepath{clip}%
\pgfsetbuttcap%
\pgfsetroundjoin%
\pgfsetlinewidth{1.053937pt}%
\definecolor{currentstroke}{rgb}{0.121569,0.466667,0.705882}%
\pgfsetstrokecolor{currentstroke}%
\pgfsetdash{}{0pt}%
\pgfpathmoveto{\pgfqpoint{5.893220in}{0.600427in}}%
\pgfpathlineto{\pgfqpoint{5.893220in}{0.901869in}}%
\pgfusepath{stroke}%
\end{pgfscope}%
\begin{pgfscope}%
\pgfpathrectangle{\pgfqpoint{3.697380in}{0.186266in}}{\pgfqpoint{3.019280in}{2.960256in}}%
\pgfusepath{clip}%
\pgfsetbuttcap%
\pgfsetroundjoin%
\pgfsetlinewidth{1.053937pt}%
\definecolor{currentstroke}{rgb}{0.121569,0.466667,0.705882}%
\pgfsetstrokecolor{currentstroke}%
\pgfsetdash{}{0pt}%
\pgfpathmoveto{\pgfqpoint{6.064770in}{0.387559in}}%
\pgfpathlineto{\pgfqpoint{6.064770in}{1.169860in}}%
\pgfusepath{stroke}%
\end{pgfscope}%
\begin{pgfscope}%
\pgfpathrectangle{\pgfqpoint{3.697380in}{0.186266in}}{\pgfqpoint{3.019280in}{2.960256in}}%
\pgfusepath{clip}%
\pgfsetbuttcap%
\pgfsetroundjoin%
\pgfsetlinewidth{1.053937pt}%
\definecolor{currentstroke}{rgb}{0.121569,0.466667,0.705882}%
\pgfsetstrokecolor{currentstroke}%
\pgfsetdash{}{0pt}%
\pgfpathmoveto{\pgfqpoint{6.236320in}{0.840913in}}%
\pgfpathlineto{\pgfqpoint{6.236320in}{1.306017in}}%
\pgfusepath{stroke}%
\end{pgfscope}%
\begin{pgfscope}%
\pgfpathrectangle{\pgfqpoint{3.697380in}{0.186266in}}{\pgfqpoint{3.019280in}{2.960256in}}%
\pgfusepath{clip}%
\pgfsetbuttcap%
\pgfsetroundjoin%
\pgfsetlinewidth{1.053937pt}%
\definecolor{currentstroke}{rgb}{0.121569,0.466667,0.705882}%
\pgfsetstrokecolor{currentstroke}%
\pgfsetdash{}{0pt}%
\pgfpathmoveto{\pgfqpoint{6.407870in}{1.678529in}}%
\pgfpathlineto{\pgfqpoint{6.407870in}{2.153687in}}%
\pgfusepath{stroke}%
\end{pgfscope}%
\begin{pgfscope}%
\pgfpathrectangle{\pgfqpoint{3.697380in}{0.186266in}}{\pgfqpoint{3.019280in}{2.960256in}}%
\pgfusepath{clip}%
\pgfsetbuttcap%
\pgfsetroundjoin%
\pgfsetlinewidth{1.053937pt}%
\definecolor{currentstroke}{rgb}{0.121569,0.466667,0.705882}%
\pgfsetstrokecolor{currentstroke}%
\pgfsetdash{}{0pt}%
\pgfpathmoveto{\pgfqpoint{6.579420in}{1.961138in}}%
\pgfpathlineto{\pgfqpoint{6.579420in}{2.485887in}}%
\pgfusepath{stroke}%
\end{pgfscope}%
\begin{pgfscope}%
\pgfpathrectangle{\pgfqpoint{3.697380in}{0.186266in}}{\pgfqpoint{3.019280in}{2.960256in}}%
\pgfusepath{clip}%
\pgfsetrectcap%
\pgfsetroundjoin%
\pgfsetlinewidth{1.053937pt}%
\definecolor{currentstroke}{rgb}{0.121569,0.466667,0.705882}%
\pgfsetstrokecolor{currentstroke}%
\pgfsetdash{}{0pt}%
\pgfpathmoveto{\pgfqpoint{3.834620in}{2.764309in}}%
\pgfpathlineto{\pgfqpoint{4.006170in}{1.268335in}}%
\pgfpathlineto{\pgfqpoint{4.177720in}{1.389022in}}%
\pgfpathlineto{\pgfqpoint{4.349270in}{1.219158in}}%
\pgfpathlineto{\pgfqpoint{4.520820in}{1.215076in}}%
\pgfpathlineto{\pgfqpoint{4.692370in}{1.128823in}}%
\pgfpathlineto{\pgfqpoint{4.863920in}{1.124066in}}%
\pgfpathlineto{\pgfqpoint{5.035470in}{1.082468in}}%
\pgfpathlineto{\pgfqpoint{5.207020in}{0.759474in}}%
\pgfpathlineto{\pgfqpoint{5.378570in}{0.675857in}}%
\pgfpathlineto{\pgfqpoint{5.550120in}{0.937999in}}%
\pgfpathlineto{\pgfqpoint{5.721670in}{0.783971in}}%
\pgfpathlineto{\pgfqpoint{5.893220in}{0.751148in}}%
\pgfpathlineto{\pgfqpoint{6.064770in}{0.778709in}}%
\pgfpathlineto{\pgfqpoint{6.236320in}{1.073465in}}%
\pgfpathlineto{\pgfqpoint{6.407870in}{1.916108in}}%
\pgfpathlineto{\pgfqpoint{6.579420in}{2.223512in}}%
\pgfusepath{stroke}%
\end{pgfscope}%
\begin{pgfscope}%
\pgfsetrectcap%
\pgfsetmiterjoin%
\pgfsetlinewidth{0.702625pt}%
\definecolor{currentstroke}{rgb}{0.000000,0.000000,0.000000}%
\pgfsetstrokecolor{currentstroke}%
\pgfsetdash{}{0pt}%
\pgfpathmoveto{\pgfqpoint{3.697380in}{0.186266in}}%
\pgfpathlineto{\pgfqpoint{3.697380in}{3.146521in}}%
\pgfusepath{stroke}%
\end{pgfscope}%
\begin{pgfscope}%
\pgfsetrectcap%
\pgfsetmiterjoin%
\pgfsetlinewidth{0.702625pt}%
\definecolor{currentstroke}{rgb}{0.000000,0.000000,0.000000}%
\pgfsetstrokecolor{currentstroke}%
\pgfsetdash{}{0pt}%
\pgfpathmoveto{\pgfqpoint{6.716660in}{0.186266in}}%
\pgfpathlineto{\pgfqpoint{6.716660in}{3.146521in}}%
\pgfusepath{stroke}%
\end{pgfscope}%
\begin{pgfscope}%
\pgfsetrectcap%
\pgfsetmiterjoin%
\pgfsetlinewidth{0.702625pt}%
\definecolor{currentstroke}{rgb}{0.000000,0.000000,0.000000}%
\pgfsetstrokecolor{currentstroke}%
\pgfsetdash{}{0pt}%
\pgfpathmoveto{\pgfqpoint{3.697380in}{0.186266in}}%
\pgfpathlineto{\pgfqpoint{6.716660in}{0.186266in}}%
\pgfusepath{stroke}%
\end{pgfscope}%
\begin{pgfscope}%
\pgfsetrectcap%
\pgfsetmiterjoin%
\pgfsetlinewidth{0.702625pt}%
\definecolor{currentstroke}{rgb}{0.000000,0.000000,0.000000}%
\pgfsetstrokecolor{currentstroke}%
\pgfsetdash{}{0pt}%
\pgfpathmoveto{\pgfqpoint{3.697380in}{3.146521in}}%
\pgfpathlineto{\pgfqpoint{6.716660in}{3.146521in}}%
\pgfusepath{stroke}%
\end{pgfscope}%
\begin{pgfscope}%
\definecolor{textcolor}{rgb}{0.000000,0.000000,0.000000}%
\pgfsetstrokecolor{textcolor}%
\pgfsetfillcolor{textcolor}%
\pgftext[x=5.207020in,y=3.229854in,,base]{\color{textcolor}\sffamily\fontsize{9.000000}{10.800000}\selectfont Scale, Error on Train = 5\%}%
\end{pgfscope}%
\end{pgfpicture}%
\makeatother%
\endgroup%